\documentclass{article}

\PassOptionsToPackage{numbers, compress}{natbib}

\usepackage[preprint]{neurips_2023}

\usepackage[utf8]{inputenc} 
\usepackage[T1]{fontenc}    
\usepackage[citecolor=blue, colorlinks]{hyperref}
\usepackage{url}            
\usepackage{booktabs}       
\usepackage{amsfonts}       
\usepackage{nicefrac}       
\usepackage{microtype}      
\usepackage{xcolor}         
\usepackage{multirow}
\usepackage{makecell}
\usepackage{subfigure}
\usepackage{float}
\usepackage{graphicx}
\usepackage{tabularx}
\usepackage{wrapfig}
\usepackage{hyperref}
\usepackage{makecell}
\usepackage{soul}
\usepackage{epsfig}
\usepackage{bbding}
\usepackage{upgreek}
\usepackage{amsmath}
\usepackage{amssymb}
\usepackage{fontawesome}
\usepackage{tikz}
\usepackage{amssymb}
\usepackage{pifont}
\newcommand{\xmark}{\ding{55}}%

\newcommand{\JHdel}[1]{\textcolor{red}{\st{#1}}}

\title{Unleash the Potential of 3D Point Cloud Modeling with A Calibrated Local Geometry-driven Distance Metric}

\author{%
  Siyu Ren and Junhui Hou\thanks{This work was supported by the Hong Kong
Research Grants Council under Grants 11202320 and 11219422. Corresponding author: Junhui Hou} \\
  Department of Computer Science, City University of Hong Kong\\
  \texttt{siyuren2-c@my.cityu.edu.hk; jh.hou@cityu.edu.hk} 
}

\begin{document}

\maketitle

\begin{abstract}
Quantifying the dissimilarity between two unstructured 3D point clouds is a challenging task, with existing metrics often relying on measuring the distance between corresponding points that can be either inefficient or ineffective. 
In this paper, we propose a novel distance metric called Calibrated Local Geometry Distance (CLGD), which computes the difference between the underlying 3D surfaces 
calibrated and induced by a set of reference points.
By associating each reference point with two given point clouds through computing its directional distances to them, the difference in directional distances of an identical reference point characterizes the geometric difference between a typical local region of the two point clouds.
Finally, CLGD is obtained by averaging the directional distance differences of all reference points. We evaluate CLGD on various optimization and unsupervised learning-based tasks,
including shape reconstruction, rigid registration, scene flow estimation, and feature representation. Extensive experiments show that CLGD achieves significantly higher accuracy under all tasks in a memory and computationally \textit{efficient} manner, compared with existing metrics. As a generic metric,  CLGD has the potential to advance 3D point cloud modeling. 
The source code is publicly available at \href{https://github.com/rsy6318/CLGD}{https://github.com/rsy6318/CLGD}.


\end{abstract}

\section{Introduction}
\label{sec:introduction}

\begin{wrapfigure}{r}{6.8cm} \small
\centering
\vspace{-0.5cm}
\subfigure[{\scriptsize EMD \cite{EMD}}]{\includegraphics[width=0.3\linewidth]{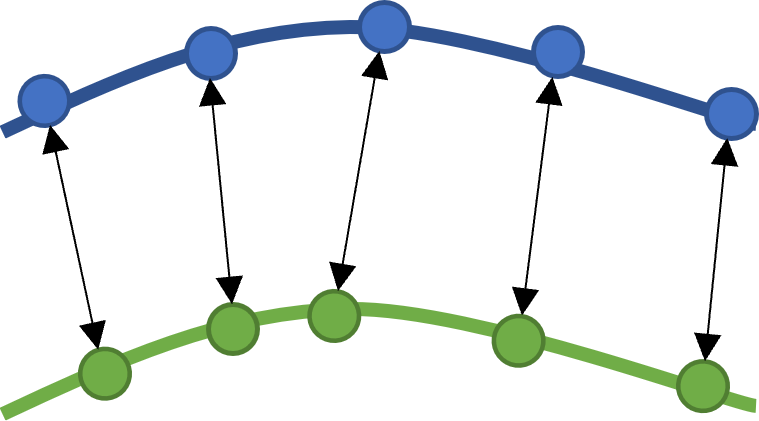}\label{CORR:EMD}}
\subfigure[{\scriptsize CD \cite{CD}}]{\includegraphics[width=0.3\linewidth]{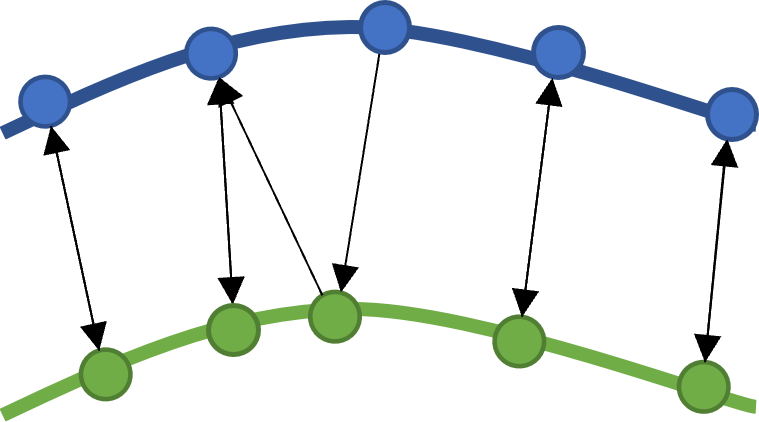}\label{CORR:CD}} 
\subfigure[{\scriptsize Ours}]{\includegraphics[width=0.3\linewidth]{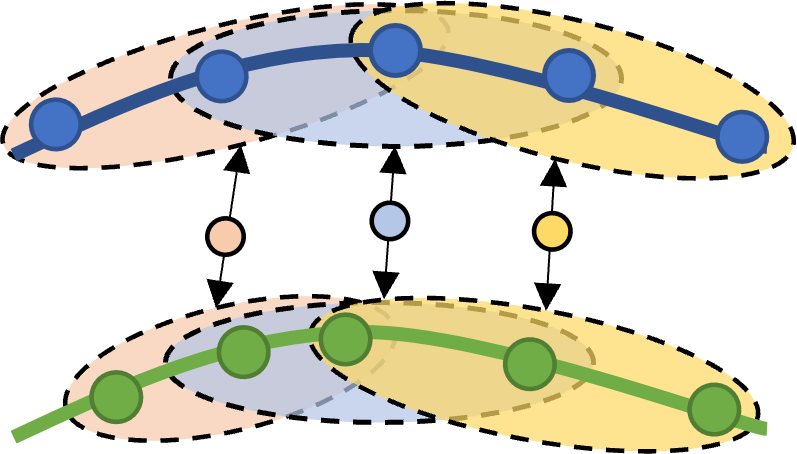}\label{CORR:GEO}} 
\vspace{-0.25cm}
\caption{  Visual illustration of different distance metrics for 3D point cloud data.
\includegraphics[width=0.03\linewidth]{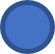} and \includegraphics[width=0.03\linewidth]{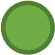} are two point clouds under evaluation,  \includegraphics[width=0.05\linewidth,height=0.01\linewidth]{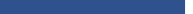} and \includegraphics[width=0.05\linewidth,height=0.01\linewidth]{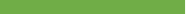} are their underlying surfaces, and the lines with arrows indicate the correspondence. In contrast to existing metrics measuring the difference between corresponding points, our metric computes the difference between the surfaces underlying 3D point clouds.}
\vspace{-0.4cm}
\end{wrapfigure}

3D point cloud data, which is a set of points defined by 3D coordinates to represent the geometric shape of an object or a scene, has been used in various fields, such as 
computer vision, 3D modeling, and robotics.
Measuring the difference between 3D point clouds is critical in many 
tasks, e.g., reconstruction, rigid registration, etc. 
Different from 2D images, where  pixel values are \textit{structured} with regular 2D coordinates, allowing us to directly compute the difference between two images pixel-by-pixel,  
3D point clouds are \textit{unstructured}, i.e., there is no 
point-wise correspondence naturally available between two point clouds, posing a great challenge. 
The two widely used distance metrics, namely Earth Mover's Distance (EMD) \cite{EMD} and Chamfer Distance (CD) \cite{CD} illustrated in Figs. \ref{CORR:EMD} and \ref{CORR:CD}, 
first build point-wise correspondence between two point clouds and then compute  
the distance between corresponding points. 
However, EMD is both memory and time-consuming as it involves solving a linear programming problem for the optimal bijection. For each point in one point cloud, CD seeks its nearest point in the other point cloud to establish the correspondence, and it could easily reach a local minimum. 
Although some improved distance metrics \cite{DCD,SWD,ARL,DPDIST} have been proposed, 
they are still either inefficient or ineffective. 


Existing distance metrics for 3D point cloud data generally concentrate on the point cloud itself,  aligning points to measure the point-wise difference. However, these metrics overlook the fact that different point clouds obtained by different sampling could represent an identical 3D surface.
In this paper, we propose an efficient yet effective distance metric named Calibrated Local Geometry Distance (CLGD). 
\textit{Unlike} previous metrics, CLGD computes the difference between the underlying 3D surfaces of two point clouds, 
as depicted in Fig. \ref{CORR:GEO}. 
Specifically, we first sample a set of 3D points called reference points, which we use to induce and calibrate the local geometry of the surfaces underlying point clouds, i.e., computing the \textit{directional distances} of each reference point to the two point clouds that approximately represent the underlying surfaces in implicit fields. 
Finally, we define CLGD as the average of 
the directional distance differences of all reference points.
We conduct extensive experiments on various tasks, including shape reconstruction, rigid registration, scene flow estimation, and feature representation, 
demonstrating its significant superiority over existing metrics.

In summary, the main contributions of this paper are: 
\begin{enumerate}
\vspace{-0.3cm}
 \item an efficient, effective, and generic distance metric for 3D point cloud data; 
\vspace{-0.13cm}
\item state-of-the-art benchmarks of rigid registration and scene flow estimation; and 
\vspace{-0.13cm}
\item potentially advancing the field of 3D point cloud processing and analysis.
\end{enumerate}
\vspace{-0.13cm}


\section{Related Work} \label{RELATED:WORKS}
\paragraph{Distance Metrics for 3D Point Clouds.}
Most of the existing distance metrics for 3D point clouds 
concentrate on the points in the point cloud. That is, they calculate the distance value based on the point-to-point distances from different point clouds, 
such as CD \cite{CD} and EMD \cite{EMD}. Specifically, EMD builds a global bi-directional mapping between the source and target point clouds. Then, the sum or mean of the distances between corresponding points is regarded as the distance between the point clouds. 
However, the computation of bijection is too expensive, especially when the number of points is large. Differently, CD builds a local mapping between point clouds by finding the nearest point in the other point cloud, making it more efficient than EMD. However, such local correspondence between point clouds may result in local minima or sub-optimal results. Hausdorff Distance (HD) \cite{HD} is modified from CD but focuses more on the outliers. Thus, it struggles to handle the details of point clouds and usually serves as an evaluation metric. 
Considering the distribution of point clouds, Wu \textit{et al.} \cite{DCD} proposed density-aware CD (DCD) by introducing density information as weights into CD, achieving a higher tolerance to the outliers.
Nguyen \textit{et al}. \cite{SWD} proposed Sliced Wasserstein Distance (SWD) 
to measure the distance between point clouds, making it more efficient and effective than EMD and CD in the shape representation task. PointNetLK \cite{POINTNETLK} and FMR \cite{FMR} convert point clouds to feature vectors through PointNet \cite{POINTNET} and utilize the distance of features to measure the difference between point clouds. However, the global feature cannot represent the details of the point clouds, and such a kind of distance metric relies heavily on the encoder. 

The above-mentioned distance metrics concentrate on points 
and ignore the geometric nature of a point cloud. 
In reality, the point clouds are sampled from the surface, and differently sampled point clouds could represent the same surface. Therefore, we should measure the difference between the underlying surfaces as the distance of the point clouds.
ARL \cite{ARL} randomly samples lines and calculates their intersections on the two point clouds' underlying surfaces approximately. It then calculates the difference between each line's intersections on the two point clouds to measure the dissimilarity of the two point clouds.
However, the calculation of the intersection is time-consuming, 
and the randomness of the lines could also make the measurement unstable. To leverage the underlying surfaces of point clouds, DPDist \cite{DPDIST} trains a network to regress the point-to-plane distance between the two point clouds. 
However, the trained network's accuracy in regressing the point-to-plane distance would decrease if the distribution of point cloud changes, limiting its generalization. 

\paragraph{Distance Metric-driven 3D Point Cloud Processing.} A distance metric is necessary for various 3D point cloud data processing and analysis tasks. One of the most common examples is the rigid registration of point clouds. The traditional registration methods, such as ICP \cite{ICP} and its variants \cite{ICP1,ICP3,FGR}, utilize the distance metrics between point clouds as the objective function to optimize. 
Some recent learning-based registration methods, such as DCP \cite{DCP}, FMR \cite{FMR}, and RPM-Net \cite{RPMNET}, could become unsupervised with the distance metrics as the alignment item in their loss functions during training. Besides, recent learning-based methods for scene flow estimation,
such as PointPWC-Net \cite{POINTPWC}, NSFP \cite{NEURALPRIOR}, and SCOOP \cite{SCOOP}, adopt  distance metrics 
as the alignment item in the loss functions to train the network without using ground-truth scene flow as supervision, and they have achieve remarkable accuracy. 
The distance metrics 
are also critical in some point cloud generation tasks, e.g., point cloud reconstruction \cite{FOLDINGNET,ATLAS}, upsampling \cite{ECNET,PUNET,PUGCN,PUGEO}, completion \cite{PCN,POINTTR,SEEDFORMER,SNOWFLAKENET,GRNET}, etc., where the difference between the generated and ground-truth point clouds is calculated as the main loss to train the networks.

\section{Proposed Method} \label{PROPOSED:METHOD}
\subsection{Problem Statement and Overview}
Given any two unstructured 3D point clouds $\mathbf{P}_{\rm 1}\in\mathbb{R}^{N_1\times 3}$ and $\mathbf{P}_{\rm 2}\in\mathbb{R}^{N_2\times 3}$ with $N_1$ and $N_2$ points respectively, we aim to construct a \textit{differentiable} distance metric, which can quantify the difference between them 
effectively and efficiently to drive downstream tasks. 
As mentioned in Section \ref{sec:introduction}, 
the problem is fundamentally challenging  due to the lack of  correspondence information between $\mathbf{P}_1$ and $\mathbf{P}_2$.  
Existing metrics generally focus on establishing the point-wise correspondence between $\mathbf{P}_1$ and $\mathbf{P}_2$ to compute the point-to-point difference, making them either ineffective or inefficient. 

In contrast to existing metrics, we address this problem by  measuring the difference between the underlying surfaces of $\mathbf{P}_1$ and $\mathbf{P}_2$. 
Generally, with a set of reference points generated, we associate each reference point with $\mathbf{P}_1$ and $\mathbf{P}_2$ to model their local surface geometry. Then we calculate the difference between the local surface geometry reference point-by-reference point and average the differences of all reference points as the final distance between $\mathbf{P}_1$ and $\mathbf{P}_2$. Such a process is memory-saving, computationally efficient, and effective. In what follows, we will introduce the technical details of the proposed distance metric.

\subsection{Generation of Reference Points} \label{REF:POINTS}
We generate a set of 3D points $\mathbf{Q}=\{\mathbf{q}_m\in\mathbb{R}^3\}_{m=1}^{M}$ named reference points, which will be used to \textit{indirectly} establish the correspondence between $\mathbf{P}_1$ and $\mathbf{P}_2$ and induce the local geometry of the surfaces underlying $\mathbf{P}_1$ and $\mathbf{P}_2$. 
Technically, after selecting either one of $\mathbf{P}_1$ and $\mathbf{P}_2$\footnote{For point cloud reconstruction-based tasks, where one point cloud is used to supervise a network to generate the other point cloud, the one used as supervision will be selected to generate $\mathbf{Q}$ because the reconstructed one is messy, lacking sufficient geometric meaning.} we add Gaussian noise to each point, 
where the standard deviation is $T$ times the distance to its nearest point in the point cloud, and repeat the noise addition process $R$ times randomly to generate $R$ reference points that are distributed \textit{near} to the underlying surface. 
Additionally, to ensure the local surface geometry induced by $\mathbf{Q}$ covers each point of $\mathbf{P}_1$ and $\mathbf{P}_2$, 
we also include 
the non-selected point cloud into $\mathbf{Q}$. 
See Table \ref{ABLATION:GEO}
for the ablative studies on this process.

\begin{wrapfigure}{r}{6.8cm}
	\centering
        \vspace{-1.3cm}
        \subfigure[]{\includegraphics[width=0.48\linewidth]{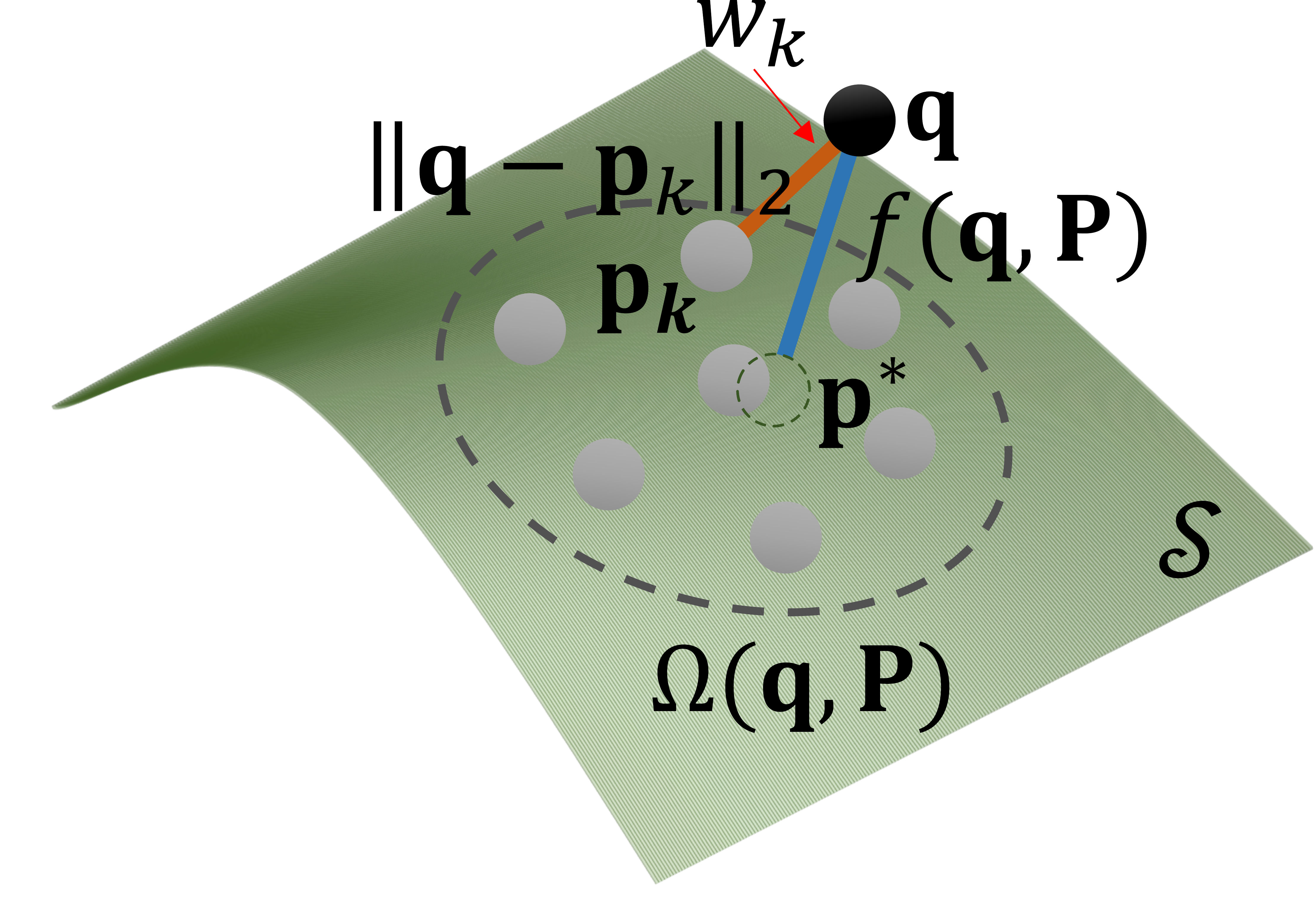}\label{VIS:DISTANCE}} 
        \subfigure[]{\includegraphics[width=0.48\linewidth]{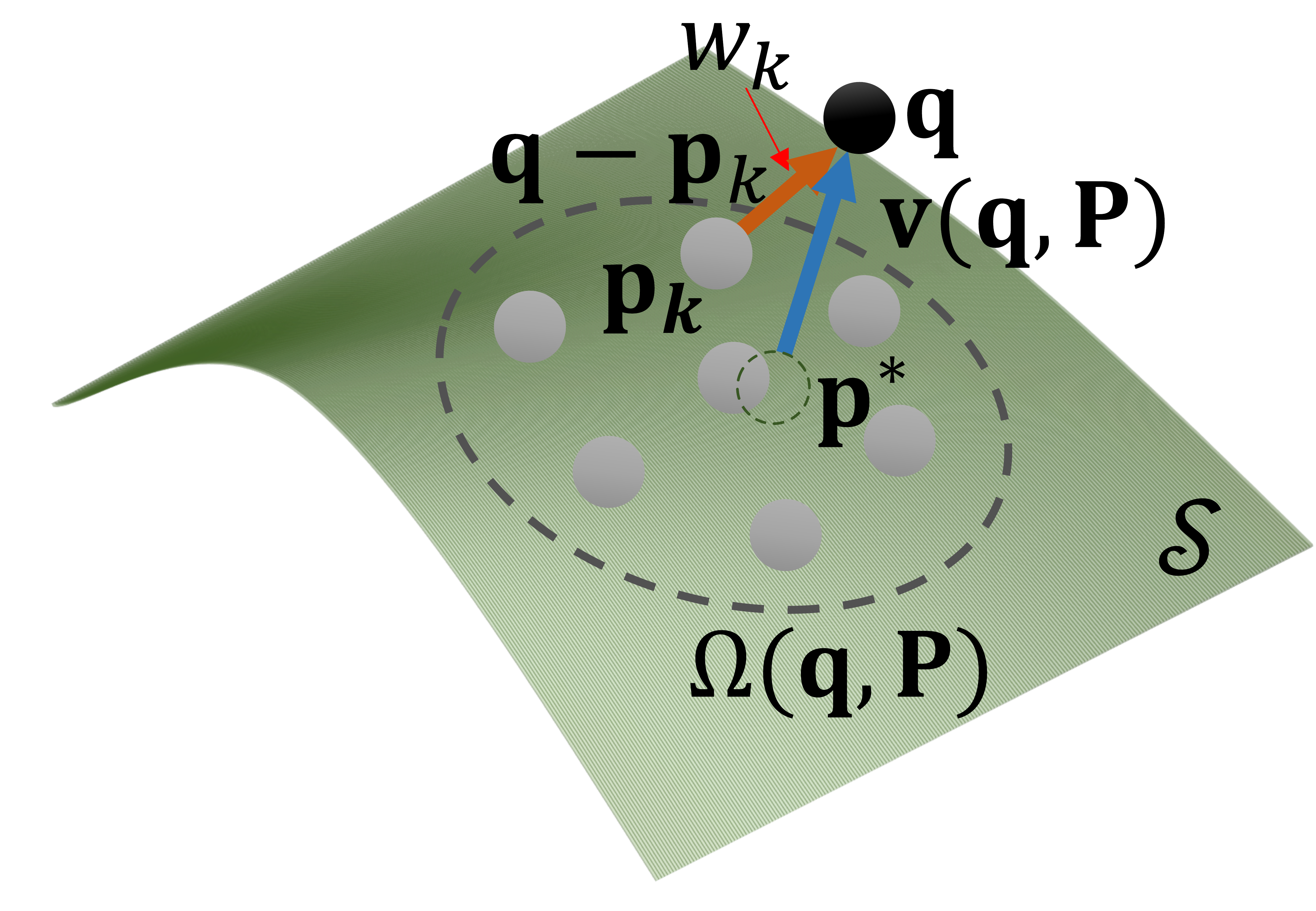}\label{VIS:DIRECTION}} 
        \vspace{-0.25cm}
	\caption{Illustration of the reference point-induced local surface geometry of a 3D point cloud. (a) $f(\mathbf{q},\mathbf{P})$. (b) $\mathbf{v}(\mathbf{q},\mathbf{P})$. 
 }
 \vspace{-0.6cm}
\end{wrapfigure}

\subsection{Calibrated Local Surface Geometry}
\label{LOCAL:GEO} 
Let $\mathbf{P}\in\{\mathbf{P}_1,\mathbf{P}_2\}$, 
$\mathbf{q}\in\mathbf{Q}$, 
and $\Omega(\mathbf{q},\mathbf{P}):=\{\mathbf{p}_k\}_{k=1}^{K}$ be the set of $\mathbf{q}$'s $K$-NN ($K$-Nearest Neighbor) points in $\mathbf{P}$. Note that we sort all points in $\Omega(\mathbf{q},\mathbf{P})$ according to their distances to $\mathbf{q}$, i.e., $\|\mathbf{q}-\mathbf{p}_1\|_2\leq\|\mathbf{q}-\mathbf{p}_2\|_2\leq...\leq\|\mathbf{q}-\mathbf{p}_K\|_2$, where $\|\cdot\|_2$ is the $\ell_2$ norm of a vector.
Let $\mathcal{S}$ be the underlying surface of $\mathbf{P}$, 
and $\mathbf{p}^*\in \mathcal{S}$ be the closest point to $\mathbf{q}$. As illustrated in Fig. \ref{VIS:DISTANCE}, we could simply approximate the distance from $\mathbf{q}$ to $\mathcal{S}$, i.e., the distance between $\mathbf{q}$ and $\mathbf{p}^*$, through the weighted averaging of $\{\|\mathbf{q}-\mathbf{p}_k\|_2\}_{k=1}^K$:
\begin{equation}\label{UDF}
    f(\mathbf{q},\mathbf{P}):=\|\mathbf{q}-\mathbf{p}^*\|_2\approx  \frac{\sum_{k=1}^{K}w(\mathbf{q},\mathbf{p}_k)\cdot\|\mathbf{q}-\mathbf{p}_k\|_2}{\sum_{k=1}^{K}w(\mathbf{q},\mathbf{p}_k)}, \ {\rm where} \ w(\mathbf{q},\mathbf{p}_k)=\frac{1}{\|\mathbf{q}-\mathbf{p}_k\|_2^2}.
\end{equation}

Similarly, as shown in Fig. \ref{VIS:DIRECTION}, we could also approximate the vector from $\mathbf{p}^*$ to $\mathbf{q}$ with the weighted averaging of $\{\mathbf{q}-\mathbf{p}_k\}_{k=1}^K$:
\begin{equation}\label{UDF_GRAD}
\mathbf{v}(\mathbf{q},\mathbf{P}):=\mathbf{q}-\mathbf{p}^*\approx\frac{\sum_{k=1}^{K}w(\mathbf{q},\mathbf{p}_k)\cdot(\mathbf{q}-\mathbf{p}_k)}{\sum_{k=1}^{K}w(\mathbf{q},\mathbf{p}_k)}.
\end{equation}
Then, we concatenate $f(\mathbf{q},\mathbf{P})$ and $\mathbf{v}(\mathbf{q},\mathbf{P})$ to form a 4D vector $\mathbf{g}(\mathbf{q},\mathbf{P})=[f(\mathbf{q},\mathbf{P})||\mathbf{v}(\mathbf{q},\mathbf{P})]\in\mathbb{R}^4$ named \textit{directional distance}, which characterizes the local geometry of the surface underlying $\mathbf{P}$ induced by 
$\mathbf{q}$ in implicit fields. 

\subsection{Local Surface Geometry-driven Distance Metric} \label{DIFF}
Calibrated by the reference point $\mathbf{q}_m\in\mathbf{Q}$, the difference between its 
directional distances 
to $\mathbf{P}_1$ and $\mathbf{P}_2$ can reflect the difference in their local surface geometry, 
i.e., the difference between 
$\Omega(\mathbf{q}_m,~\mathbf{P}_1)$ and $\Omega(\mathbf{q}_m,~\mathbf{P}_2)$. Specifically, we calculate the difference between the directional distances of $\mathbf{q}_m$ as  
\begin{equation}
    d(\mathbf{q}_m,\mathbf{P}_{\rm 1}, \mathbf{P}_{\rm 2})=\|\mathbf{g}(\mathbf{q}_m,\mathbf{P}_{\rm 1})-\mathbf{g}(\mathbf{q}_m,\mathbf{P}_{\rm 2})\|_1,
\end{equation}
where $\|\cdot\|_1$ computes the $\ell_1$ norm of a vector. Note that $\mathbf{g}(\mathbf{q}_m,\mathbf{P}_{\rm 1})$ and $\mathbf{g}(\mathbf{q}_m,\mathbf{P}_{\rm 2})$ share \textit{identical} weights $\{ w \}_{k=1}^{K}$, i.e., the weights of the point cloud selected to generate reference points are also applied to the other point cloud. 
The proposed distance metric for 3D point clouds is finally defined as the weighted sum of $d(\mathbf{q}_m,\mathbf{P}_1,\mathbf{P}_2)$:
\begin{equation} \label{CLGD}
    \mathcal{D}_{\rm CLGD}(\mathbf{P}_{\rm 1}, \mathbf{P}_{\rm 2})=\frac{1}{M}\sum_{\mathbf{q}_m\in\mathbf{Q}}s(\mathbf{q}_m)\cdot d(\mathbf{q}_m,\mathbf{P}_{\rm 1}, \mathbf{P}_{\rm 2}), 
\end{equation}
where $s(\mathbf{q}_m)=\texttt{Exp}(-\beta\cdot d(\mathbf{q}_m,\mathbf{P}_1,\mathbf{P}_2))$ is the confidence score of $d(\mathbf{q}_m,~\mathbf{P}_1,~\mathbf{P}_2)$ with $\beta\geq 0$ being a hyperparameter. 
$s(\mathbf{q}_m)$ is introduced to cope with the case where $\mathbf{P}_1$ and $\mathbf{P}_2$ are partially overlapped.


\section{Experiments}
\label{EXPERIMENT}
To demonstrate the effectiveness and superiority of the proposed CLGD, we applied it to a wide range of downstream tasks, including shape reconstruction, rigid point cloud registration, scene flow estimation, and feature representation, and compared it with 
EMD \cite{EMD}, CD \cite{CD}, and ARL \cite{ARL}\footnote{
ARL was primarily designed for rigid point cloud registration, and thus, we compared our CLGD with ARL only in the registration task.}. 
In all experiments,
we set $R=10$, $T=3$, and $K=5$. 
We set $\beta=3$ in rigid registration due to the partially overlapping point clouds, and $\beta=0$ in the other three tasks. 
We conducted all experiments on an NVIDIA RTX 3090 with Intel(R) Xeon(R) CPU.

\subsection{3D Shape Reconstruction} 
We consider a learning-based point cloud shape reconstruction task. 
Technically, we followed FoldingNet \cite{FOLDINGNET} to construct a reconstruction network, where 
regular 2D grids distributed on a square area $[-\delta,~\delta]^2$ 
are fed into MLPs to regress 3D point clouds. 
The network was trained by minimizing the distance between the reconstructed point cloud and the given point cloud
in an overfitting manner, i.e., each shape has its individual network parameters. Additionally, based on the mechanism of the reconstruction network and 
differential geometry, we could obtain the normal vectors of the reconstructed point cloud through the backpropagation of the network. 
Finally, we used SPSR \cite{SPSR} to recover the mesh from the resulting point cloud and its normal vectors. We refer the readers to the \textit{Supplementary Material} for more details.

\paragraph{Implementation Details.} We utilized three categories of the ShapeNet dataset \cite{SHAPENET}, namely chair, sofa, and table, each containing 200 randomly selected shapes.
For each shape, we normalized it within a unit cube and sampled 4096 points from its surface uniformly using PDS \cite{PDS} to get the point cloud. 
As for the input regular 2D grids, we set $\delta=0.3$ and the size to be 
$64\times 64$. We used the ADAM optimizer to optimize the network for $10^4$ iterations with a learning rate of $10^{-3}$. 


\begin{table}[h] \small 
\centering
\vspace{-0.3cm}
\caption{Quantitative comparisons of reconstructed point clouds and surfaces under different loss functions. The best results are highlighted in \textbf{bold}. $\downarrow$ (resp. $\uparrow$) indicates the smaller (resp. the larger), the better.}\label{OVERFITTING}
\begin{tabular}{l|l|c c c | c c c } 
\toprule
\multirow{2}{*}{Shape} & \multirow{2}{*}{Loss} & \multicolumn{3}{c|}{Point Cloud} & \multicolumn{3}{c}{Triangle Mesh}  \\
\cline{3-8}
 &  & CD ($\times 10^{-2}$) $\downarrow$ & HD ($\times 10^{-2}$) $\downarrow$ & P2F ($\times 10^{-3}$) $\downarrow$ & NC $\uparrow$ & F-0.5\% $\uparrow$ & F-1\% $\uparrow$ \\
\hline
\multirow{3}{*}{Chair} & EMD    & 2.935 & 12.628 & 9.777 & 0.781 & 0.277 & 0.524   \\
                        & CD    & 2.221 & 9.430 & 4.543 & 0.839 & 0.465 & 0.721  \\
                        & Ours  & 1.898 & 6.787 & 2.591 & 0.908 & 0.709 & 0.913  \\
\hline
\multirow{3}{*}{Sofa}   & EMD   & 2.534 & 7.355 & 7.433 & 0.879 & 0.459 & 0.717  \\
                        & CD    & 1.972 & 5.323 & 3.392 & 0.920 & 0.668 & 0.887 \\
                        & Ours  & 1.770 & 4.191 & 2.040 & 0.940 & 0.806 & 0.949  \\   
\hline
\multirow{3}{*}{Table}  & EMD & 2.996 & 11.374 & 8.921 & 0.768 & 0.243 & 0.473  \\
                        & CD & 2.272 & 8.881 & 4.353 & 0.824 & 0.403 & 0.658 \\
                        & Ours & 1.974 & 6.302 & 2.480 & 0.900 & 0.643 & 0.873 \\  
\hline
\multirow{3}{*}{\textbf{Average}} & EMD & 2.821 & 10.452 & 8.720 & 0.809 & 0.326 & 0.571  \\
                     & CD & 2.155 & 7.878 & 4.096 & 0.861 & 0.521 & 0.755  \\
                    & Ours & \textbf{1.880} & \textbf{5.760} & \textbf{2.370} & \textbf{0.916} & \textbf{0.719} & \textbf{0.911}  \\   
\bottomrule
\end{tabular}
\end{table}

\begin{figure}[h] \small 
\centering
{
\begin{tikzpicture}[]
\node[] (a) at (0.5,6) {\includegraphics[width=0.14\textwidth]{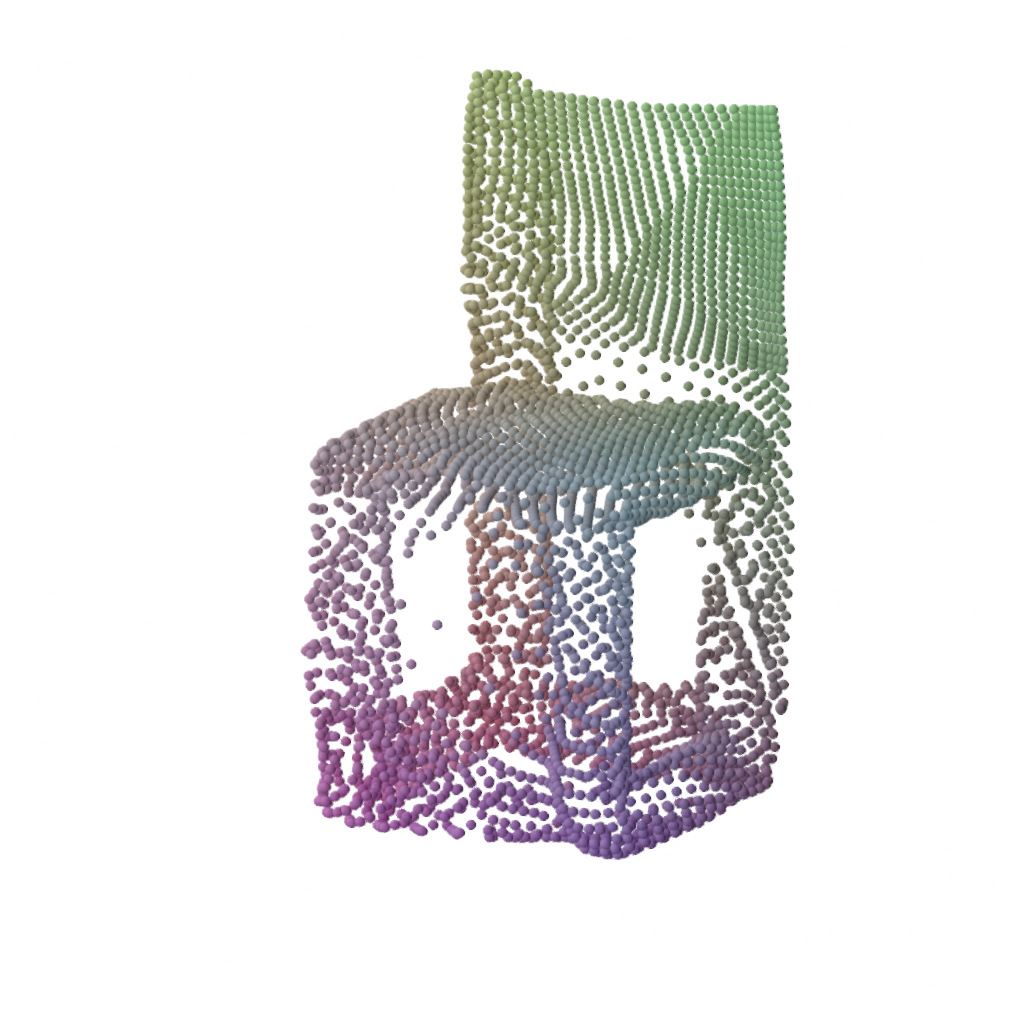} };
\node[] (a) at (0.5-0.1,4) {\includegraphics[width=0.14\textwidth]{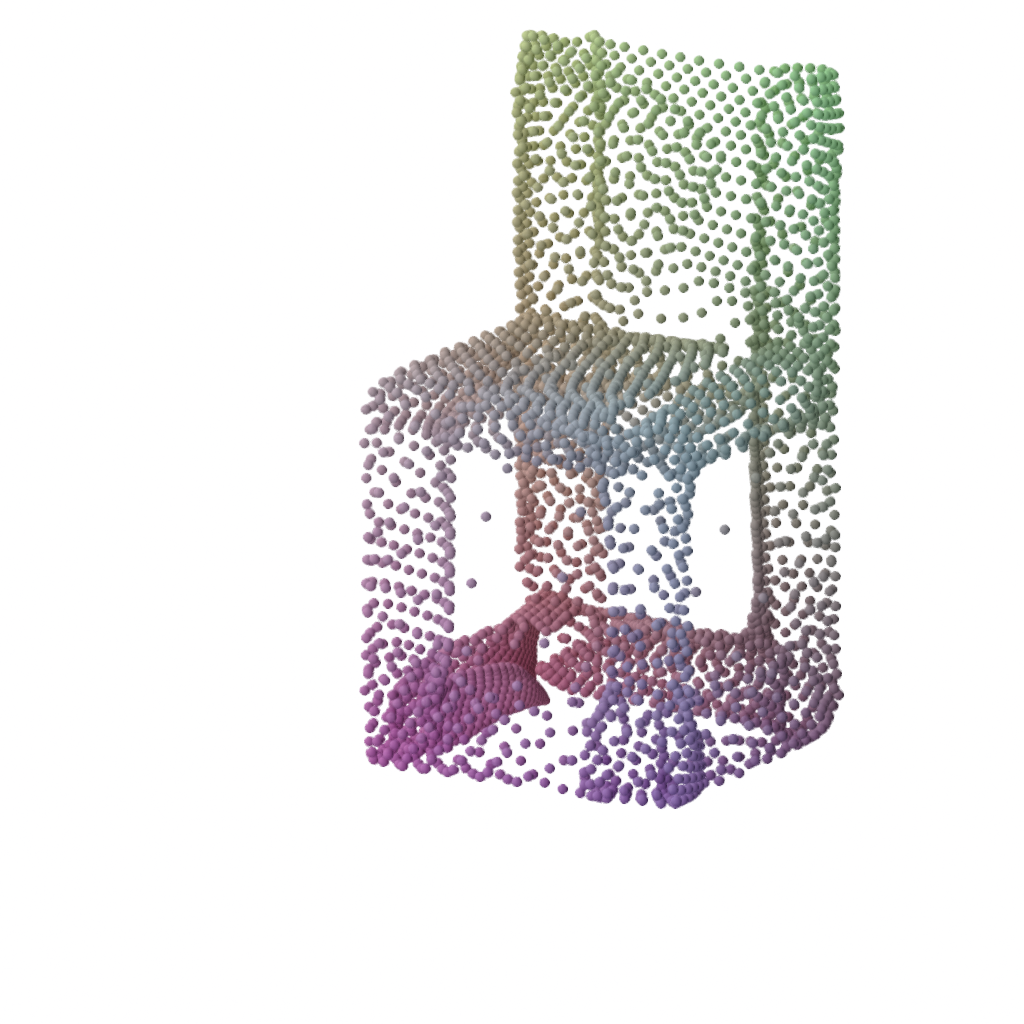}} ;
\node[] (a) at (0.5,2) {\includegraphics[width=0.14\textwidth]{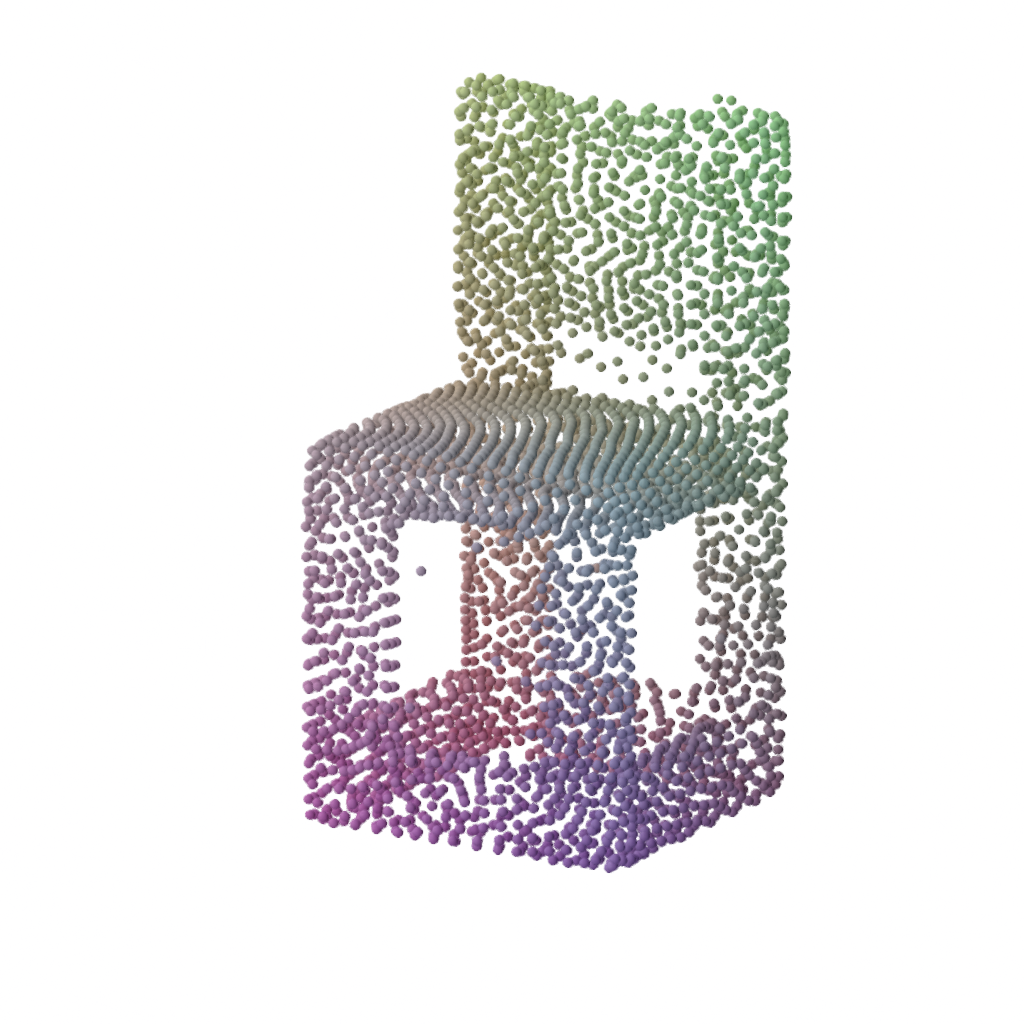}} ;
\node[] (a) at (0.5,0) {\includegraphics[width=0.14\textwidth]{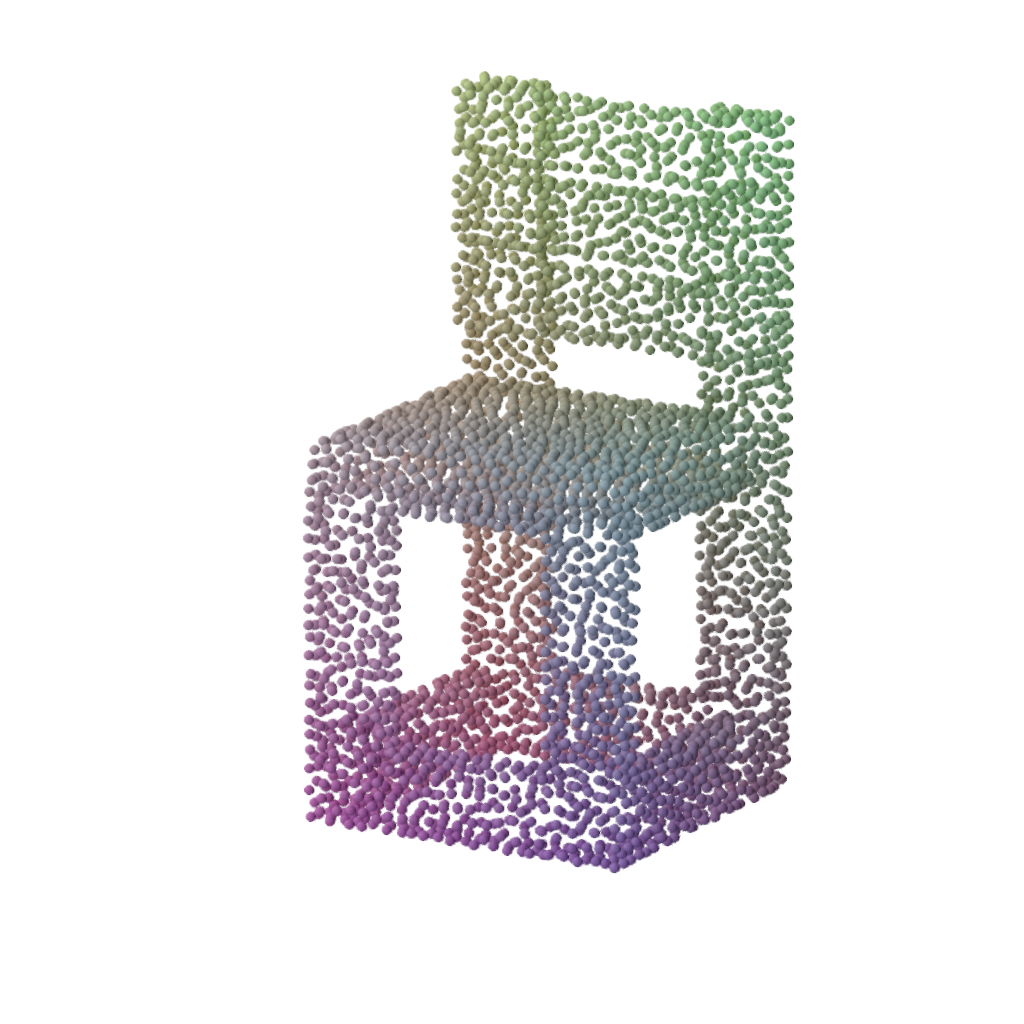}} ;

\node[] (b) at (19/8,6) {\includegraphics[width=0.14\textwidth]{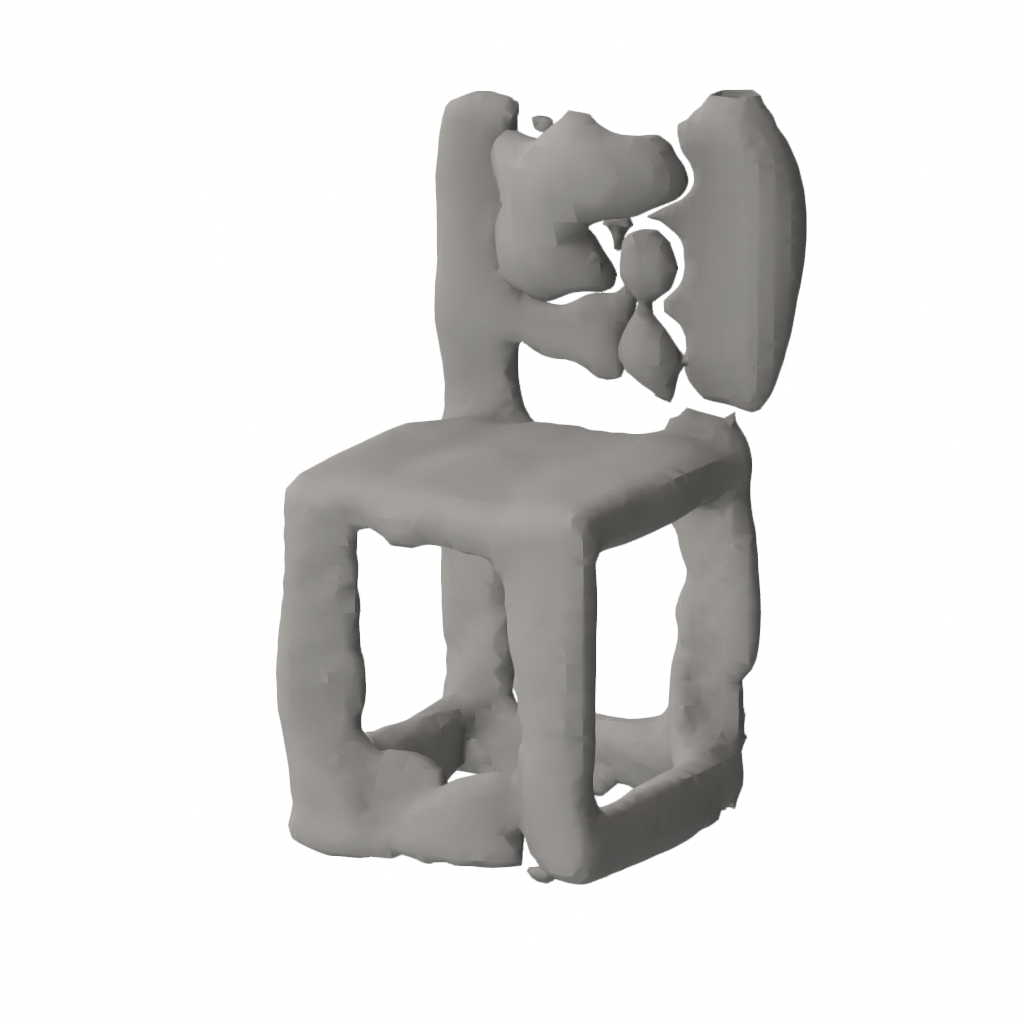}};
\node[] (b) at (19/8-0.1,4) {\includegraphics[width=0.14\textwidth]{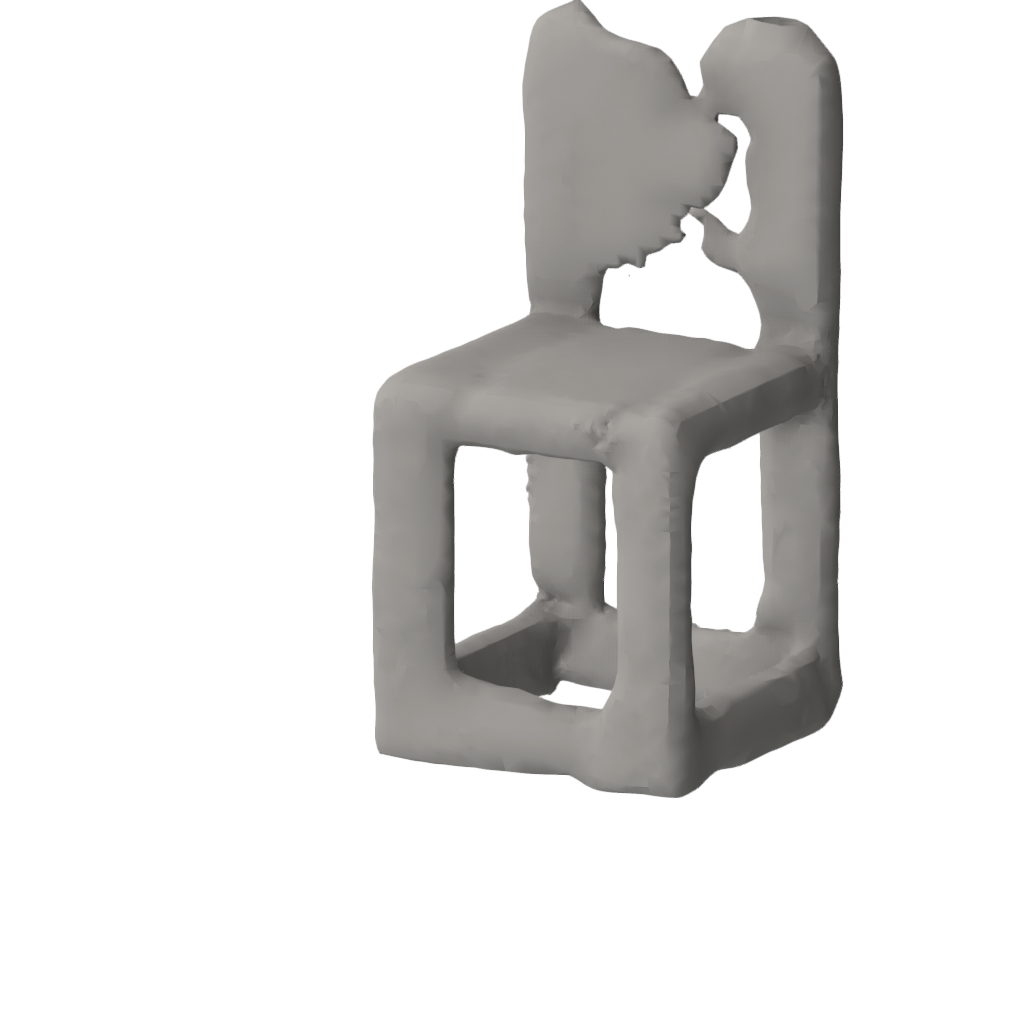}};
\node[] (b) at (19/8,2) {\includegraphics[width=0.14\textwidth]{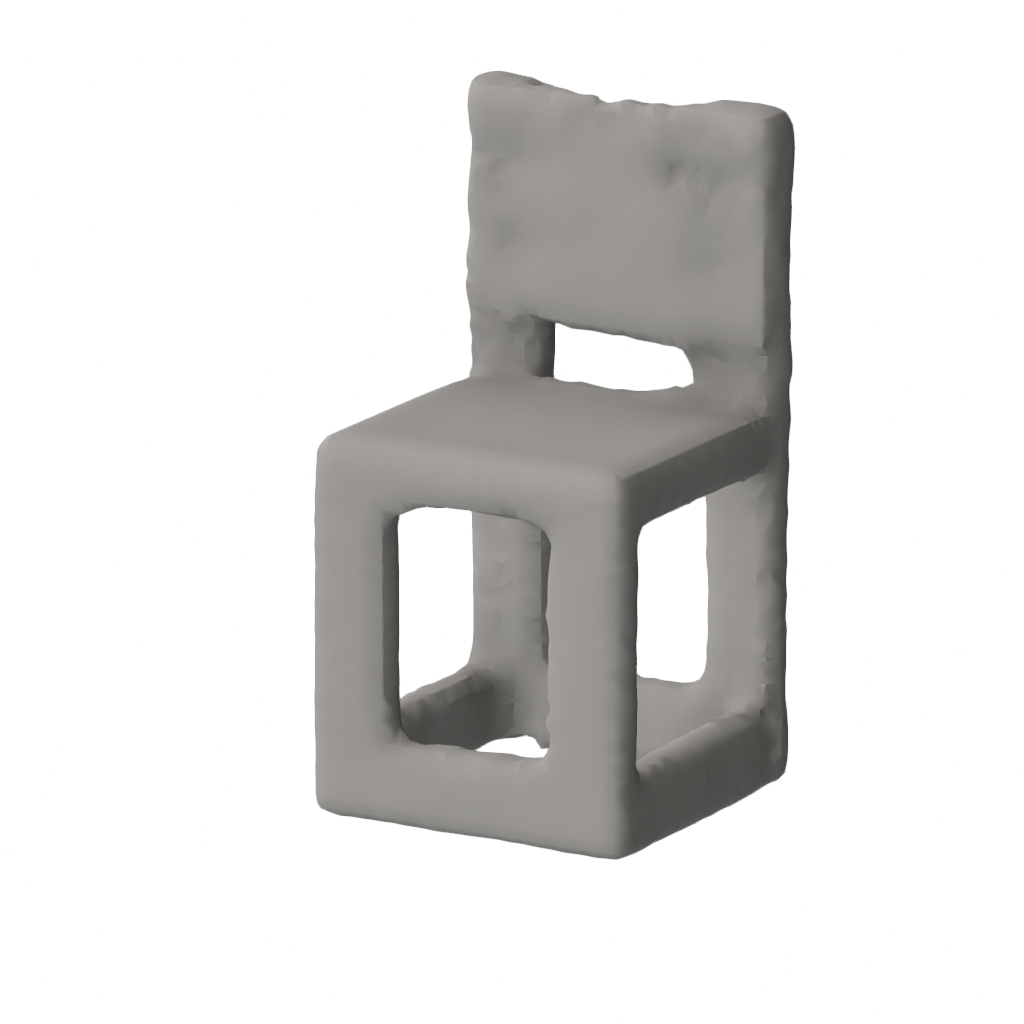}};
\node[] (b) at (19/8,0) {\includegraphics[width=0.14\textwidth]{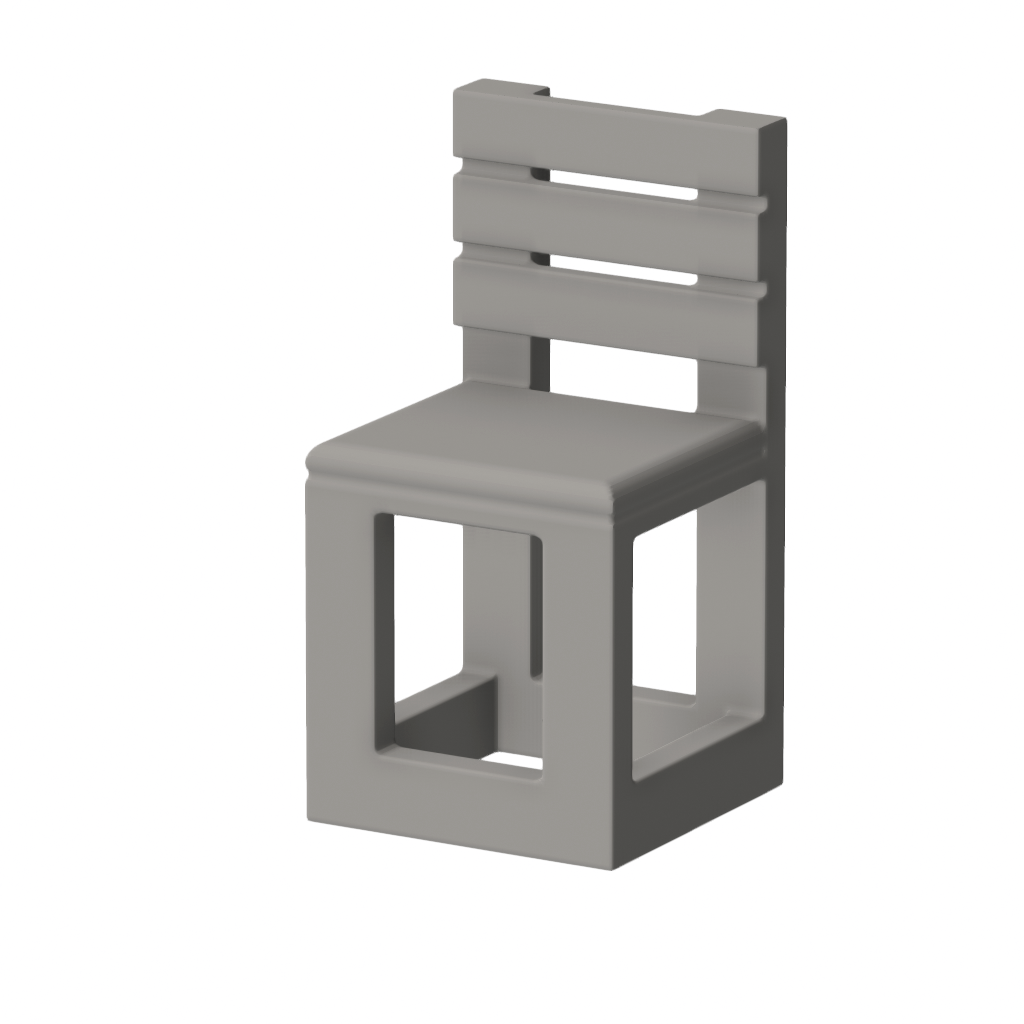}};

\node[] (c) at (19/8*2,6) {\includegraphics[width=0.2\textwidth]{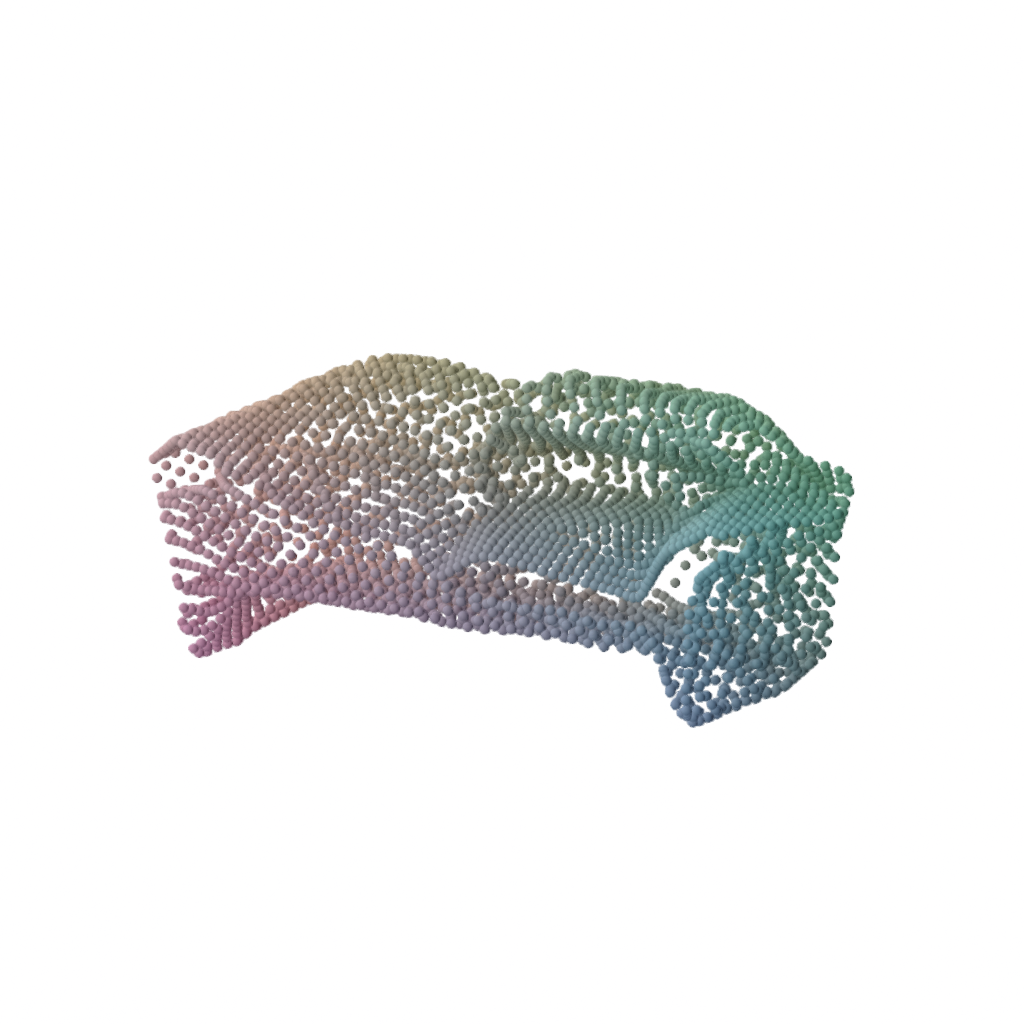}};
\node[] (c) at (19/8*2-0.1,4) {\includegraphics[width=0.2\textwidth]{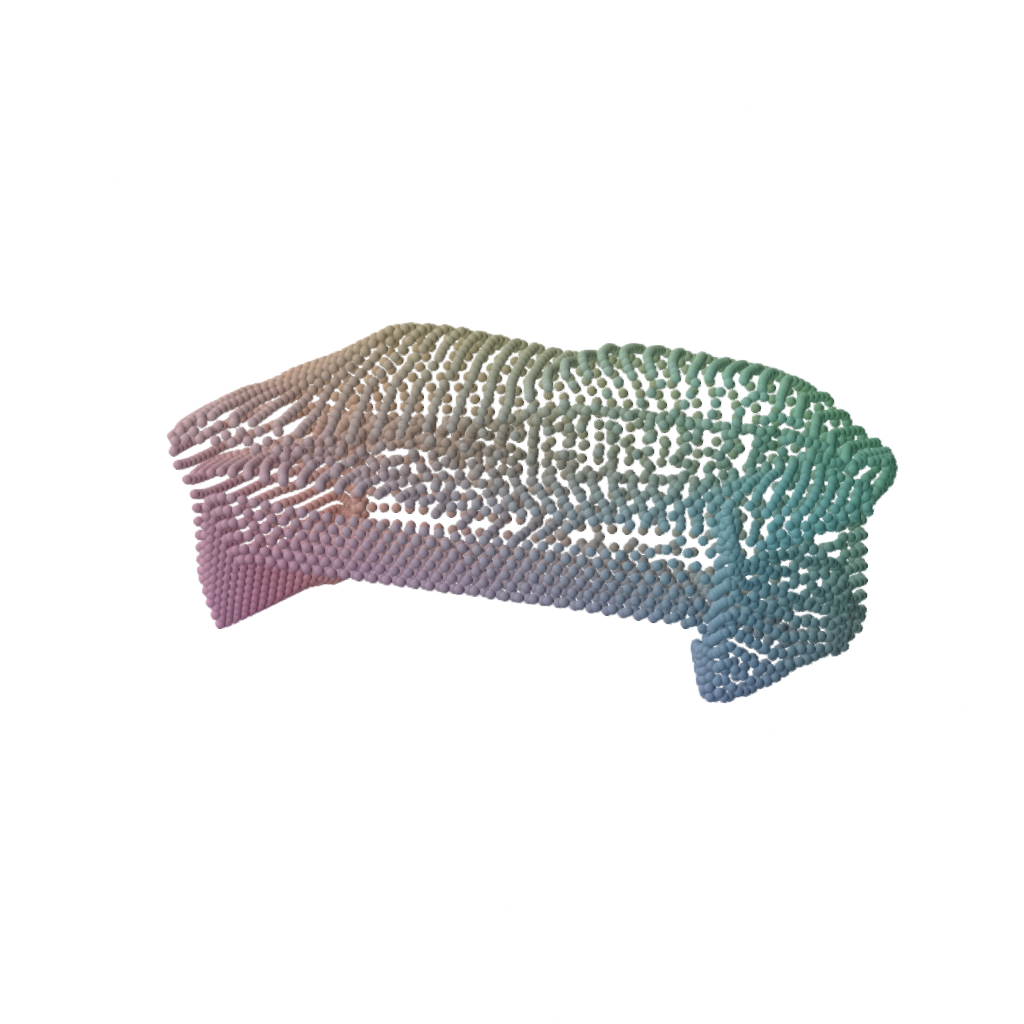}};
\node[] (c) at (19/8*2,2) {\includegraphics[width=0.2\textwidth]{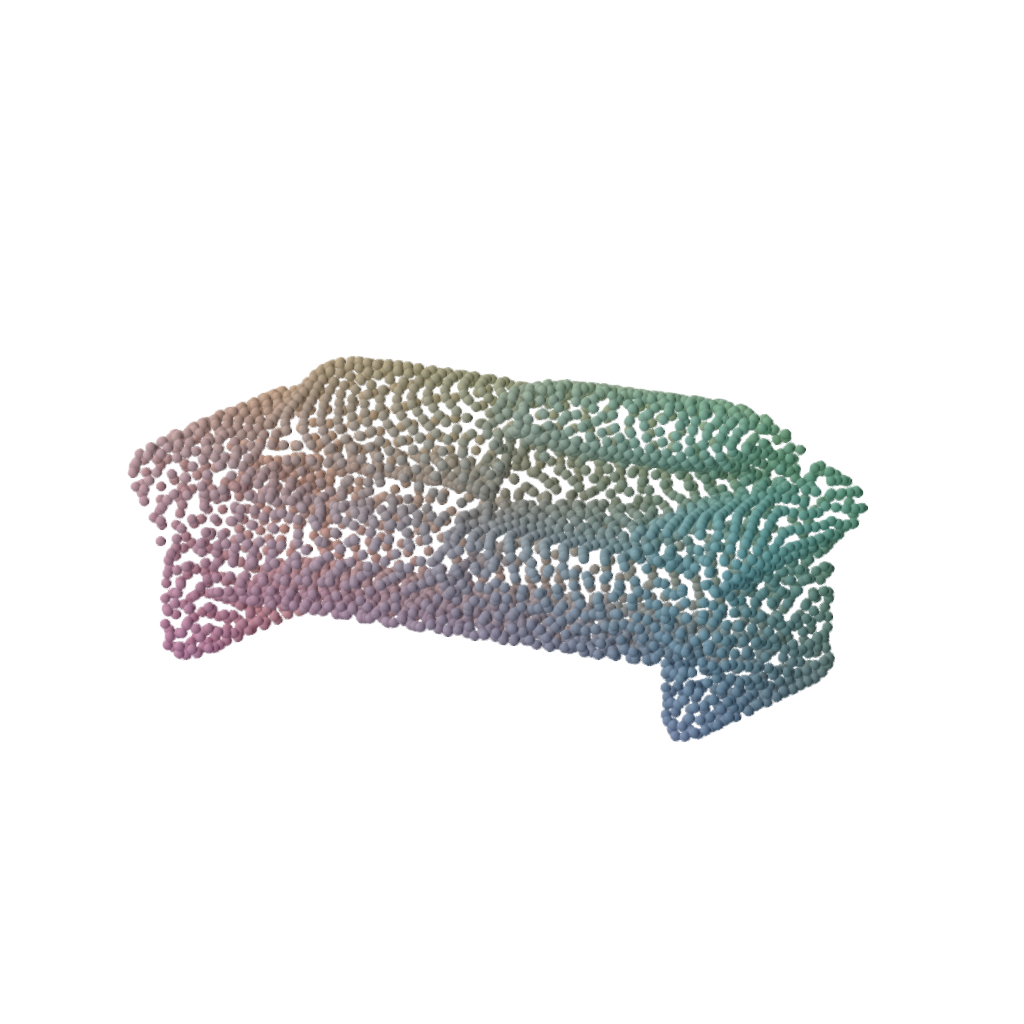} };
\node[] (c) at (19/8*2,0) {\includegraphics[width=0.2\textwidth]{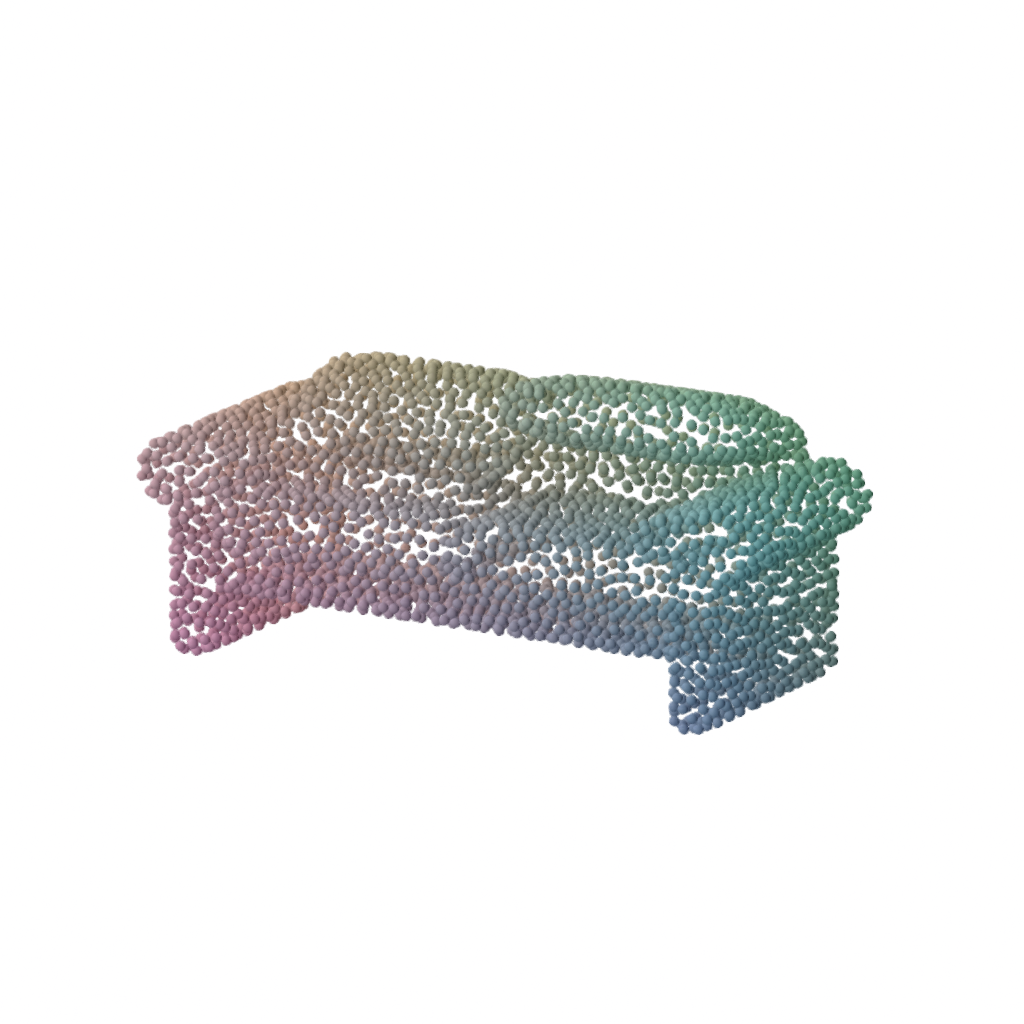} };

\node[] (d) at (19/8*3,6) {\includegraphics[width=0.2\textwidth]{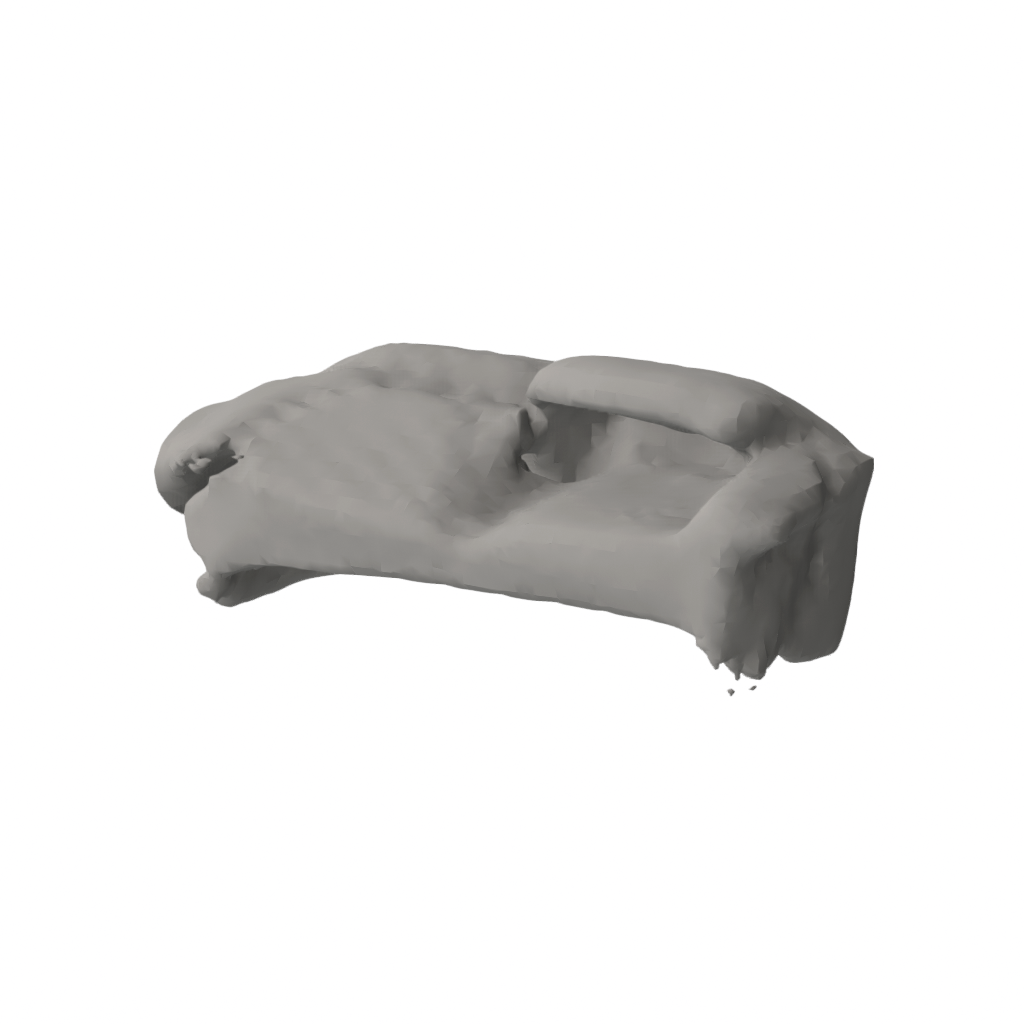}};
\node[] (d) at (19/8*3,4) {\includegraphics[width=0.2\textwidth]{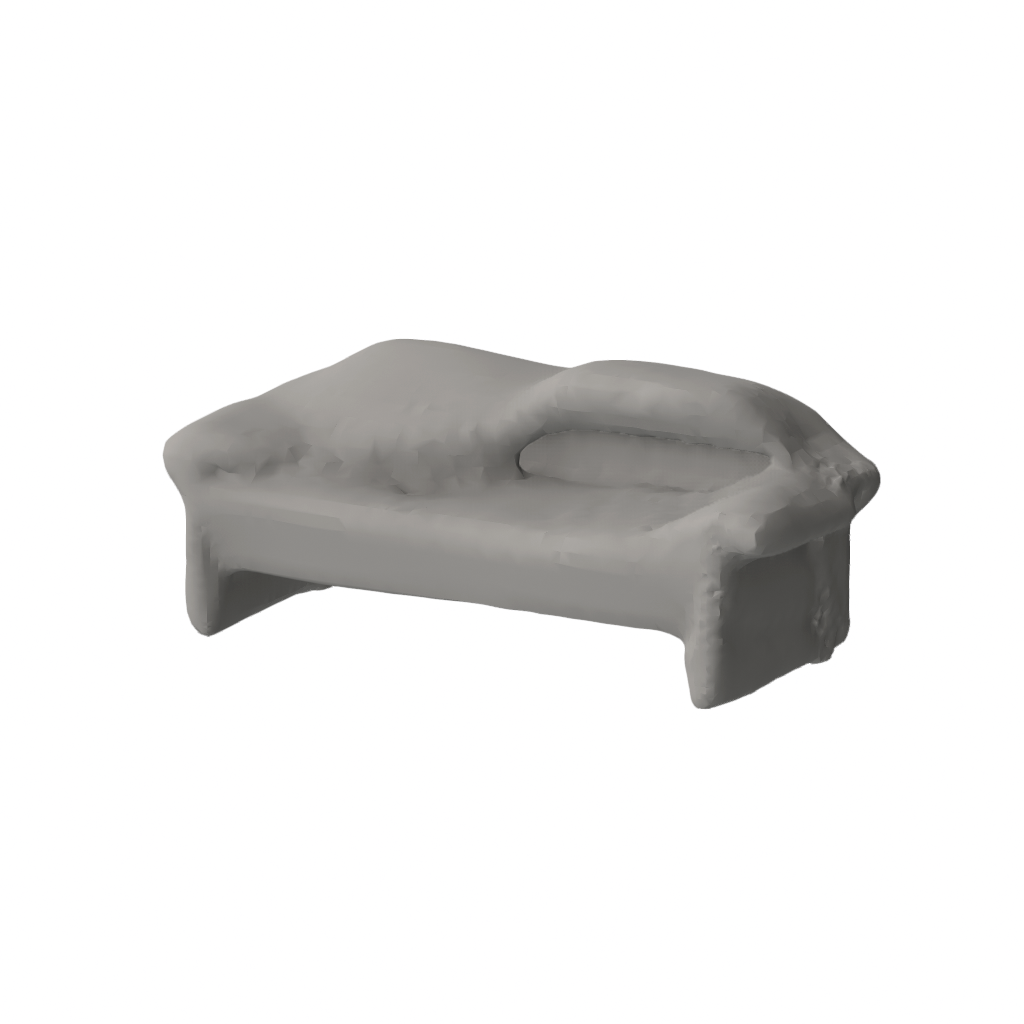}};
\node[] (d) at (19/8*3,2) {\includegraphics[width=0.2\textwidth]{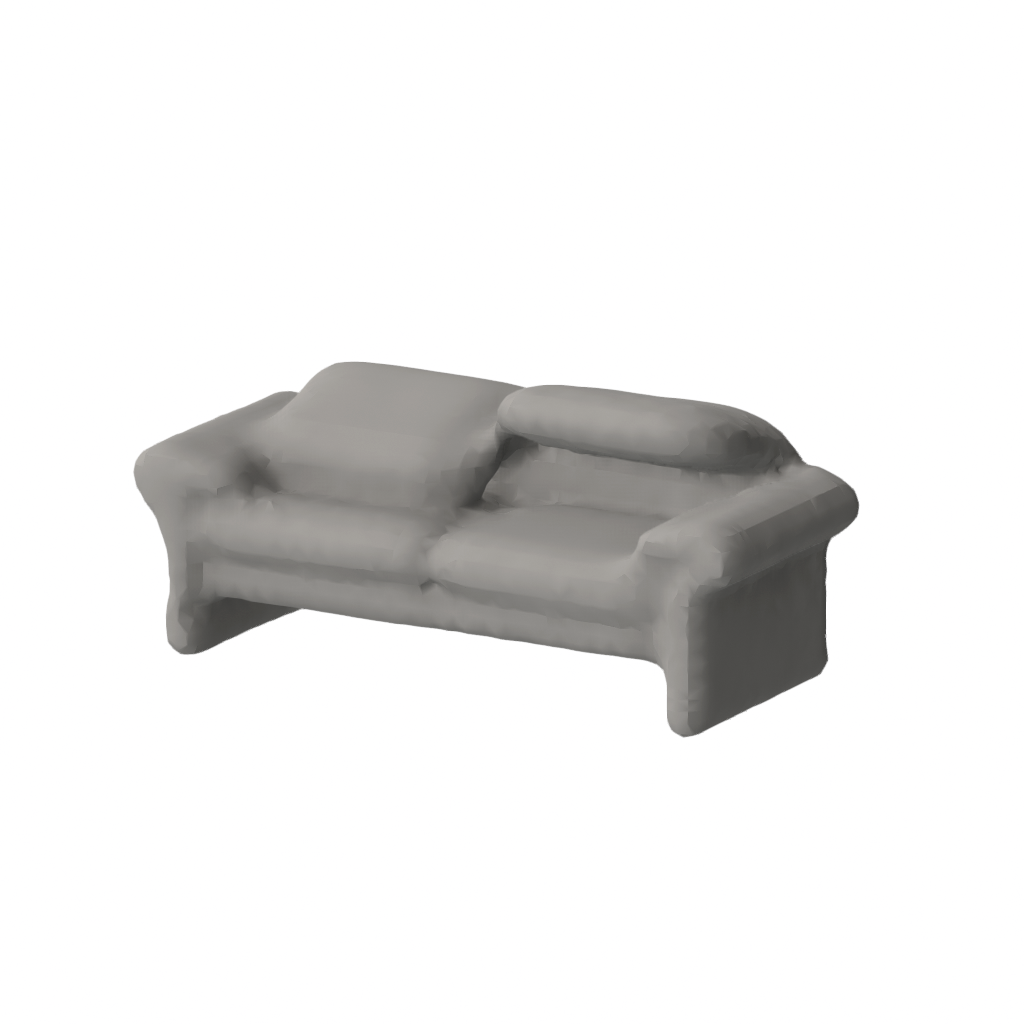} };
\node[] (d) at (19/8*3,0) {\includegraphics[width=0.2\textwidth]{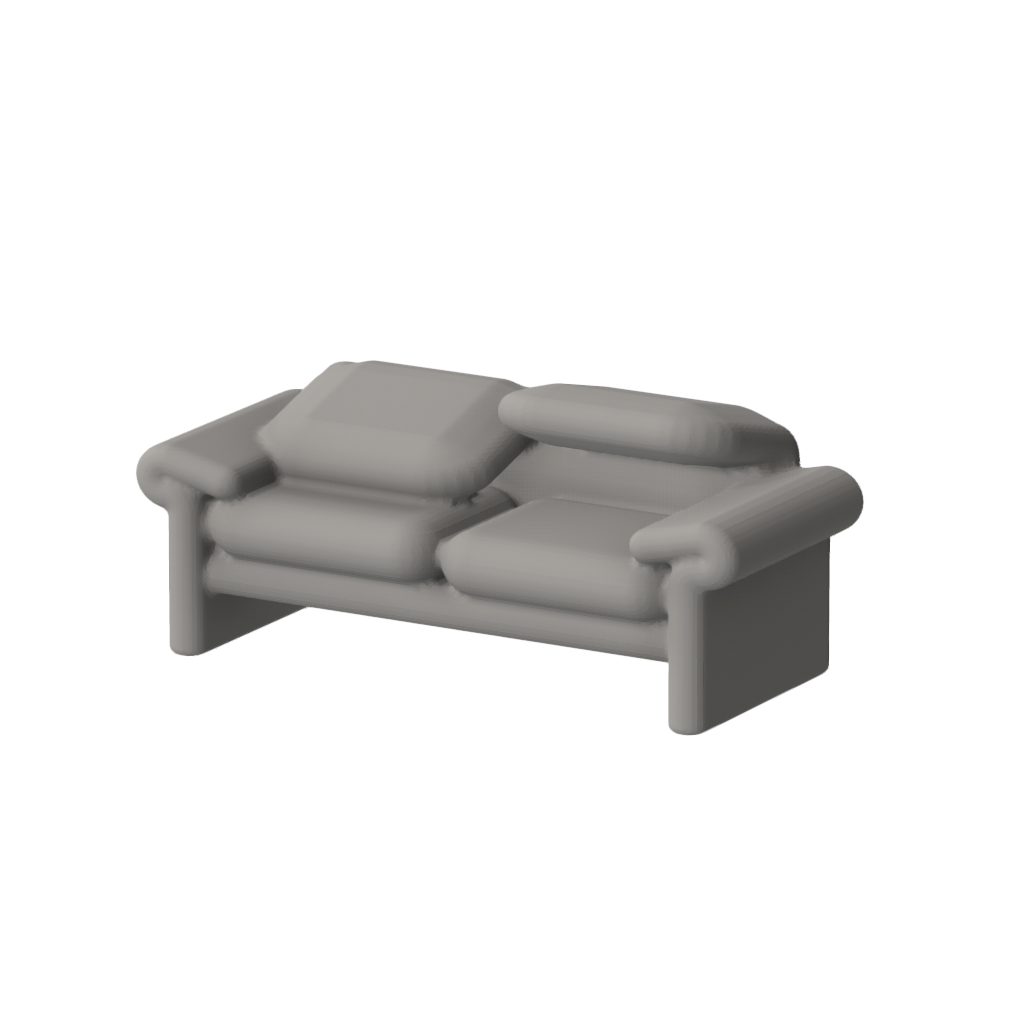} };

\node[] (e) at (19/8*4,6) {\includegraphics[width=0.2\textwidth]{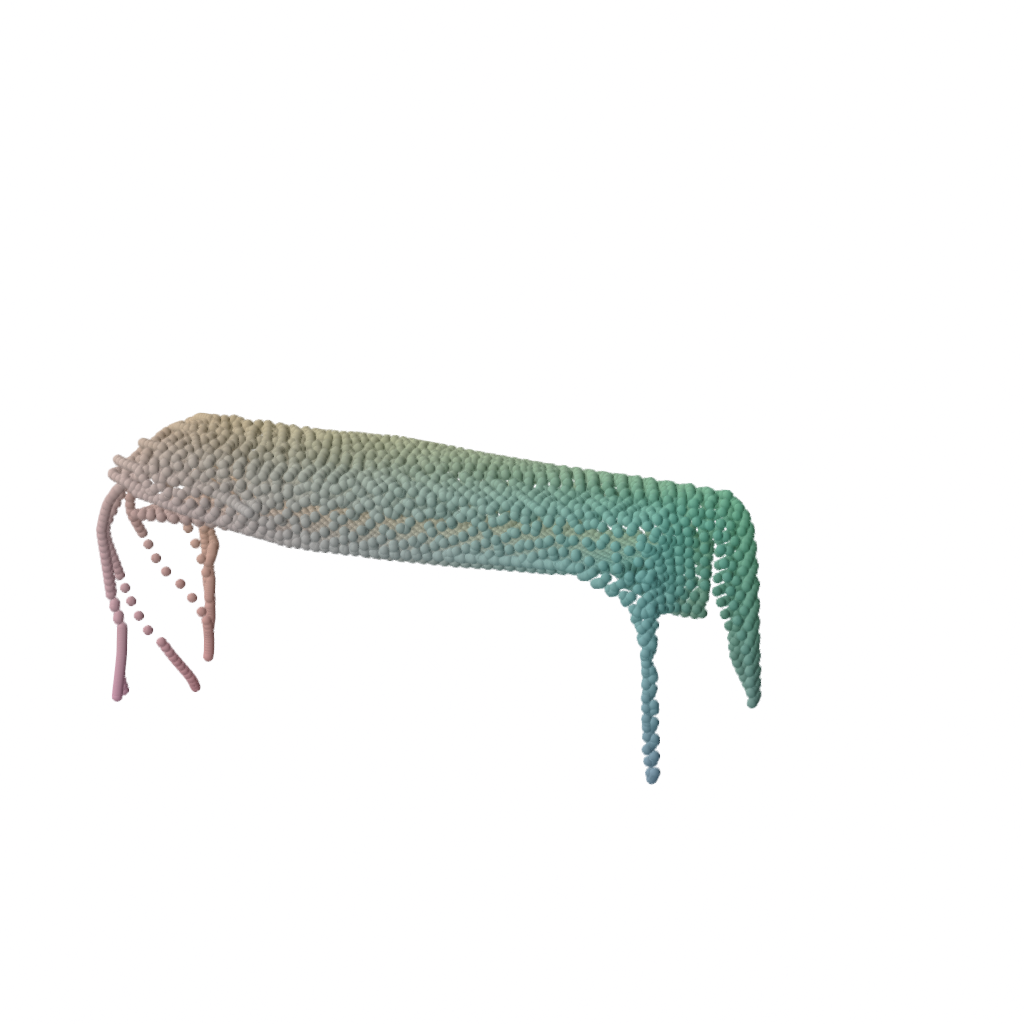}};
\node[] (e) at (19/8*4,4) {\includegraphics[width=0.2\textwidth]{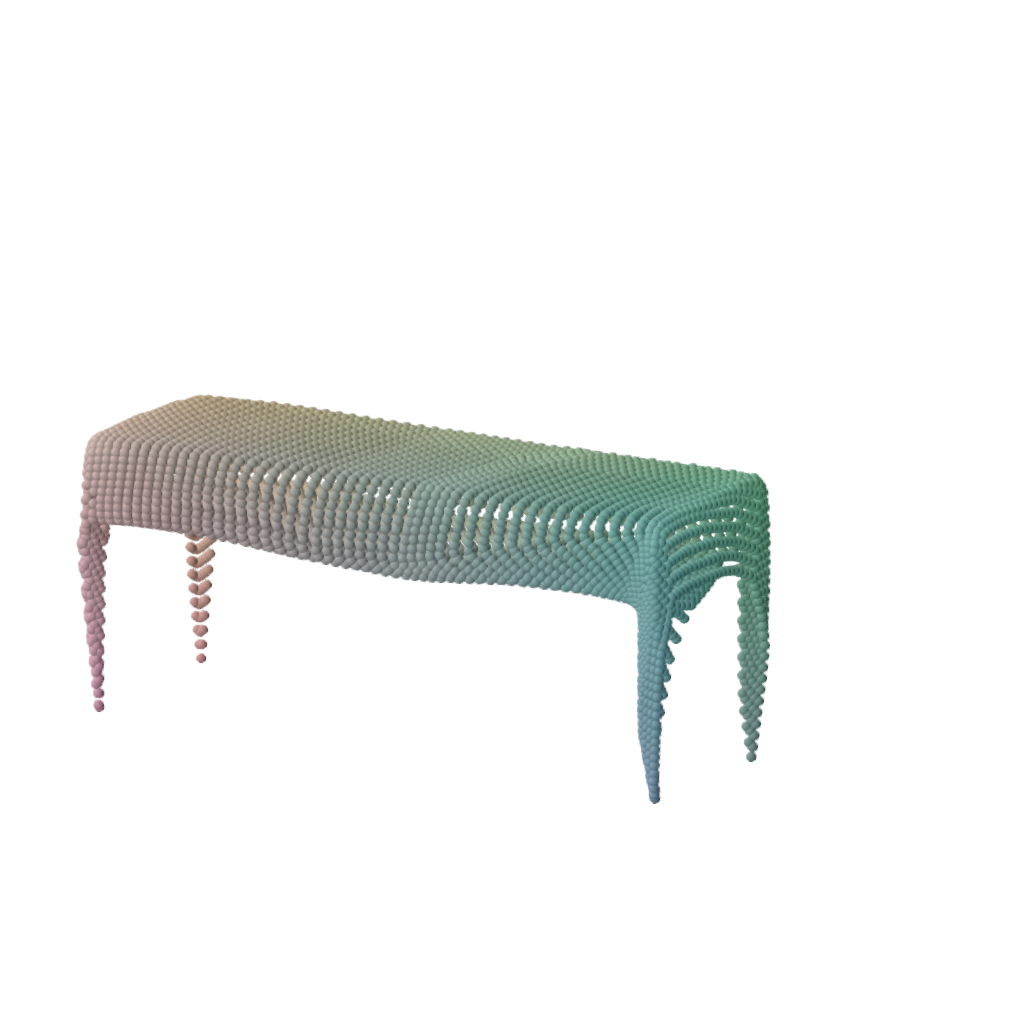}};
\node[] (e) at (19/8*4,2) {\includegraphics[width=0.2\textwidth]{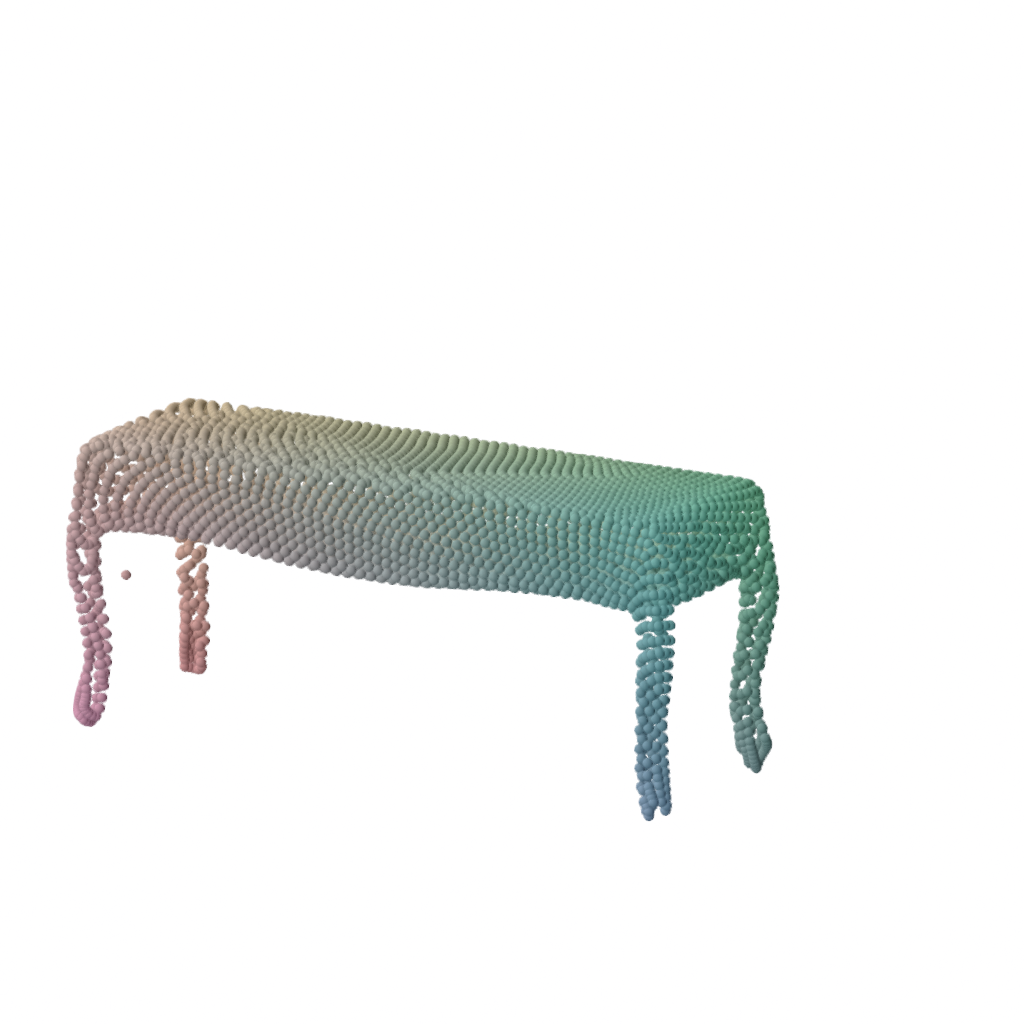} };
\node[] (e) at (19/8*4,0) {\includegraphics[width=0.2\textwidth]{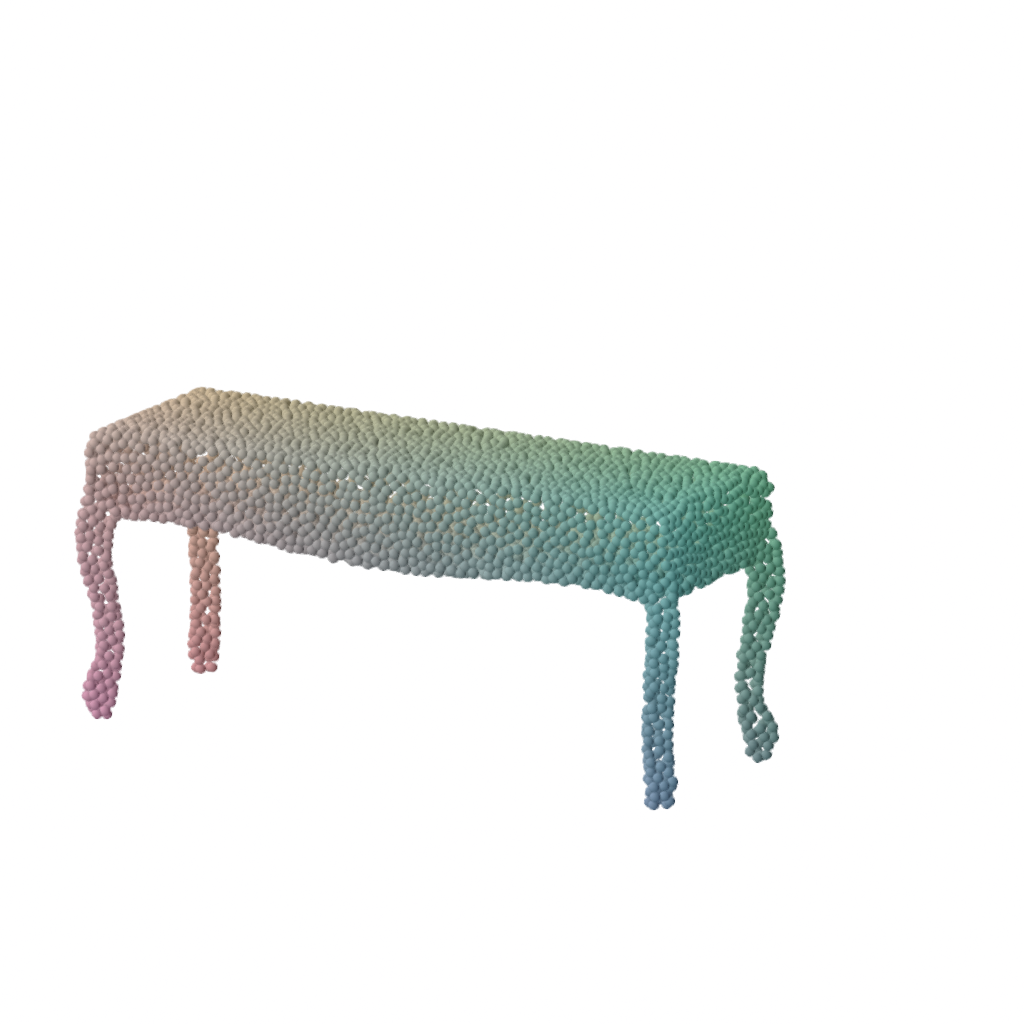} };

\node[] (f) at (19/8*5,6) {\includegraphics[width=0.2\textwidth]{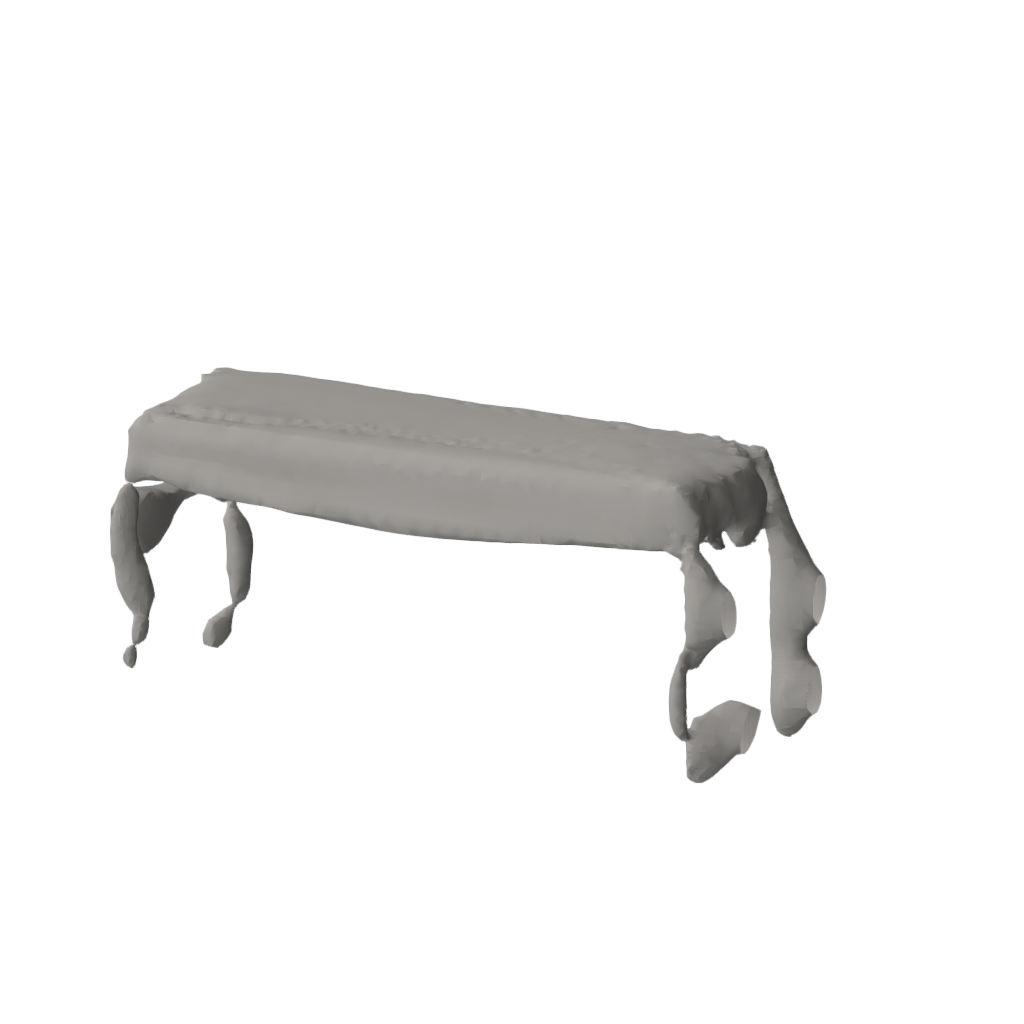}};
\node[] (f) at (19/8*5,4) {\includegraphics[width=0.2\textwidth]{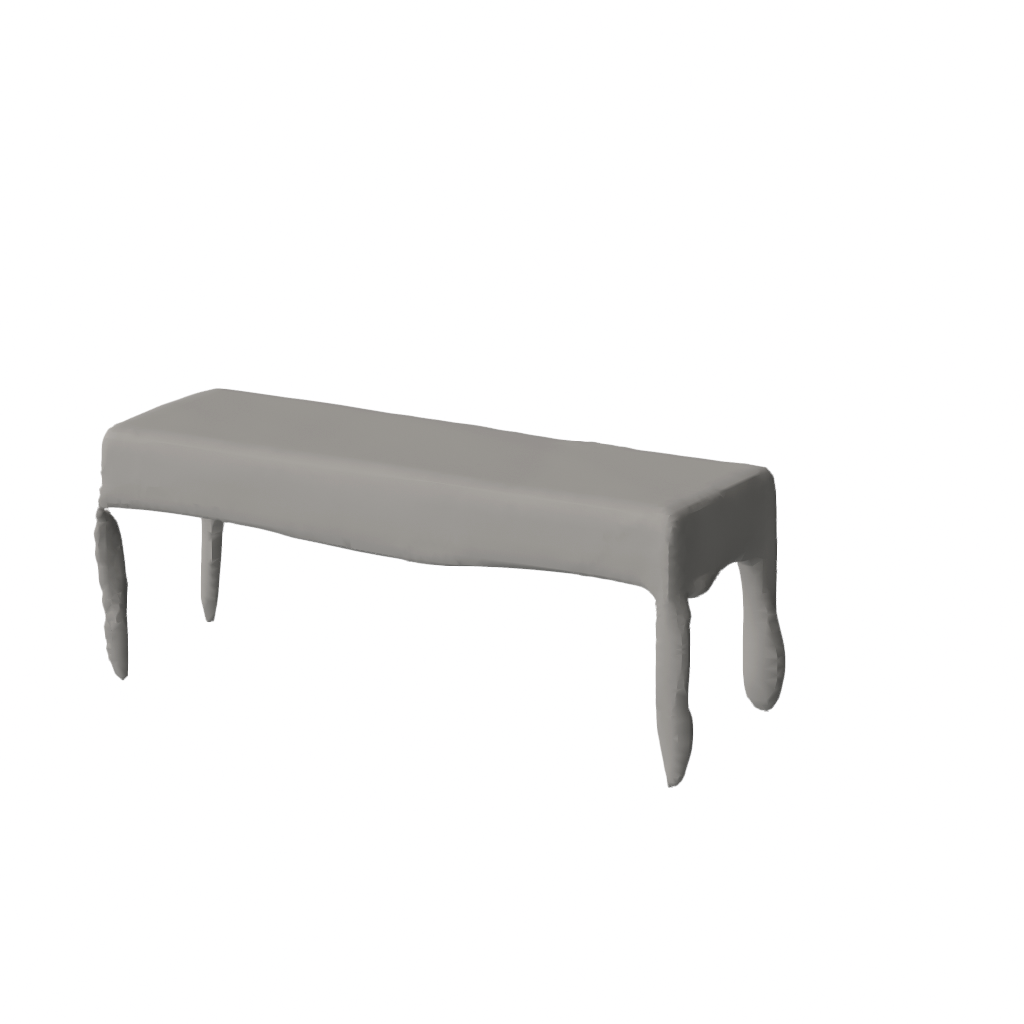}};
\node[] (f) at (19/8*5,2) {\includegraphics[width=0.2\textwidth]{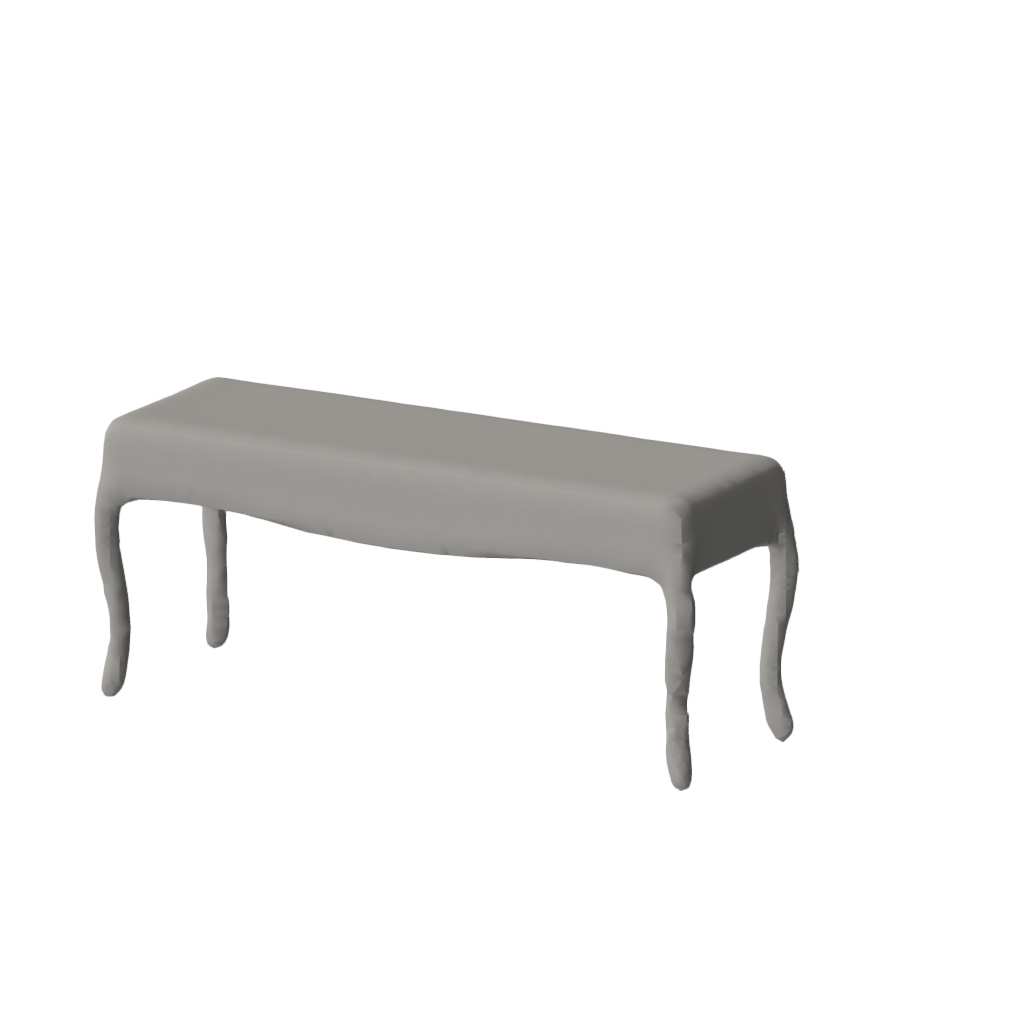} };
\node[] (f) at (19/8*5,0) {\includegraphics[width=0.2\textwidth]{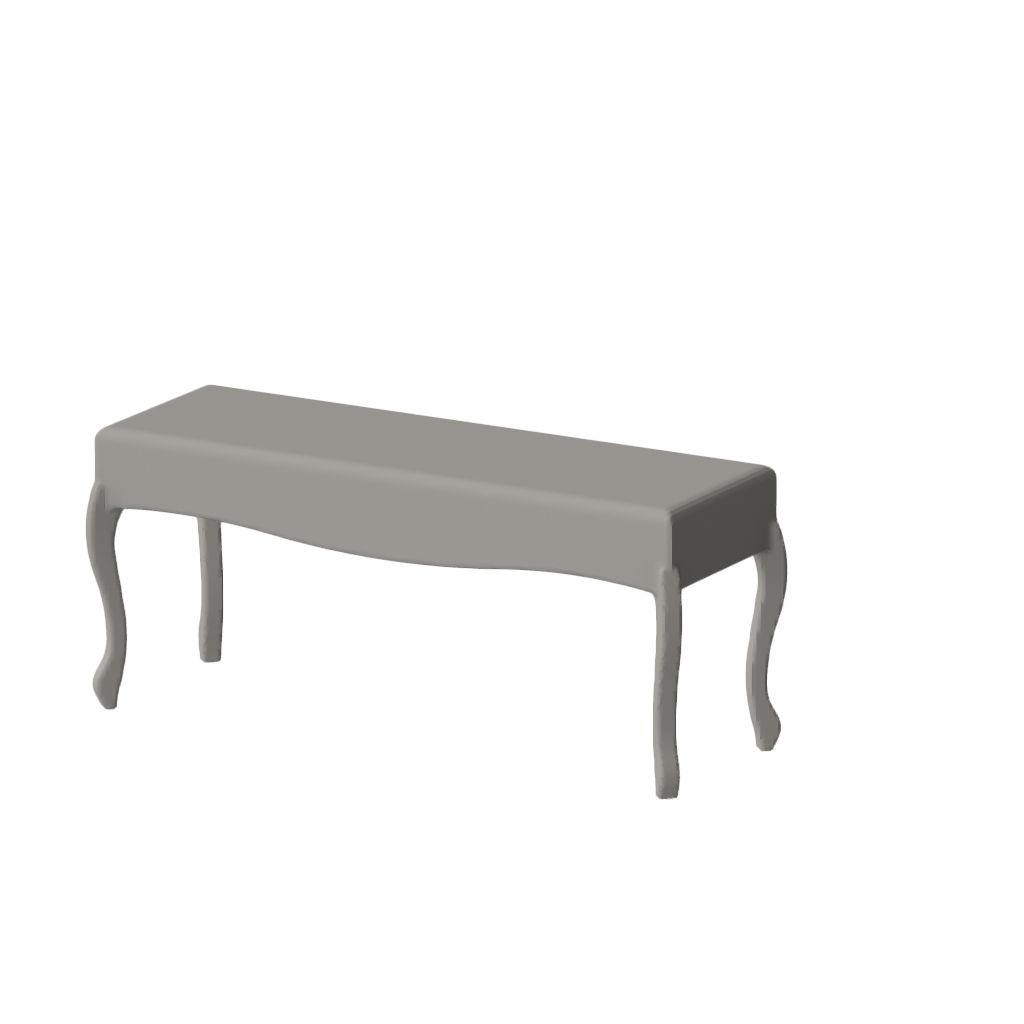} };
\end{tikzpicture}
}
\vspace{-1cm}
\caption{Visual comparisons of reconstructed shapes in the form of point clouds and surfaces under different distance metrics. From top to bottom: EMD, CD, Ours, and GT.} 
\label{OVERFITTING:VIS}
\label{SHAPENET:VIS} 
\end{figure}

\begin{wrapfigure}{r}{7cm} \small
	\centering
        \subfigure[Running time]{\includegraphics[width=0.48\linewidth]{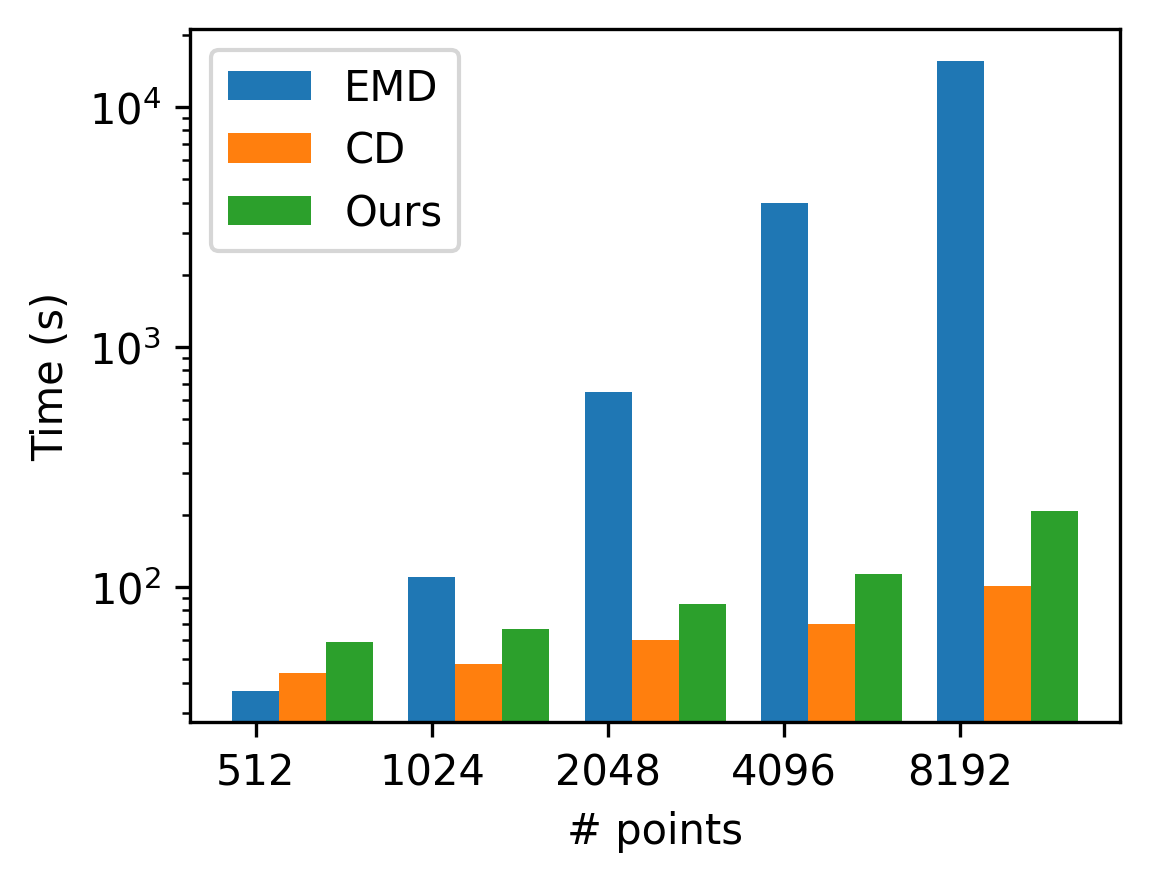}\label{TIME}} 
        \subfigure[GPU Memory]{\includegraphics[width=0.48\linewidth]{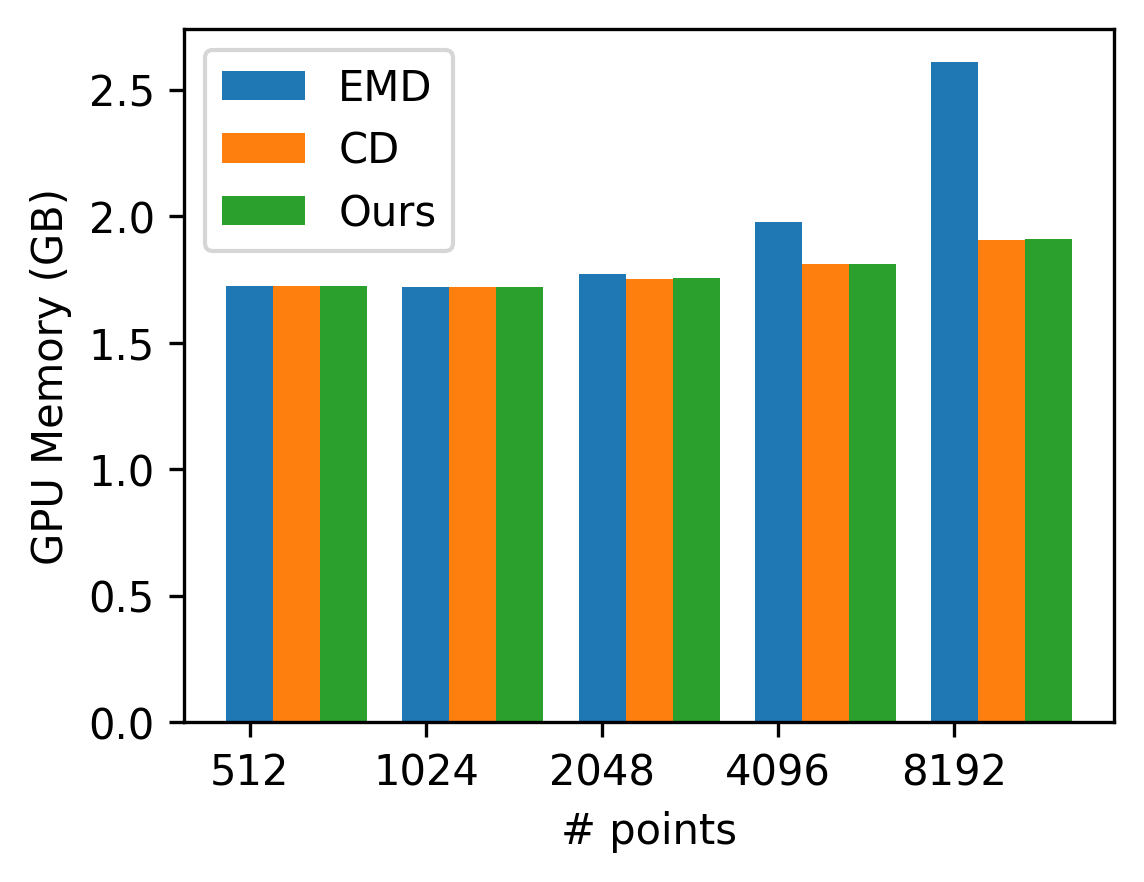}\label{GPU}} 
        \vspace{-0.3cm}
	\caption{Efficiency comparison under 3D shape reconstruction with various numbers of points. 
 }\label{EFFICIENCY}
\vspace{-0.2cm}
\end{wrapfigure}
\paragraph{Comparisons.} 
To quantitatively compare 
different distance metrics, we employed 
CD, Hausdorff distance (HD), and the point-to-surface distance (P2F) to evaluate the accuracy of reconstructed point 
clouds.
As for the reconstructed triangle meshes, we used Normal Consistency (NC) and F-Score 
with thresholds of 0.5\% and 1\%, denoted as F-0.5\% and F-1\%, as the evaluation metrics.
Table \ref{OVERFITTING} and Fig. \ref{OVERFITTING:VIS} show the numerical and visual results, respectively, where both quantitative accuracy and visual quality of reconstructed shapes by the network trained with our CLGD are much better. 
Besides, due to the difficulty in establishing the optimal bijection between a relatively large number of points, i.e., 4096 points in our experiments, the network trained with EMD produces even worse shapes than that with CD. 


\paragraph{Efficiency Analysis.} We compared the running time 
in Fig. \ref{TIME}, where it can be seen that with the number of points increasing, the reconstruction driven by EMD requires much more time to optimize than that by CD and our CLGD. 
Fig. \ref{GPU} compares the GPU memory consumption, showing that these three distance metrics are comparable when the number of points is relatively small, 
but when dealing with a large number of points, EMD requires more GPU memory. 

\subsection{Rigid Registration} \label{RIGID:REGISTRATION}
Given source and target point clouds, denoted as $\mathbf{P}_{\rm src}\in\mathbb{R}^{N_{\rm src}\times 3}$ and $\mathbf{P}_{\rm tgt}\in\mathbb{R}^{N_{\rm tgt}\times 3}$, respectively, rigid registration is to seek a spatial transformation $[\mathbf{R},\mathbf{t}]$ to align $\mathbf{P}_{\rm src}$ with $\mathbf{P}_{\rm tgt}$, where $\mathbf{R}\in\mathbb{R}^{3\times 3}$ is the rotation matrix and $\mathbf{t}\in\mathbb{R}^3$ is the translation vector. 
The optimization-based registration  directly solves the following problem: 
\begin{equation}
\{\hat{\mathbf{R}},\hat{\mathbf{t}}\}=\mathop{\arg\min}_{\mathbf{R},\mathbf{t}}\mathcal{D}\left(\mathbf{RP}_{\rm src}+\mathbf{t},\mathbf{P}_{\rm tgt}\right),\label{REG:EQ} 
\end{equation}
where $\mathcal{D}(\cdot,~\cdot)$ is the distance metric that can be EMD, CD, ARL, or our CLGD. 
We also consider unsupervised learning-based rigid registration.  Specifically, following \cite{ARL}, we modify RPM-Net \cite{RPMNET}, a supervised learning-based registration method, to be unsupervised by using a distance metric to drive the learning of the network. See the \textit{Supplementary Material} for more details.
We also selected two common rigid registration methods as baselines, i.e., ICP \cite{ICP} and FGR \cite{FGR}.


\begin{wraptable}{r}{8.3cm}
\small
\centering
\vspace{-0.6cm}
\caption{Quantitative comparisons of rigid registration on the Human dataset \cite{ARL}.}\label{REGISTRATION:TABLE}
\begin{tabular}{l|l|c c}
\toprule[1.2pt]
~ & Method &  ${\rm RE}$ ($^\circ$) $\downarrow$ & TE (m) $\downarrow$    \\
\hline
\multirow{6}{*}{Optimization-based}     &  ICP \cite{ICP} & 1.276 & 0.068  \\
                                        & FGR \cite{FGR} & 19.684  & 0.384  \\ 
                                        \cline{2-4}
                                        & EMD \cite{EMD} & 7.237 & 0.642  \\
                                         & CD \cite{CD} & 10.692 & 0.421  \\
                                        & ARL \cite{ARL} & 1.245 & 0.085  \\
                                        & Ours & \textbf{1.040}  & \textbf{0.040}  \\
\hline 
\multirow{4}{*}{\makecell[l]{Unsupervised learning \\ (RPM-Net \cite{RPMNET})}}   &  EMD \cite{EMD}   & 7.667 & 0.638  \\
                                    &  CD \cite{CD}  & 2.197 & 0.287  \\
                                    &  ARL \cite{ARL} & 1.090  &  0.075  \\
                                    & Ours  & \textbf{0.793} & \textbf{0.053}  \\
                                   
\bottomrule[1.2pt]
\end{tabular}
\vspace{-0.2cm}
\end{wraptable}

\paragraph{Implementation Details.}
We used the Human dataset provided in \cite{ARL}, containing 5000 pairs of source and target point clouds for training and 500 pairs for testing. The source point clouds are 1024 points, while the target ones are 2048 points, i.e., they are partially overlapped. 
We used Open3D \cite{OPEN3D} to implement ICP and FGR. For the optimization-based methods, we utilized Lie algebraic to represent the transformation and optimized it with the ADAM optimizer for 1000 iterations with a learning rate of 0.02. For the unsupervised learning-based methods, we kept the same training settings as \cite{ARL}. 
We refer the reader to \textit{Supplemental Material} for more details.

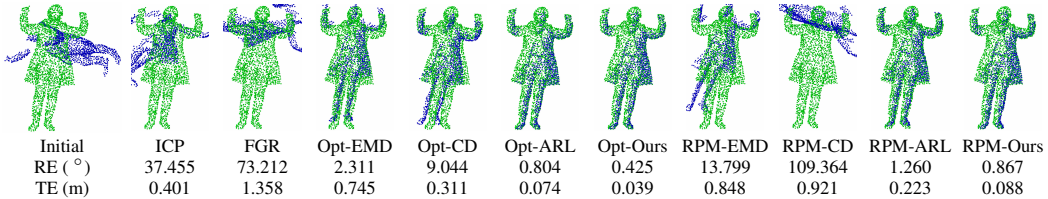
\begin{figure}[h] \small 
\centering
{
\begin{tikzpicture}[]
\node[] (a) at (0-0.2,1.35) {\includegraphics[width=0.12\textwidth]{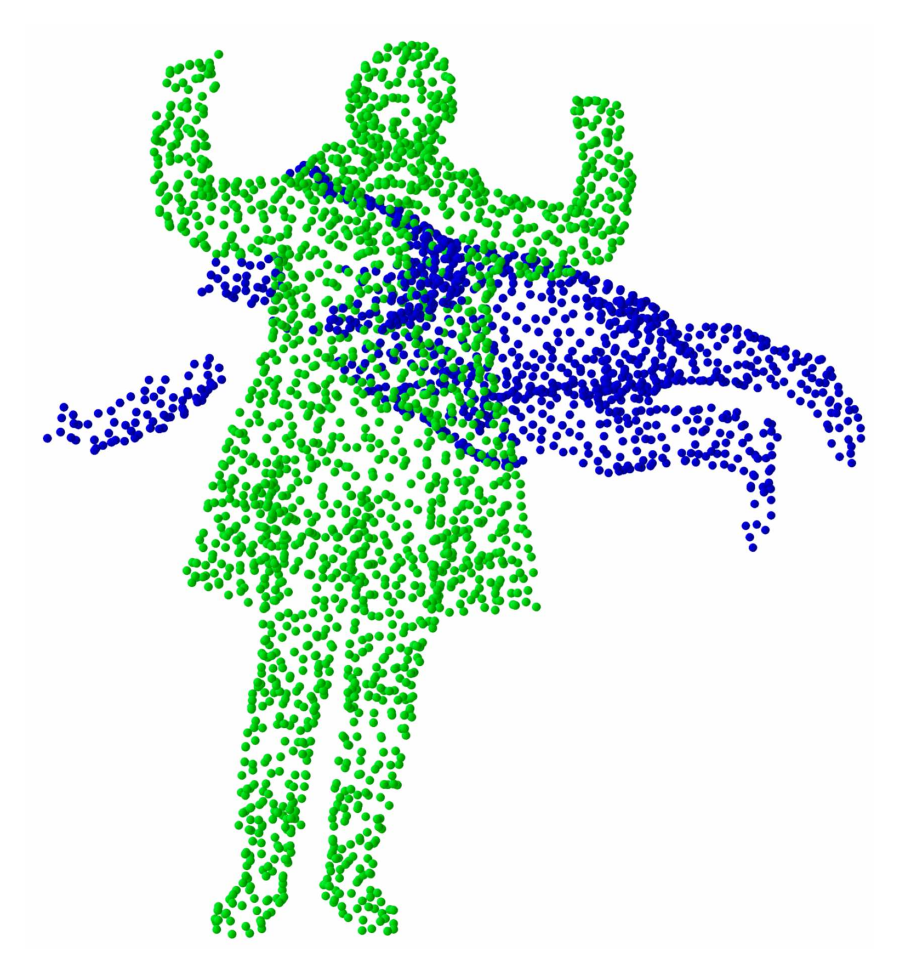} };
\node[] (a) at (0-0.2,0) {\scriptsize \makecell[c]{ Initial\\  RE ($\ ^\circ$)\\  TE (m)}};

\node[] (b) at (12.3/10,1.35) {\includegraphics[width=0.08\textwidth]{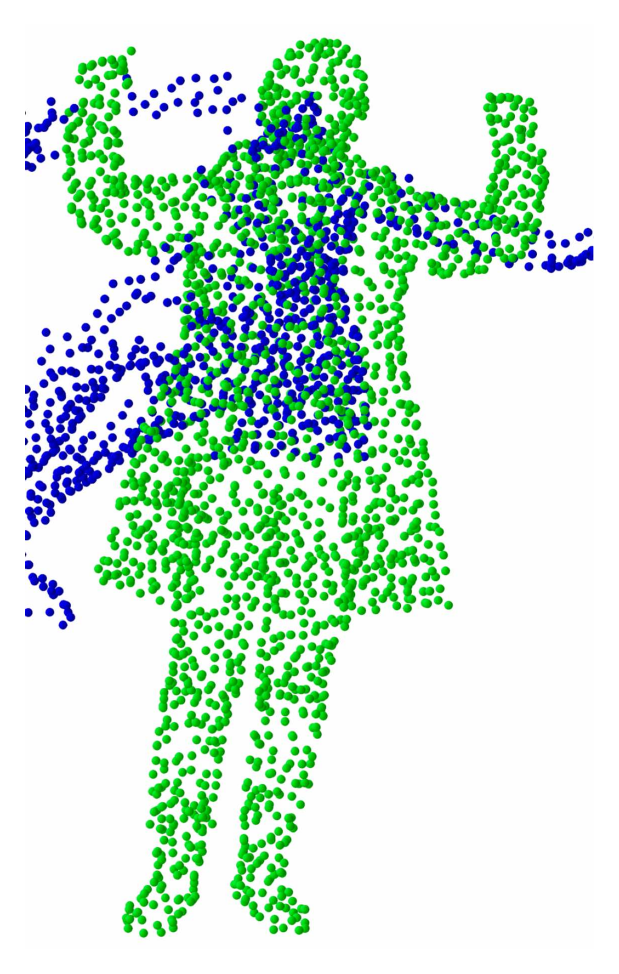}};
\node[] (b) at (12.3/10,0) {\scriptsize\makecell[c]{ ICP\\37.455\\0.401}};

\node[] (b) at (12.3/10*2,1.35) {\includegraphics[width=0.08\textwidth]{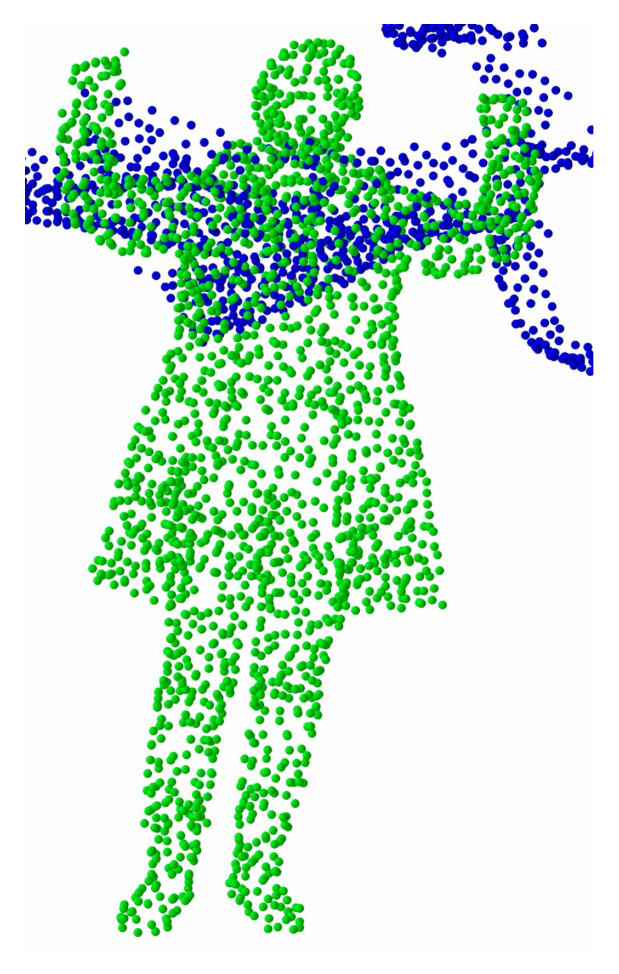}};
\node[] (b) at (12.3/10*2,0) {\scriptsize\makecell[c]{ FGR\\73.212\\1.358}};

\node[] (c) at (12.3/10*3,1.35) {\includegraphics[width=0.08\textwidth]{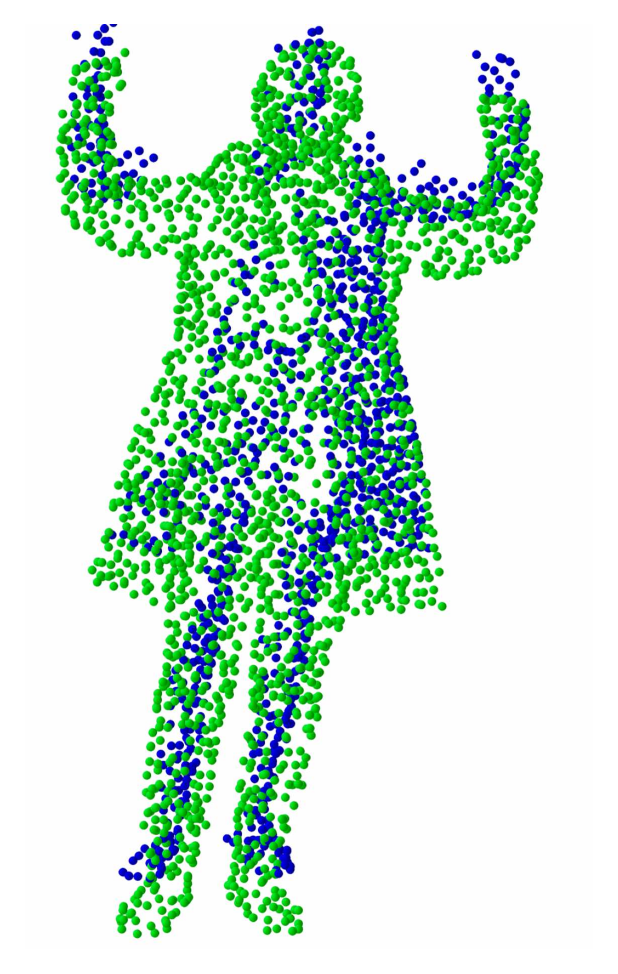}};
\node[] (b) at (12.3/10*3,0) {\scriptsize\makecell[c]{ Opt-EMD\\2.311\\0.745}};

\node[] (c) at (12.3/10*4,1.35) {\includegraphics[width=0.08\textwidth]{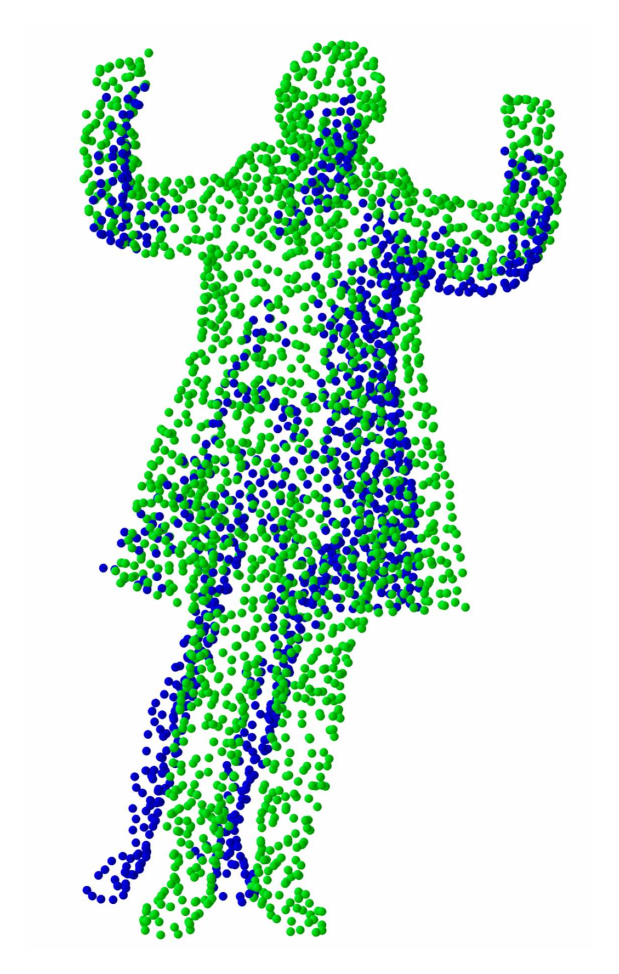} };
\node[] (b) at (12.3/10*4,0) {\scriptsize\makecell[c]{ Opt-CD\\9.044\\0.311}};

\node[] (d) at (12.3/10*5,1.35) {\includegraphics[width=0.08\textwidth]{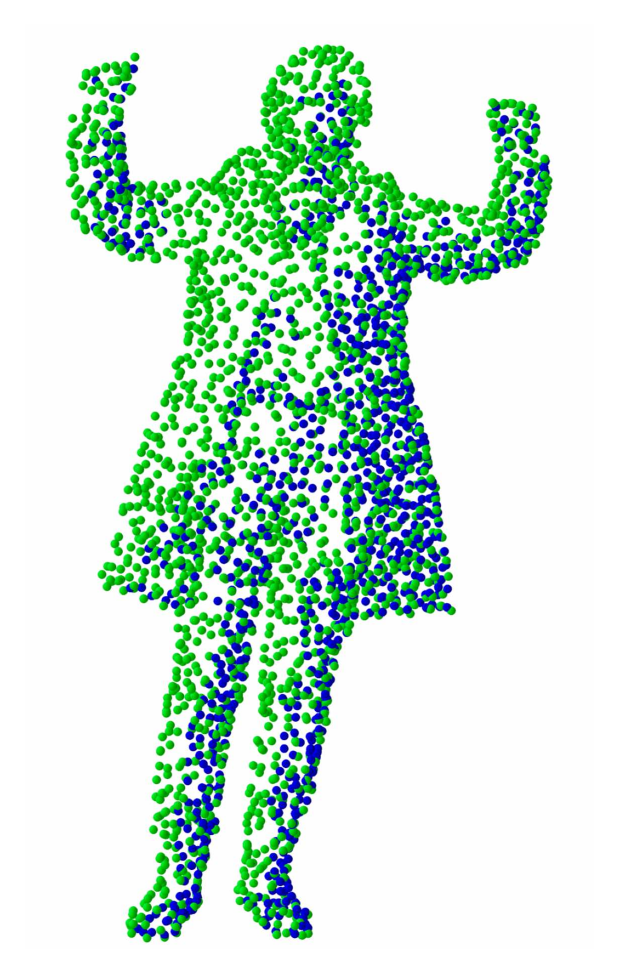}};
\node[] (d) at (12.3/10*5,0) {\scriptsize\makecell[c]{Opt-ARL\\0.804\\0.074}};

\node[] (d) at (12.3/10*6,1.35) {\includegraphics[width=0.08\textwidth]{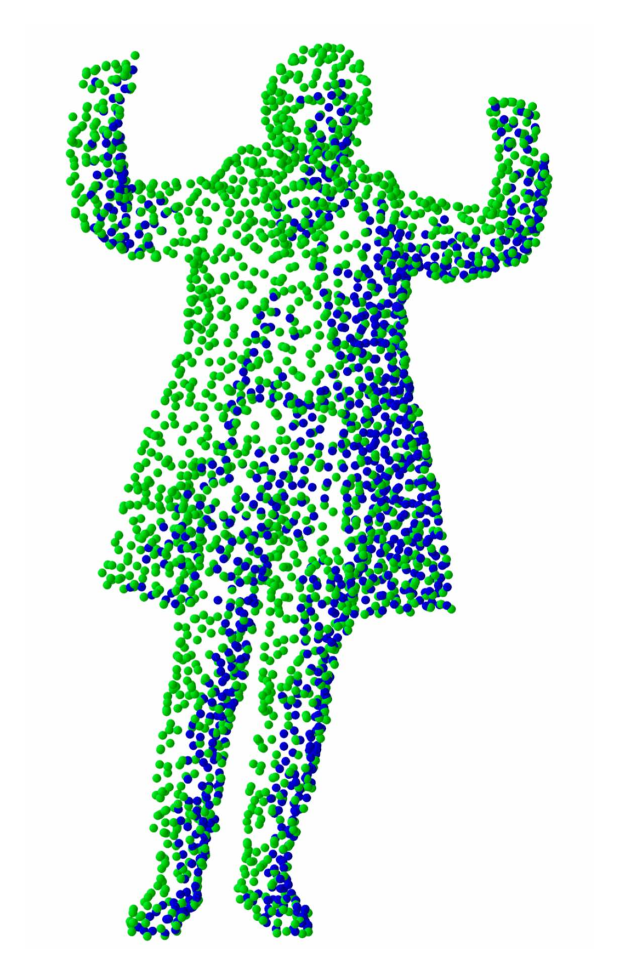} };
\node[] (d) at (12.3/10*6,0) {\scriptsize\makecell[c]{Opt-Ours\\0.425\\0.039}};

\node[] (e) at (12.3/10*7,1.35) {\includegraphics[width=0.08\textwidth]{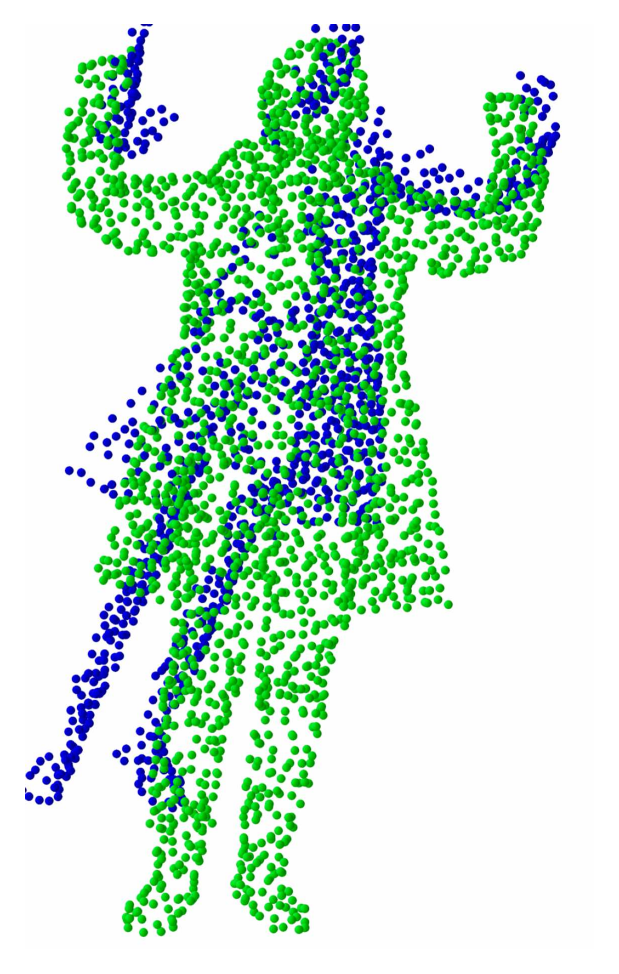}};
\node[] (d) at (12.3/10*7,0) {\scriptsize\makecell[c]{RPM-EMD\\13.799\\0.848}};

\node[] (e) at (12.3/10*8,1.35) {\includegraphics[width=0.08\textwidth]{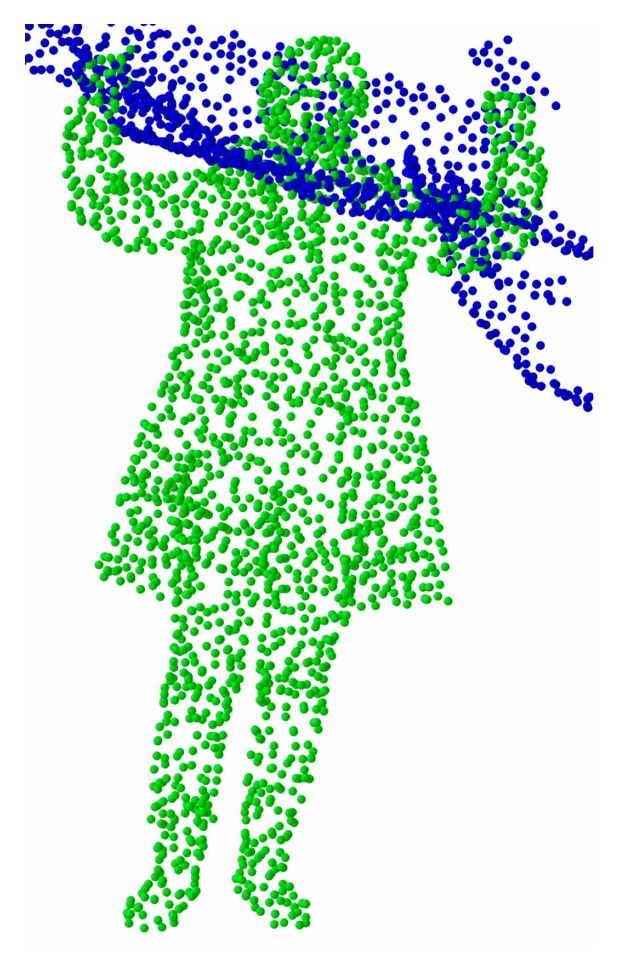} };
\node[] (d) at (12.3/10*8,0) {\scriptsize\makecell[c]{RPM-CD\\109.364\\0.921}};

\node[] (f) at (12.3/10*9,1.35) {\includegraphics[width=0.08\textwidth]{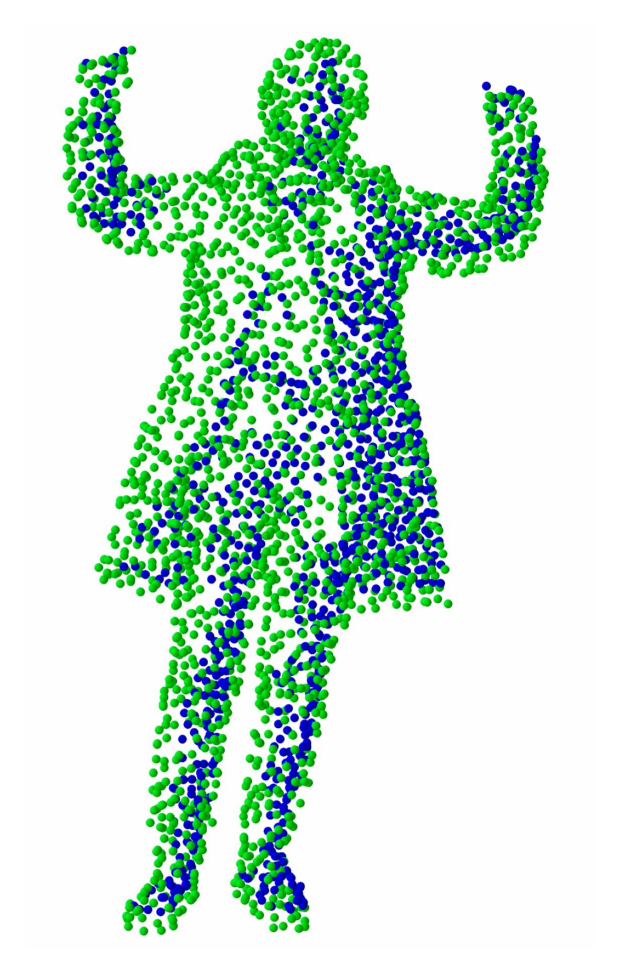}};
\node[] (d) at (12.3/10*9,0) {\scriptsize\makecell[c]{RPM-ARL\\1.260\\0.223}};

\node[] (f) at (12.3,1.35) {\includegraphics[width=0.08\textwidth]{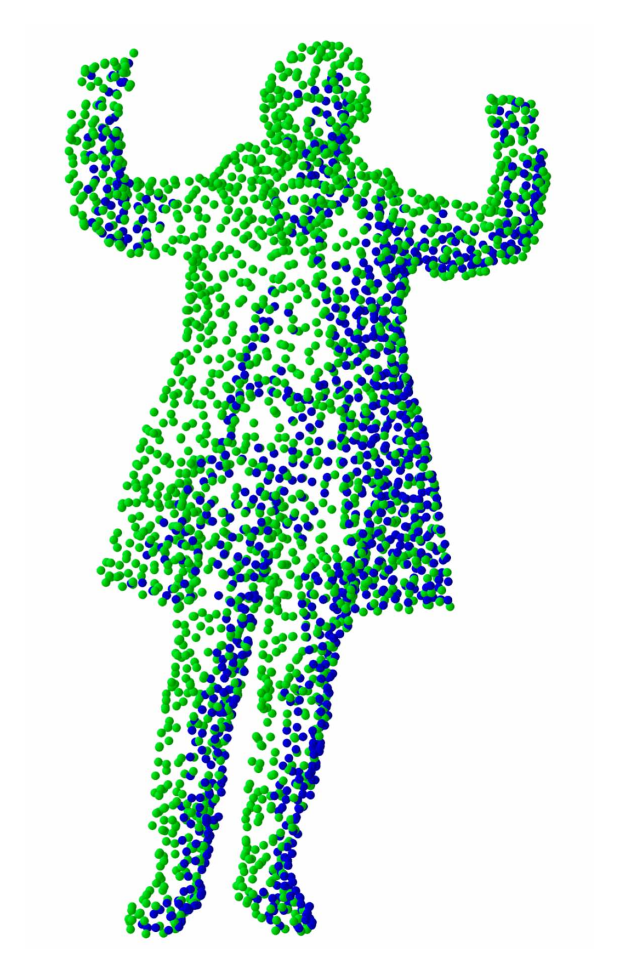} };
\node[] (d) at (12.3,0) {\scriptsize\makecell[c]{RPM-Ours\\0.867\\0.088}};

\end{tikzpicture}
}
\vspace{-0.6cm}
\caption{Visual comparisons of rigid registration results. The \textcolor{blue}{blue} and \textcolor{green}{green} points represent the source and target point clouds, respectively. \color{cyan}{\faSearch~} Zoom in to see details.} \label{REGISTRATION:FIG}
\label{SHAPENET:VIS} 
\vspace{-0.3cm}
\end{figure}

\paragraph{Quantitative Comparison.} 
Following previous registration work \cite{METRIC}, we adopted \textit{Rotation Error (RE)} and \textit{Translation Error (TE)} as the evaluation metrics. As shown in Table \ref{REGISTRATION:TABLE} and Fig. \ref{REGISTRATION:FIG}, our distance metric outperforms the baseline methods in both optimization-based and unsupervised learning methods. ICP, FGR, as long as EMD and CD, can easily get local optimal solutions since they struggle to properly handle the outliers in the non-overlapping regions. ARL \cite{ARL} and our CLGD do not have such a disadvantage, but the randomly sampled lines in ARL decrease the registration accuracy.

\begin{wraptable}{r}{6.7cm} \small
\centering
\vspace{-0.2cm}
\caption{{\small Running time (s) and GPU memory (GB) costs of different distance metrics under the optimization-based rigid registration task.}} \label{REGISTRATION:EFFICIENCY}
\setlength{\tabcolsep}{2.9mm}{
\begin{tabular}{l|c c}
\toprule
 & Time  &  GPU Memory \\
\hline
  EMD \cite{EMD} & 11 & 1.680\\
  CD \cite{CD} & 11 & 1.809\\
  ARL \cite{ARL} & 281 & 7.202 \\
  Ours & 13 & 1.813\\               
\bottomrule
\end{tabular}}
\vspace{-0.2cm}
\end{wraptable}

\paragraph{Efficiency Analysis.}
Table \ref{REGISTRATION:EFFICIENCY} lists the running time and GPU memory costs of the optimization-based registration driven by different metrics
\footnote{Note that for the unsupervised learning-based methods, different distance metrics are only used to train the network, and the inference time of a trained network with different metrics is equal. 
}. 
Since the number of points  is relatively small, EMD has comparable running time and GPU memory consumption to CD and our CLGD. In contrast, ARL \cite{ARL} is much less efficient because of the calculation of the intersections.

\subsection{Scene Flow Estimation}
This task 
aims to predict point-wise offsets $\mathbf{F}\in\mathbb{R}^{N_{\rm src}\times 3}$, which can align source point cloud 
$\mathbf{P}_{\rm src}$ to  target point cloud 
$\mathbf{P}_{\rm tgt}$. The optimization-based methods directly solve 
\begin{equation}
    \hat{\mathbf{F}}=\mathop{\arg\min}_{\mathbf{F}}\mathcal{D}(\mathbf{P}_{\rm src}+\mathbf{F},\mathbf{P}_{\rm tgt})+\alpha\mathcal{R}_{\rm smooth}(\mathbf{F}), \label{SCENEFLOW:EQ}
\end{equation}
where $\mathcal{R}_{\rm smooth}(\cdot)$ is the spatial smooth regularization term 
and the hyperparameter $\alpha>0$ balances the two iterms. 
Besides, we also evaluated the proposed CLGD by incorporating it into unsupervised learning-based frameworks, i.e., replacing the distance metrics of existing unsupervised learning-based frameworks with our CLGD to train the network, 
We adopted two SOTA unsupervised scene flow estimation methods named NSFP \cite{NEURALPRIOR} and SCOOP \cite{SCOOP}. 


\paragraph{Implementation Details.}
We used the Flyingthings3D dataset \cite{FLYINGTHINGS3D} preprocessed by \cite{HPLFLOWNET}, where 
$N_{\rm src}=N_{\rm tgt}=8192$. For the optimization-based methods, we used $\ell_2$-smooth regularization, 
set $\alpha=50$, 
and optimized the scene flow directly with the ADAM optimizer for 500 iterations with a learning rate of 0.01. For the unsupervised learning-based methods, we 
adopted  the same training settings as their original papers.

\begin{table}[h] \small
\centering
\caption{Quantitative comparisons of scene flow estimation on the Flyingthings3D dataset \label{SCENEFLOW:TABLE}.}
\begin{tabular}{l|l|c c c c}
\toprule 
~ & Method &  EPE3D(m)$\downarrow$ & Acc-0.05 $\uparrow$   & Acc-0.1$\uparrow$ & Outliers $\downarrow$ \\
\hline
\multirow{3}{*}{Optimization-based} & {EMD} \cite{EMD} & 0.3681 & 0.1894 & 0.4226 & 0.7838 \\
                                    & {CD} \cite{CD} & 0.1557 & 0.3489 & 0.6581 & 0.6799 \\
                                    & Ours & \textbf{0.0843}  & \textbf{0.5899} & \textbf{0.8722} & \textbf{0.4707} \\
                                    
                                    
\hline
\multirow{4}{*}{\makecell[l]{Unsupervised Learning}}    
                                    & NSFP \cite{NEURALPRIOR} & 0.0899 & 0.6095 & 0.8496 & 0.4472 \\
                                    & NSFP \cite{NEURALPRIOR} + Ours & \textbf{0.0662} & \textbf{0.7346} & \textbf{0.9107} & \textbf{0.3426} \\
                                    \cline{2-6}
                                    & SCOOP  \cite{SCOOP}           & 0.0839 & 0.5698 & 0.8516 & 0.4834 \\
                                    & SCOOP \cite{SCOOP} + Ours    & \textbf{0.0742} & \textbf{0.6134} & \textbf{0.8858} & \textbf{0.4497} \\

\bottomrule 
\end{tabular}
\end{table}

\begin{wraptable}{r}{6cm} \small
\centering
\vspace{-0.65cm}
\caption{{\small Running time (s) and GPU memory (GB) costs of different distance metrics under the optimization-based scene flow estimation task.}} \label{EFFICIENCY:SCENEFLOW}
\begin{tabular}{l|c c}
\toprule
 & Time &  GPU Memory \\
\hline
  EMD & 1011 & 2.204\\
  CD & 7 & 1.704\\
  Ours & 14 & 1.725\\               
\bottomrule
\end{tabular}
\vspace{-0.4cm}
\end{wraptable}

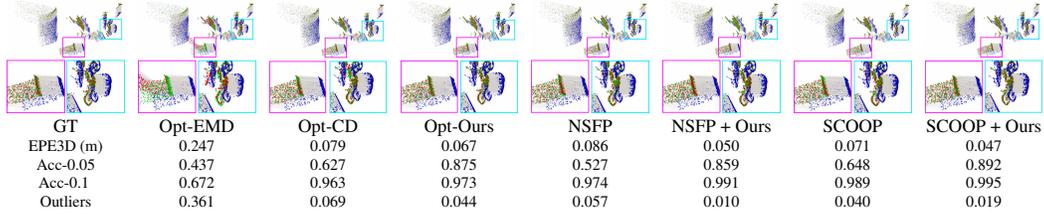
\begin{figure}[tbp] \small 
\centering
{
\begin{tikzpicture}[]
\node[] (a) at (0,1.45) {\includegraphics[width=0.117\textwidth]{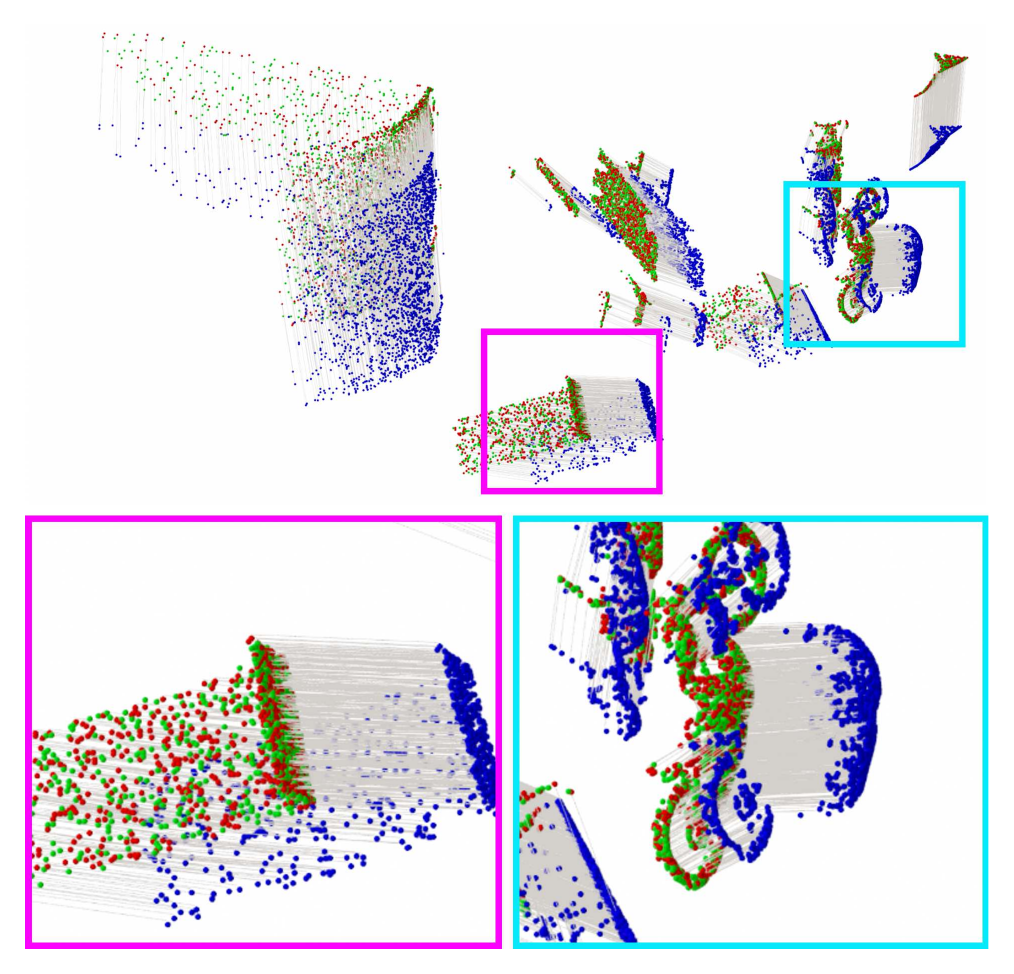} };
\node[] (a) at (0,0) { \tiny \makecell[c]{ {\scriptsize GT}\\ {\tiny EPE3D (m)}\\  {\tiny Acc-0.05} \\ {\tiny Acc-0.1\ }  \\{\tiny Outliers\ }}};

\node[] (b) at (12.2/7,1.45) {\includegraphics[width=0.117\textwidth]{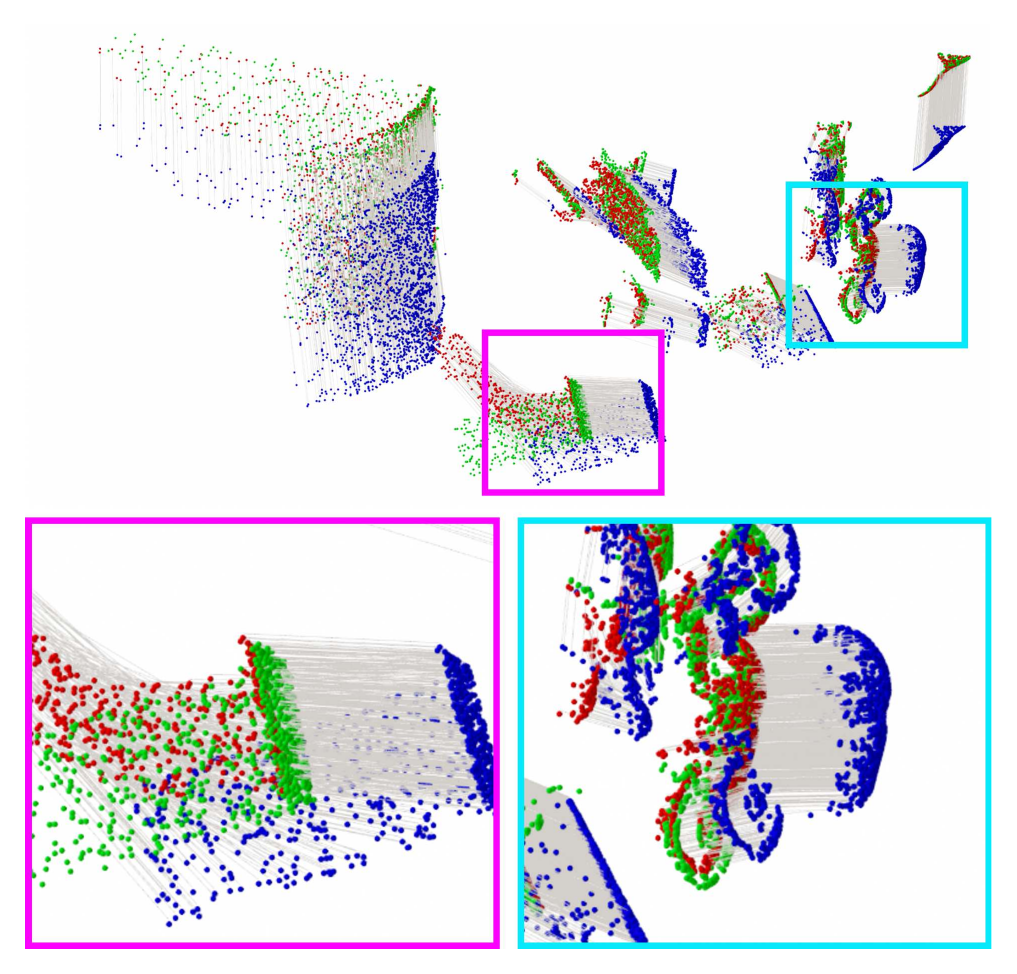}};
\node[] (b) at (12.2/7,0) {\tiny\makecell[c]{ {\scriptsize Opt-EMD}\\ {0.247} \\ {0.437} \\ {0.672} \\ {0.361} }};

\node[] (b) at (12.2/7*2,1.45) {\includegraphics[width=0.117\textwidth]{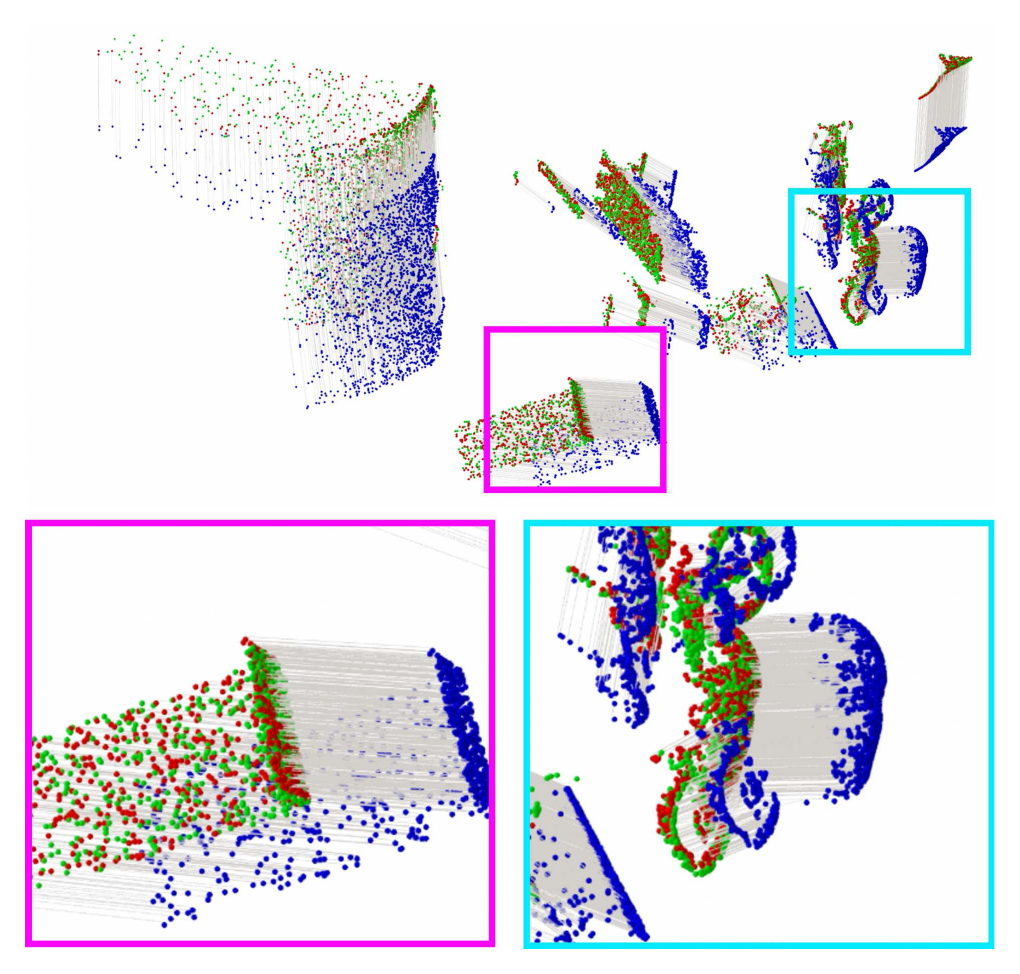}};
\node[] (b) at (12.2/7*2,0) {\tiny\makecell[c]{ {\scriptsize Opt-CD} \\ {0.079} \\ {0.627} \\ {0.963} \\ {0.069} }};

\node[] (c) at (12.2/7*3,1.45) {\includegraphics[width=0.117\textwidth]{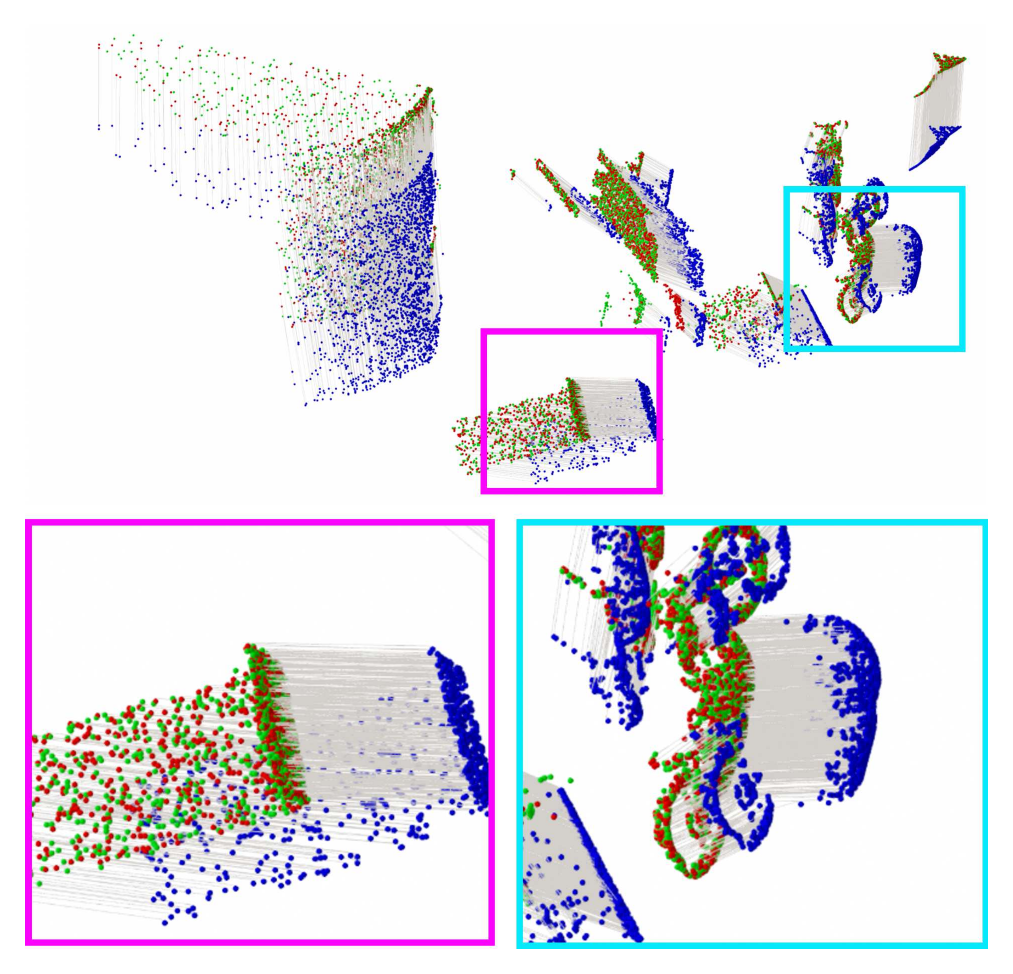}};
\node[] (b) at (12.2/7*3,0) {\tiny\makecell[c]{ {\scriptsize Opt-Ours} \\ {0.067} \\ {0.875} \\ {0.973} \\ {0.044} }};

\node[] (c) at (12.2/7*4,1.45) {\includegraphics[width=0.117\textwidth]{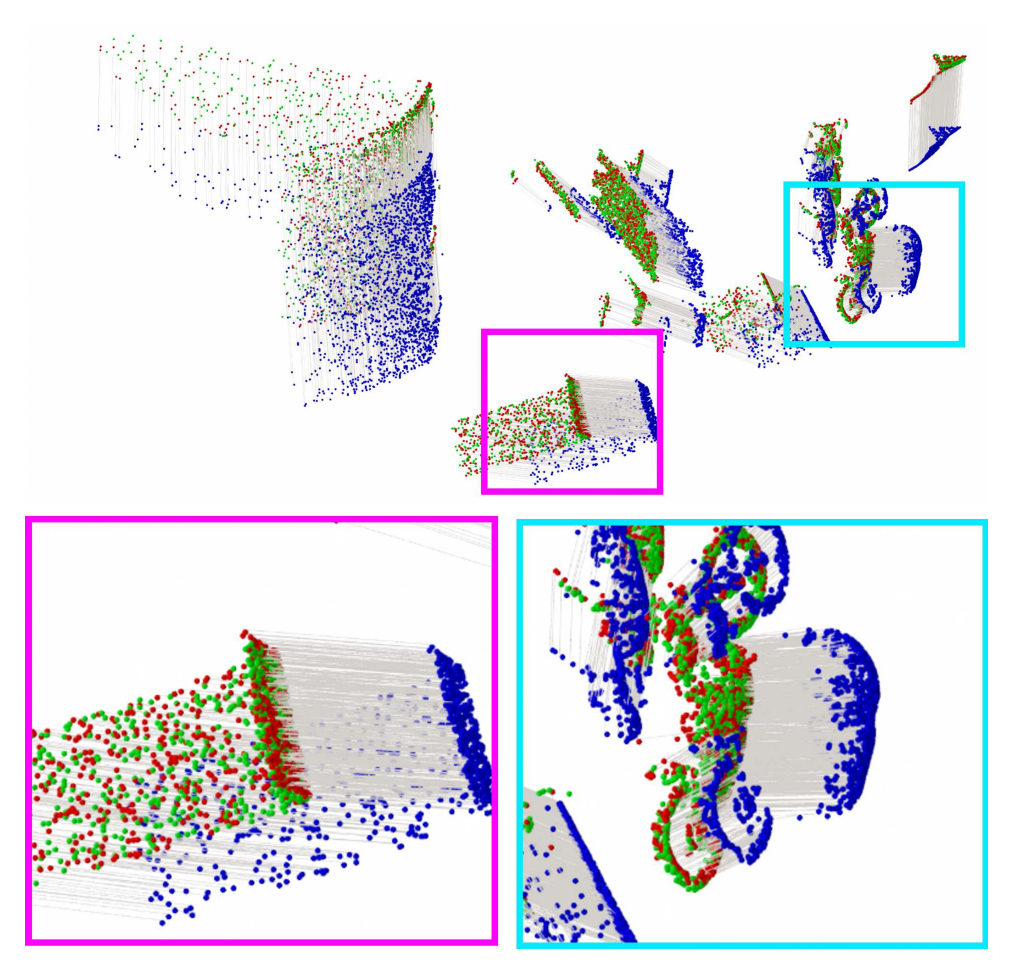} };
\node[] (b) at (12.2/7*4,0) {\tiny\makecell[c]{ {\scriptsize NSFP} \\ {0.086} \\ {0.527} \\ {0.974} \\ {0.057} }};

\node[] (d) at (12.2/7*5,1.45) {\includegraphics[width=0.117\textwidth]{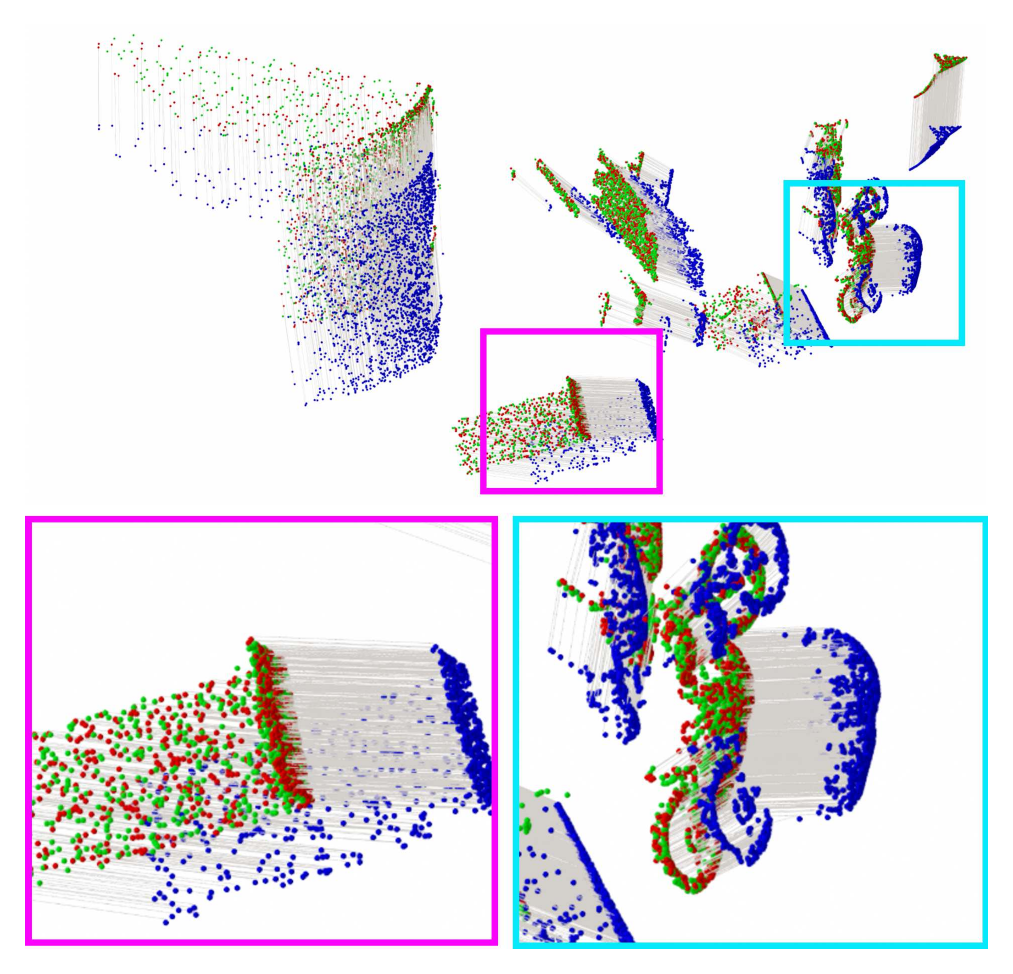}};
\node[] (d) at (12.2/7*5,0) {\tiny\makecell[c]{ {\scriptsize NSFP + Ours} \\ {0.050} \\ {0.859} \\ {0.991} \\ {0.010} }};

\node[] (d) at (12.2/7*6,1.45) {\includegraphics[width=0.117\textwidth]{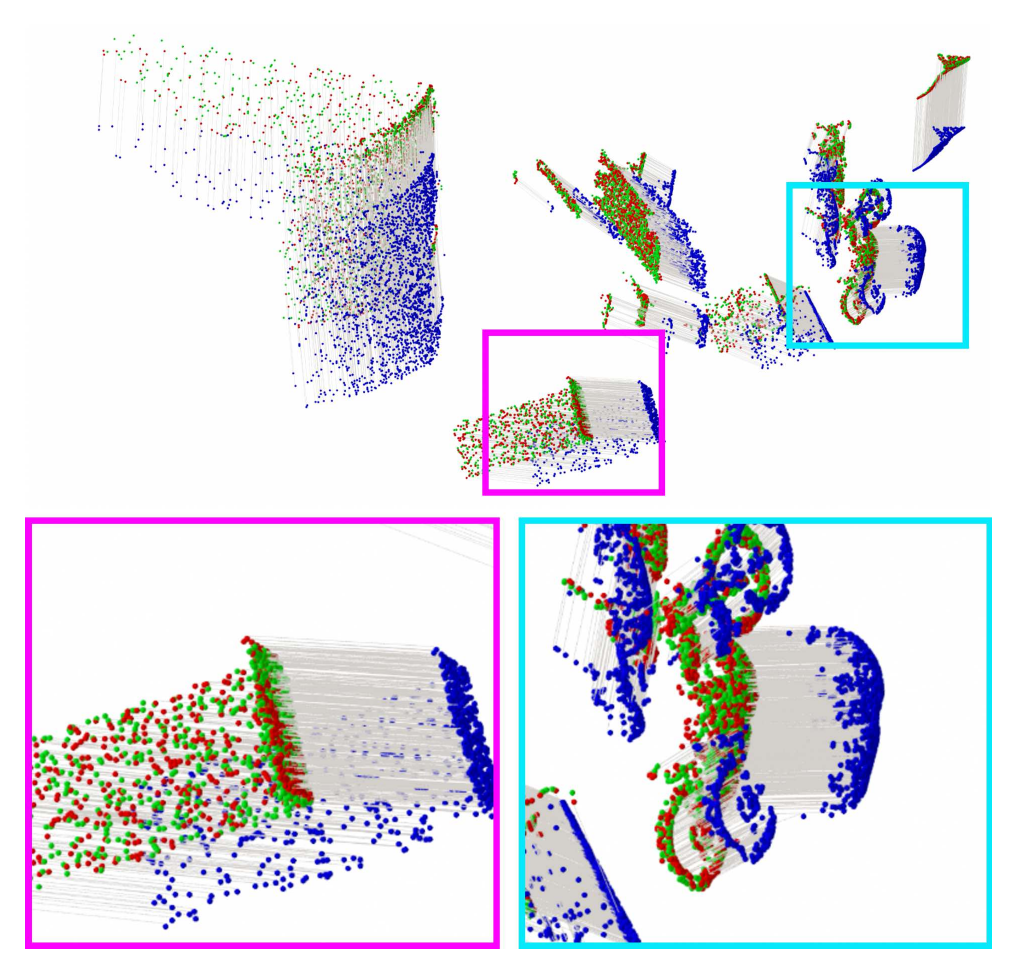} };
\node[] (d) at (12.2/7*6,0) {\tiny\makecell[c]{ {\scriptsize SCOOP} \\ {0.071} \\ {0.648} \\ {0.989} \\ {0.040} }};

\node[] (e) at (12.2/7*7,1.45) {\includegraphics[width=0.117\textwidth]{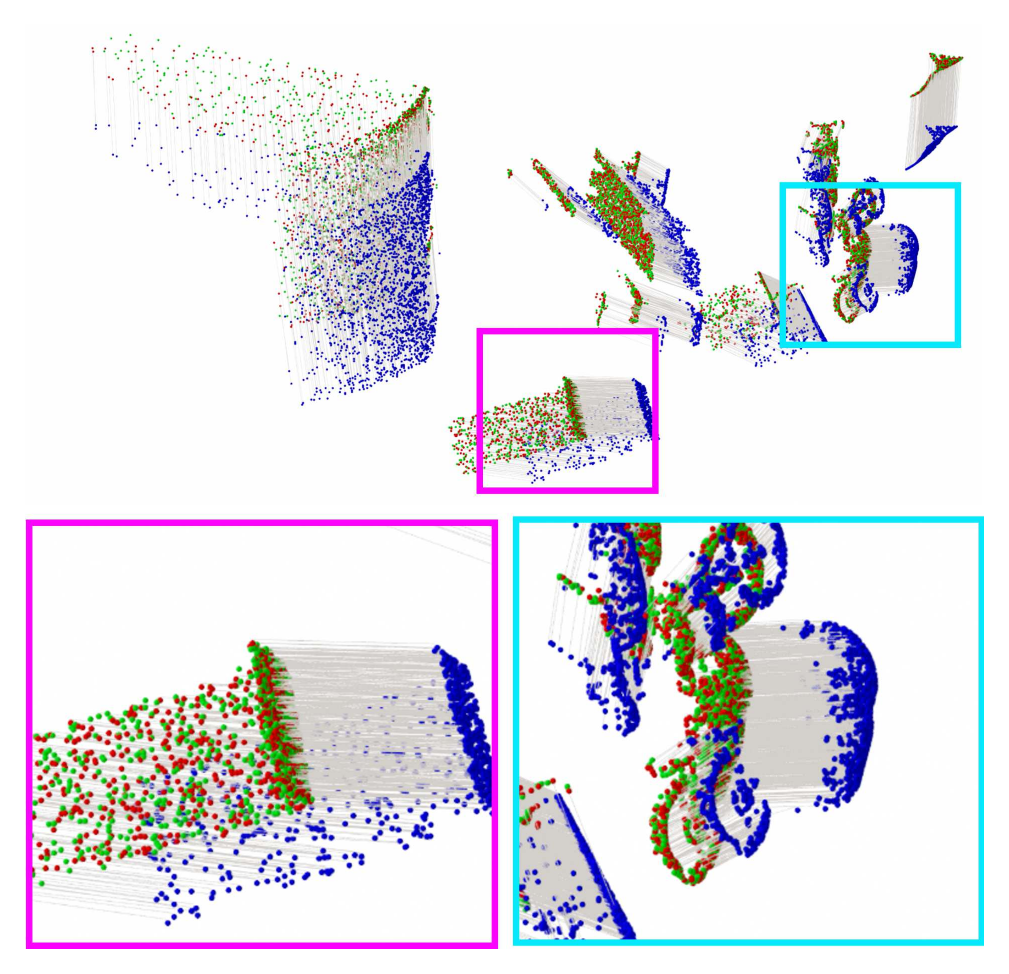}};
\node[] (d) at (12.2/7*7,0) {\tiny\makecell[c]{ {\scriptsize SCOOP + Ours} \\ {0.047} \\ {0.892} \\ {0.995} \\ {0.019} }};
\end{tikzpicture}
}
\vspace{-0.6cm}
\caption{Visual comparisons of scene flow estimation. The \textcolor{blue}{blue} and \textcolor{green}{green} points represent the source and target point clouds, respectively, and the \textcolor{red}{red} points are the warped source point cloud with estimated scene flows. \color{cyan}{\faSearch~} Zoom in to see details.} \label{SCENEFLOW:FIG}
\label{SHAPENET:VIS} 
\vspace{-0.3cm}
\end{figure}

\paragraph{Comparison.}
Following \cite{NEURALPRIOR,SCOOP}, we employed \textit{End Point Error (EPE)}, \textit{Flow Estimation Accuracy (Acc)} with thresholds 0.05 and 0.1 (denoted as \textit{Acc-0.05} and \textit{Acc-0.1}), and \textit{Outliers} as the evaluation metrics. 
From the results shown in Table \ref{SCENEFLOW:TABLE} and Fig. \ref{SCENEFLOW:FIG}, it can be seen that  our CLGD drives much more accurate scene flows than EMD and CD under the optimization-based framework, 
and our CLGD further boosts the accuracy of SOTA unsupervised learning-based methods to a significant extent, demonstrating its superiority and the importance of the distance metric in 3D point cloud modeling. 

\vspace{-0.3cm}

\paragraph{Efficiency Analysis.}
We also compared the running time and GPU memory cost of different metrics 
in the optimization-based framework. As shown in 
Table \ref{EFFICIENCY:SCENEFLOW},
EMD consumes much more time and GPU memory than CD and our CLGD. 

\subsection{Feature Representation} 
In this experiment, we trained an auto-encoder with different distance metrics used as the reconstruction loss to evaluate their abilities. 
Technically, an input point cloud is encoded into 
a global feature through the encoder, which is further decoded to reconstruct the input point cloud through a decoder.  
After training, we used the encoder to represent point clouds as features for classification by an SVM classifier. 
\paragraph{Implementation Details.} We built a auto-encoder with MLPs 
and used the ShapeNet \cite{SHAPENET} and ModelNet40 \cite{MODELNET40} datasets for training and testing, respectively. 
We trained the network for 300 epochs using the  ADAM optimizer with a learning rate of $10^{-3}$. We refer the readers to the \textit{Supplemental Material} for more details.

\begin{wraptable}{r}{5.5cm} \small
\centering
\vspace{-0.65cm}
\caption{Classification accuracy by SVM on 
ModelNet40 \cite{MODELNET40}. 
} \label{CLASS:ACC}
\begin{tabular}{l|c c  c }
\toprule
 Loss Function & EMD & CD & Ours  \\
\hline
Accuracy (\%) & 78.12 & 78.89 & 81.28 \\
\bottomrule
\end{tabular}
\end{wraptable} 

\begin{figure}[h]
\centering
\subfigure[EMD]{\includegraphics[width=0.23\linewidth]{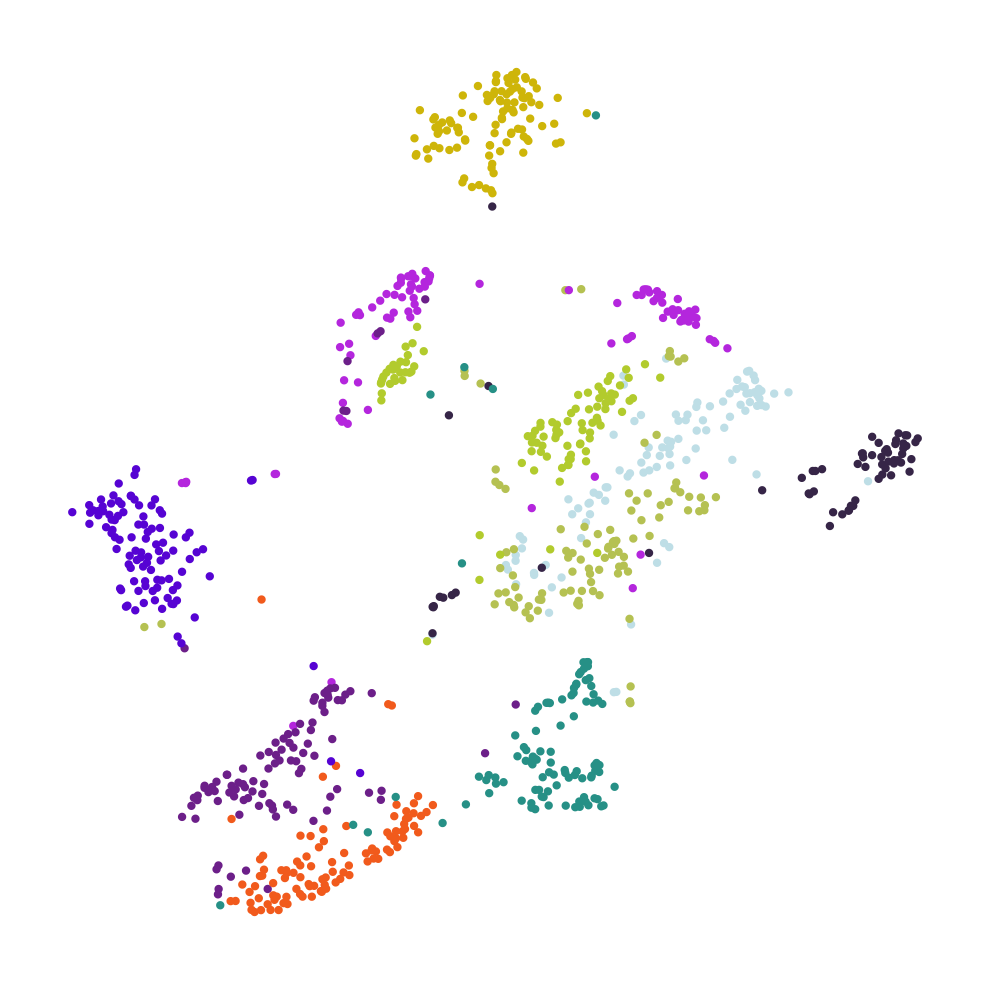}} 
\subfigure[CD]{\includegraphics[width=0.23\linewidth]{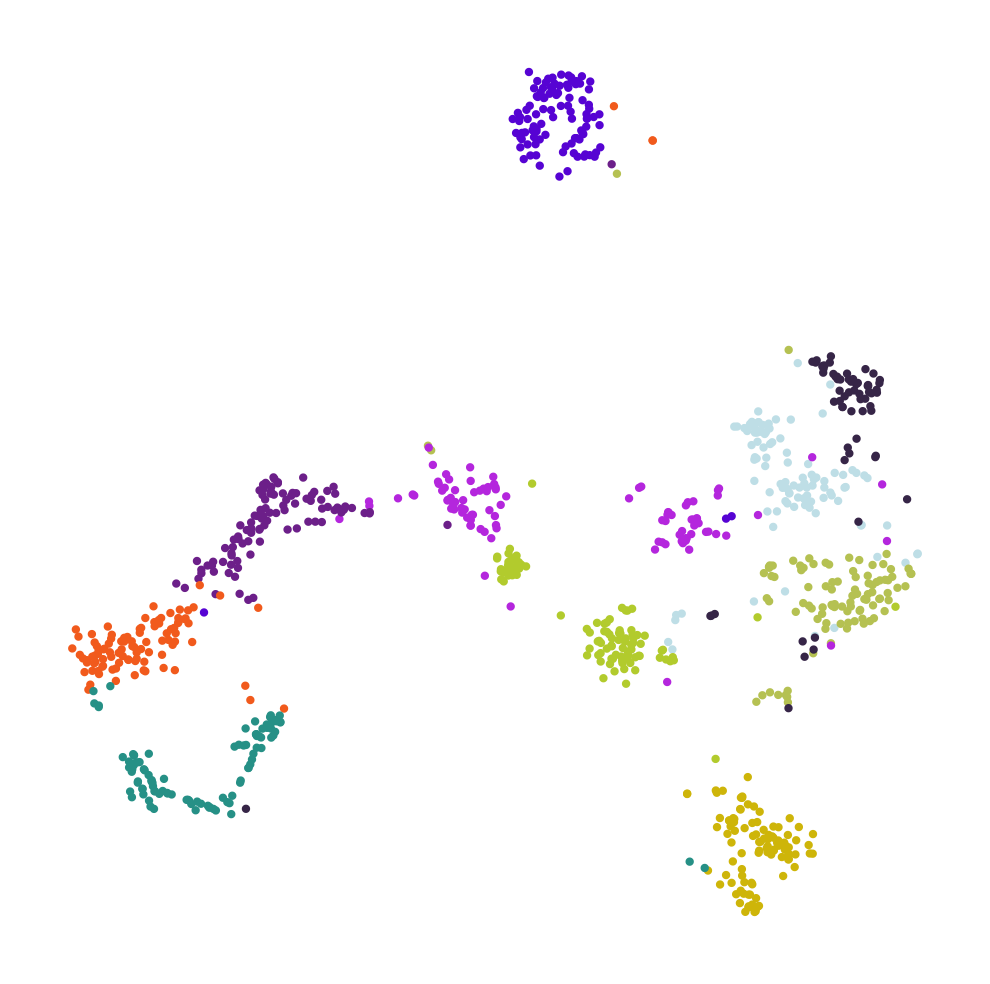}} 
\subfigure[Ours]{\includegraphics[width=0.23\linewidth]{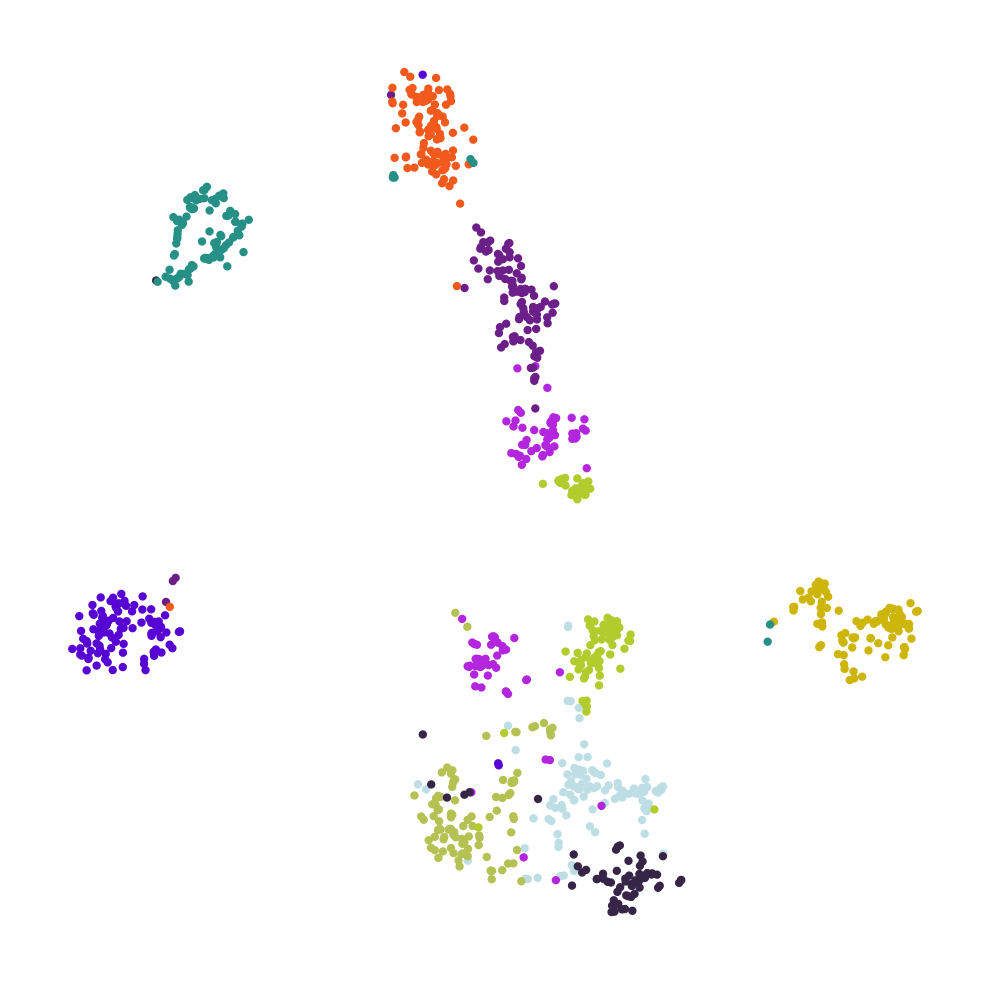}} 
\includegraphics[width=0.25\linewidth]{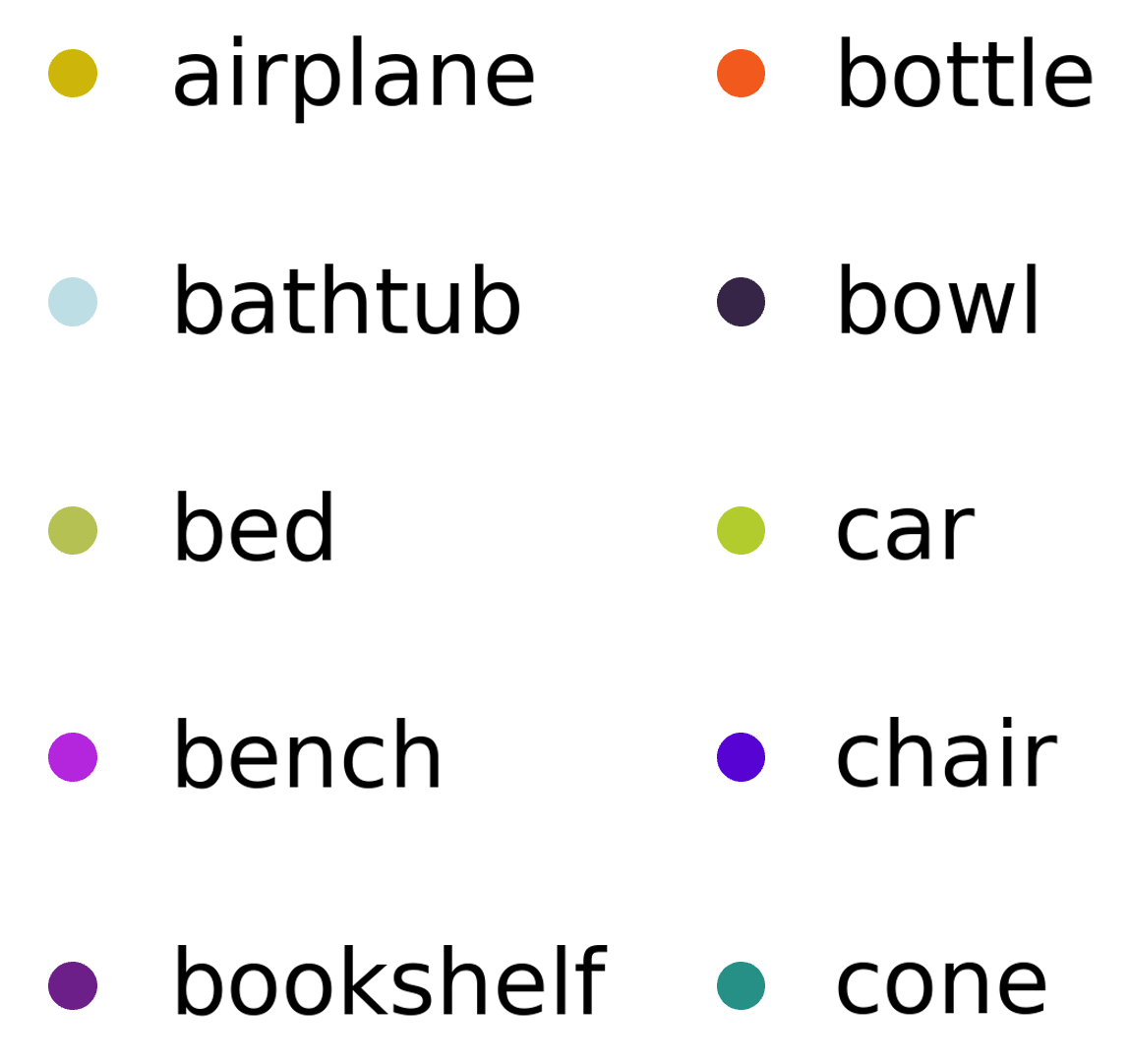}
\vspace{-0.3cm}
\caption{The T-SNE clustering visualization of the features obtained from the auto-encoder trained with different distance metrics.}\label{CLASS:FIG}
\vspace{-0.3cm}
\end{figure}

\paragraph{Comparison.} As listed in Table \ref{CLASS:ACC}, the higher classification accuracy by our CLGD demonstrates the auto-encoder driven by our CLGD can learn more discriminative features. Besides, we also used T-SNE \cite{TSNE} to visualize the 2D embeddings of the global features of 1000 shapes from 10 categories in Fig. \ref{CLASS:FIG} to show the advantages of our method more intuitively.


\vspace{-0.3cm}

\subsection{Ablation Study} \label{SEC:ABLATION}
We carried out comprehensive ablation studies on shape reconstruction and rigid registration to verify the rationality of the design in our CLGD.

\paragraph{Directional Distance.}
We removed $f$ or $\mathbf{v}$ from the directional distance $\mathbf{g}$ when modeling the local surface geometry in Section \ref{LOCAL:GEO}. The accuracy of reconstructed 3D mesh shapes 
is listed in Table \ref{ABLATION:GEO}, where it can be seen that removing either one of $f$ and $\mathbf{v}$ would decrease the reconstruction accuracy, especially $\mathbf{v}$, because using $f$ or $\mathbf{v}$ alone could hardly characterize the underlying surfaces of point clouds. 

\paragraph{Size of $\Omega(\mathbf{q},\mathbf{P})$.} We changed the size of $\Omega(\mathbf{q},\mathbf{P})$ by varying the value of $K$ used in Eqs. (\ref{UDF}) and (\ref{UDF_GRAD}). 
As shown in Table \ref{ABLATION:KVALUE}, a  small or large value of $K$ would decrease the performance of CLGD because 
a small $K$ only covers a tiny region on the point cloud, resulting in that $\mathbf{g}(\mathbf{q},\mathbf{P})$ cannot represent the local surface geometry induced by $\mathbf{q}$, 
while a large $K$ 
includes too many points, making $\mathbf{g}(\mathbf{q},\mathbf{P})$ would become smooth and ignore the details of the local surface geometry induced by $\mathbf{q}$. 

\paragraph{Reference Points.} 
We studied how the number of reference points affects CLGD by varying the value of $R$. 
As shown in Table \ref{ABLATION:M},
too few reference points cannot sufficiently capture the local surface geometry difference, 
while too many reference points are not necessary for improving performance but 
compromise efficiency.
We also studied how the distribution of reference points  affects CLGD by varying the value of $T$, concluding that 
the reconstruction accuracy decreases if the reference points are either too close to or too far from the underlying surface.
Finally, from Table \ref{ABLATION:SRC}, it can be seen that after removing the source point cloud $\mathbf{P}_{\rm src}$ from $\mathbf{Q}$, the reconstruction accuracy decreases, demonstrating the necessity of $\mathbf{P}_{\rm src}$ as reference points. 

\begin{table}[h]\small
\centering
\vspace{-0.4cm}
\caption{Results of the ablative studies about our CLGD under the 3D shape reconstruction task. 
}\label{ABLATION:SRC} \label{ABLATION:M} \label{ABLATION:GEO}\label{ABLATION:KVALUE}
\begin{tabular}{c|c |c c c | c c c | c}
\toprule
Geometry & $K$ & $R$ & $T$ & $\mathbf{P}_{\rm src}\subset\mathbf{Q}$ &  NC  $\uparrow$ & F-0.5\% $\uparrow$& F-1\% $\uparrow$& Time (s) $\downarrow$\\
\hline
$f$ &5&10&3&   \checkmark  & 0.791 & 0.611  & 0.776 &123  \\
$\mathbf{v}$&5&10&3& \checkmark & 0.907 & 0.675 & 0.888 & 122  \\
\hline
$\mathbf{g}$&1&10&3& \checkmark  & 0.913 & 0.687  & 0.888   & 119\\
$\mathbf{g}$&3&10&3& \checkmark  & 0.916 & 0.717 & 0.908  & 124 \\
$\mathbf{g}$ &10&10&3&\checkmark  & 0.913 & 0.712 & 0.907 & 132 \\
\hline
$\mathbf{g}$  &5&1&3& \checkmark   & 0.899 & 0.610 & 0.831 & 110  \\
$\mathbf{g}$  &5&5&3&        \checkmark   & 0.912 & 0.686 & 0.886 & 112 \\
$\mathbf{g}$  &5&20&3&        \checkmark  &0.913 & 0.693 & 0.893 & 149\\
\hline
$\mathbf{g}$&5&10&1&    \checkmark  & 0.916 & 0.722 & 0.909 & 126\\
$\mathbf{g}$&5&10&5&   \checkmark  & 0.911 & 0.692 & 0.896 & 126\\
$\mathbf{g}$&5&10&10&  \checkmark  & 0.891 & 0.610 & 0.836 & 126\\
\hline
$\mathbf{g}$&5&10&3&     \xmark  & 0.915 & 0.714 & 0.909 & 123 \\
\hline
$\mathbf{g}$&5&10&3& \checkmark  & 0.916 & 0.719 & 0.911 & 126  \\
\bottomrule
\end{tabular}
\vspace{-0.3cm}
\end{table}

\begin{wraptable}{r}{5cm} \small
\centering
\vspace{-0.65cm}
\caption{Effect of the value of $\beta$ on rigid registration accuracy.
}\label{ABLATION:BETA}
\setlength{\tabcolsep}{4.5mm}{
\begin{tabular}{c|c c c c }
\toprule
 $\beta$ & RE ($^\circ$) & TE (m) \\
\hline
 0 & 5.637 & 0.286  \\
 1 & 1.598 & 0.080  \\
 3 & 1.040 & 0.040  \\
 5 & 1.290 & 0.037  \\
 10 & 1.769 & 0.046  \\
\bottomrule
\end{tabular}}
\vspace{-0.3cm}
\end{wraptable}

\paragraph{Effectiveness of $s(\mathbf{q})$ in Rigid Registration.}
We set various values of $\beta$ in Eq. \eqref{CLGD} in the rigid registration task. 
As shown in Table \ref{ABLATION:BETA}, when $\beta=0$, the registration accuracy decreases significantly because the source and target point clouds are partially overlapped, and those reference points, which 
correspond to 
the non-overlapping regions and negate 
the optimization process, are still equally taken into account. 
On the contrary, if the value of $\beta$ is large, the registration accuracy only drops slightly. 
The reason is that the weights of the reference points with similar local surface geometry differences would be quite different, and the details of the point clouds may be ignored,  making the registration accuracy slightly affected.

\section{Conclusion and Discussion}\label{CONCLUSION}
We have introduced CLGD, a novel, robust, and generic distance metric for 3D point clouds. 
CLGD measures the difference between the local surfaces underlying 3D point clouds under evaluation, 
which is significantly different from existing metrics. 
Our experiments have demonstrated the significant superiority of CLGD in terms of both efficiency and effectiveness for various tasks, including shape reconstruction, rigid registration, scene flow estimation, and feature representation. Besides, comprehensive ablation studies have validated its rationality. 
We believe CLGD will advance the field of 3D point cloud processing and analysis.

Our proposed CLGD measures the differences between two point clouds by modeling the local geometry on their underlying surfaces. As a result, it is important that the points of a point cloud are distributed on the surfaces rather than completely random, as the local geometry induced by reference points in a randomly distributed point cloud would be meaningless. Consequently, in some generation tasks, CLGD requires a relatively good initialization in the beginning. Otherwise, it may perform worse than distance metrics that focus on the point cloud, such as EMD and CD.

{
\small
\bibliographystyle{abbrv}
\bibliography{ref}
}

\end{document}


\maketitle

\setcounter{equation}{6}
\setcounter{figure}{7}

In this supplementary material, we provide more details and analyses of the experiments conducted in the manuscript. 
\section{Shape Reconstruction}
\begin{figure*}[h]
    \centering
    \includegraphics[width=0.85\linewidth]{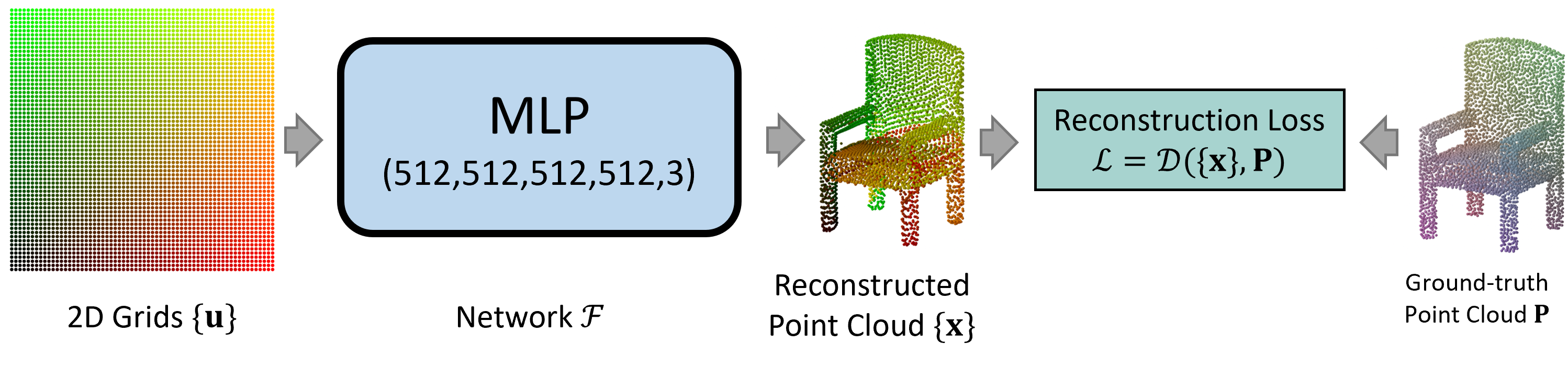}
    \vspace{-0.2cm}
    \caption{The network architecture used for 3D shape reconstruction. 
    }
    \label{SHAPERECONSTRUCTION:MLP}
\end{figure*}

\paragraph{Network Architecture.} Fig. \ref{SHAPERECONSTRUCTION:MLP} shows the architecture of the neural network $\mathcal{F}(\cdot) :\mathbb{R}^2\to\mathbb{R}^3$ used for 3D shape reconstruction, which maps a 2D grid $\mathbf{u}:=[u_1,~u_2]^\mathsf{T}$ to a 3D coordinate $\mathbf{x}:=[x,~y,~z]^\mathsf{T}$, i.e., $\mathbf{x}=\mathcal{F}(\mathbf{u})$. 
\paragraph{Normal Estimation.}

According to the differential geometric property of $\mathcal{F}$, we could obtain the normal vector of a typical 3D point $\mathbf{x}$ corresponding to   $\mathbf{u}$ by
\begin{equation}
    \mathbf{n}(\mathbf{x})=\frac{\frac{\partial \mathcal{F}(\mathbf{u})}{\partial u_1}\times \frac{\partial \mathcal{F}(\mathbf{u})}{\partial u_2} \ \ }{\|\frac{\partial \mathcal{F}(\mathbf{u})}{\partial u_1}\times \frac{\partial \mathcal{F}(\mathbf{u})}{\partial u_2}\|_2},
\end{equation}
where $\times$ is the cross product of two vectors, and the partial derivatives can be calculated through the backpropagation of the network. The resulting normal vectors as well as the 3D coordinates are used for mesh reconstruction via SPSR \cite{SPSR}. 

\paragraph{Evaluation Metric.}
Let $\mathcal{S}_{\rm rec}$ and $\mathcal{S}_{\rm GT}$ denote the reconstructed and ground-truth 3D meshes, respectively, on which we randomly sample $N_{\rm eval}=10^5$ points, denoted as $\mathbf{P}_{\rm rec}$ and $\mathbf{P}_{\rm GT}$. For each point on $\mathbf{P}_{\rm rec}$ and $\mathbf{P}_{\rm GT}$, the normal of the triangle face where it is sampled is considered to be its normal vector, and the normal sets of $\mathbf{P}_{\rm rec}$ and $\mathbf{P}_{\rm GT}$ are denoted as $\mathbf{N}_{\rm rec}$ and $\mathbf{N}_{\rm GT}$, respectively.

Let $\texttt{NN\_Norm}(\mathbf{x},\mathbf{P})$ be the operator that returns the normal vector of the point $\mathbf{x}$'s nearest point in the point cloud $\mathbf{P}$. The NC is defined as 
\begin{equation}
\begin{aligned}
\texttt{NC}(\mathcal{S}_{\rm rec},\mathcal{S}_{\rm GT})&=\frac{1}{N_{\rm eval}}\sum_{\mathbf{x}\in\mathbf{P}_{\rm rec}}\left|\mathbf{N}_{\rm rec}(\mathbf{x})\cdot\texttt{NN\_Normal}(\mathbf{x},\mathbf{P}_{\rm GT})\right|\\
&+\frac{1}{N_{\rm eval}}\sum_{\mathbf{x}\in\mathbf{P}_{\rm GT}}\left|\mathbf{N}_{\rm GT}(\mathbf{x})\cdot\texttt{NN\_Normal}(\mathbf{x},\mathbf{P}_{\rm rec})\right|.
\end{aligned}
\end{equation}
The F-Score is defined as the harmonic mean between the precision and the recall of points that lie within a certain distance threshold $\epsilon$ between $\mathcal{S}_{\rm rec}$ and $
\mathcal{S}_{\rm GT}$,
\begin{equation}
\texttt{F-Score}(\mathcal{S}_{\rm rec},\mathcal{S}_{\rm GT},\epsilon)=\frac{2\cdot\texttt{Recall}\cdot\texttt{Precision}}{\texttt{Recall}+\texttt{Precision}},
\end{equation}
where
\begin{equation}
    \begin{aligned}
        \texttt{Recall}(\mathcal{S}_{\rm rec},\mathcal{S}_{\rm GT},\epsilon)&=\left|\left\{\mathbf{x}_1\in\mathbf{P}_{\rm rec},\ {\rm s.t.} \mathop{\rm min}_{\mathbf{x}_2\in\mathbf{P}_{\rm GT}}\|\mathbf{x}_1-\mathbf{x}_2\|_2<\epsilon\right\}\right|,\\
        \texttt{Precision}(\mathcal{S}_{\rm rec},\mathcal{S}_{\rm GT},\epsilon)&=\left|\left\{\mathbf{x}_2\in\mathbf{P}_{\rm GT},\ {\rm s.t.} \mathop{\rm min}_{\mathbf{x}_1\in\mathbf{P}_{\rm rec}}\|\mathbf{x}_1-\mathbf{x}_2\|_2<\epsilon\right\}\right|. \nonumber
    \end{aligned}
\end{equation}


\section{Rigid Registration}
\paragraph{Evaluation Metric.}
Let $[\hat{\mathbf{R}},\hat{\mathbf{t}}]$ and $[\mathbf{R}_{\rm GT}, \mathbf{t}_{\rm Gt}]$ be the estimated and ground-truth transformation, respectively. The two evaluation metrics named Rotation Error (RE) and Translation Error (TE) are defined as 
\begin{equation}
\begin{aligned}
    \texttt{RE}(\hat{\mathbf{R}},\mathbf{R}_{\rm GT})=\angle(\mathbf{R}^{-1}_{\rm GT}\hat{\mathbf{R}}),\quad
    \texttt{TE}(\hat{\mathbf{t}},\mathbf{t}_{\rm GT})=\|\hat{\mathbf{t}}-\mathbf{t}_{\rm GT}\|_2,
\end{aligned}
\end{equation}
where $\angle (\mathbf{A})={\rm arccos}(\frac{\texttt{trace}(\mathbf{A})-1}{2})$ returns the angle of rotation matrix $\mathbf{A}$ in degrees.

\paragraph{Unsupervised Learning-based Method.}
\begin{figure}[h]
\centering
\includegraphics[width=0.95\textwidth]{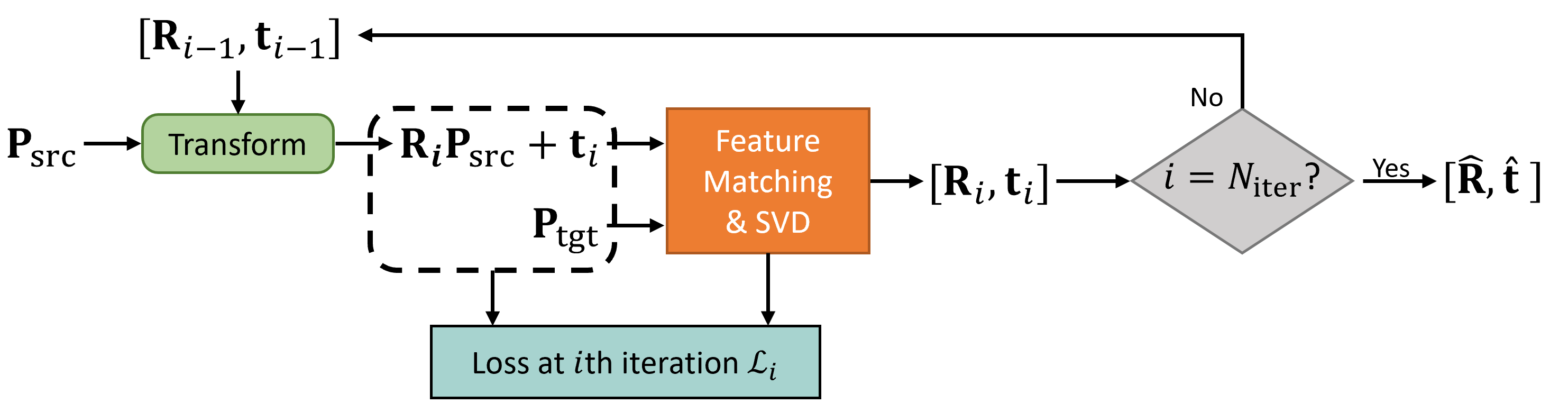}
\caption{The overall pipeline of the unsupervised learning-based rigid registration built upon 
RPM-Net \cite{RPMNET}. The loss $\mathcal{L}_i$ is defined in Eq. \eqref{RPM:LOSS}. $N_{\rm iter}$ is the total number of iterations, and $[\mathbf{R}_i,\mathbf{t}_i]$ is the resulting transformation at the $i$-th iteration.  See the original paper for detailed structures.}
\label{RPM:STRUC}
\end{figure}

\begin{figure}[h] \small 
\centering
{
\begin{tikzpicture}[]



\node[] (a) at (0,6) {\includegraphics[width=0.24\textwidth]{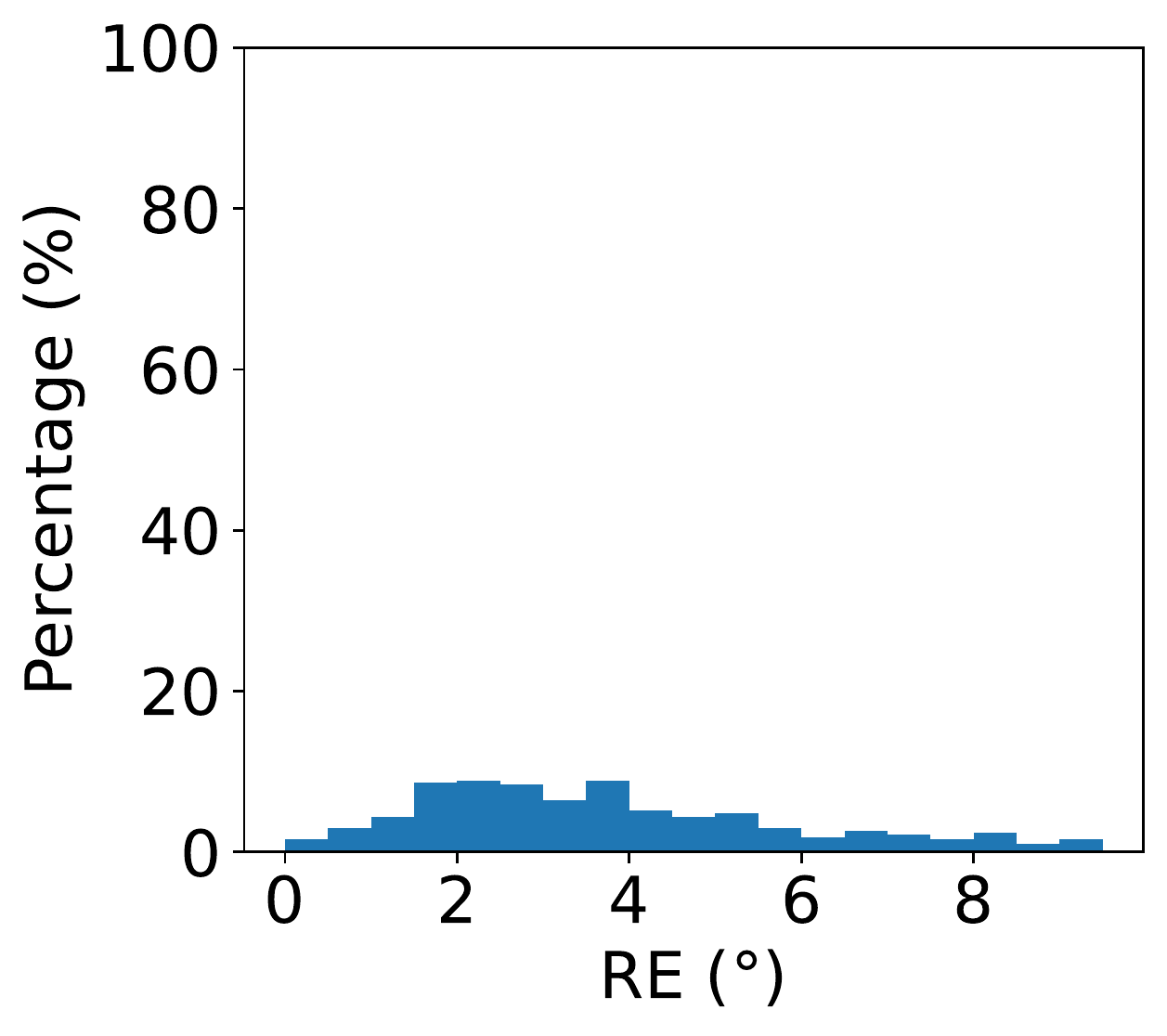}} ;
\node[] (a) at (10.5/3-0.1,6) {\includegraphics[width=0.24\textwidth]{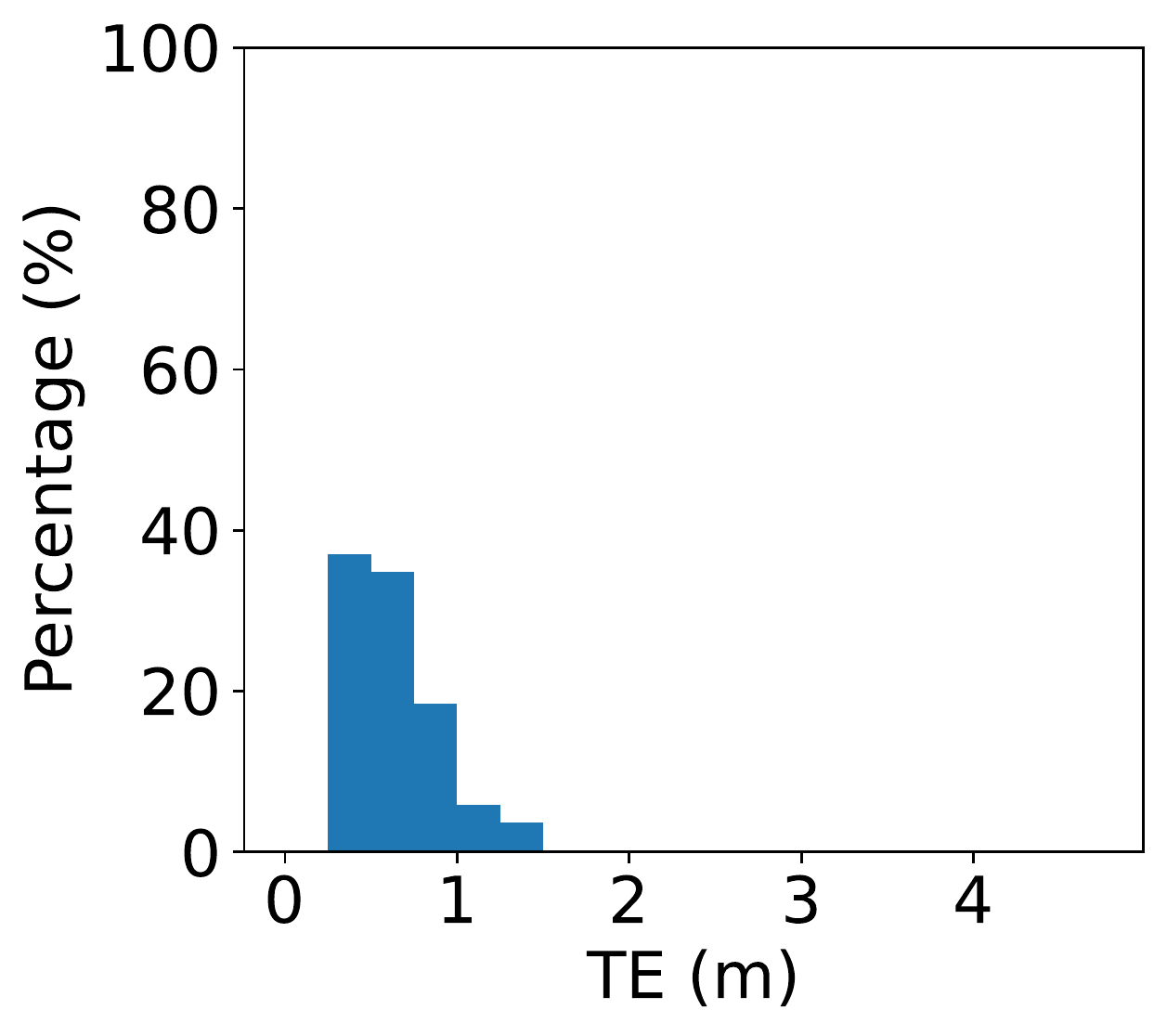}} ;
\node[] (a) at (10.5/6-0.05,4.5) {(a) EMD \cite{EMD}} ;

\node[] (a) at (10.5/3*2+0.1,6) {\includegraphics[width=0.24\textwidth]{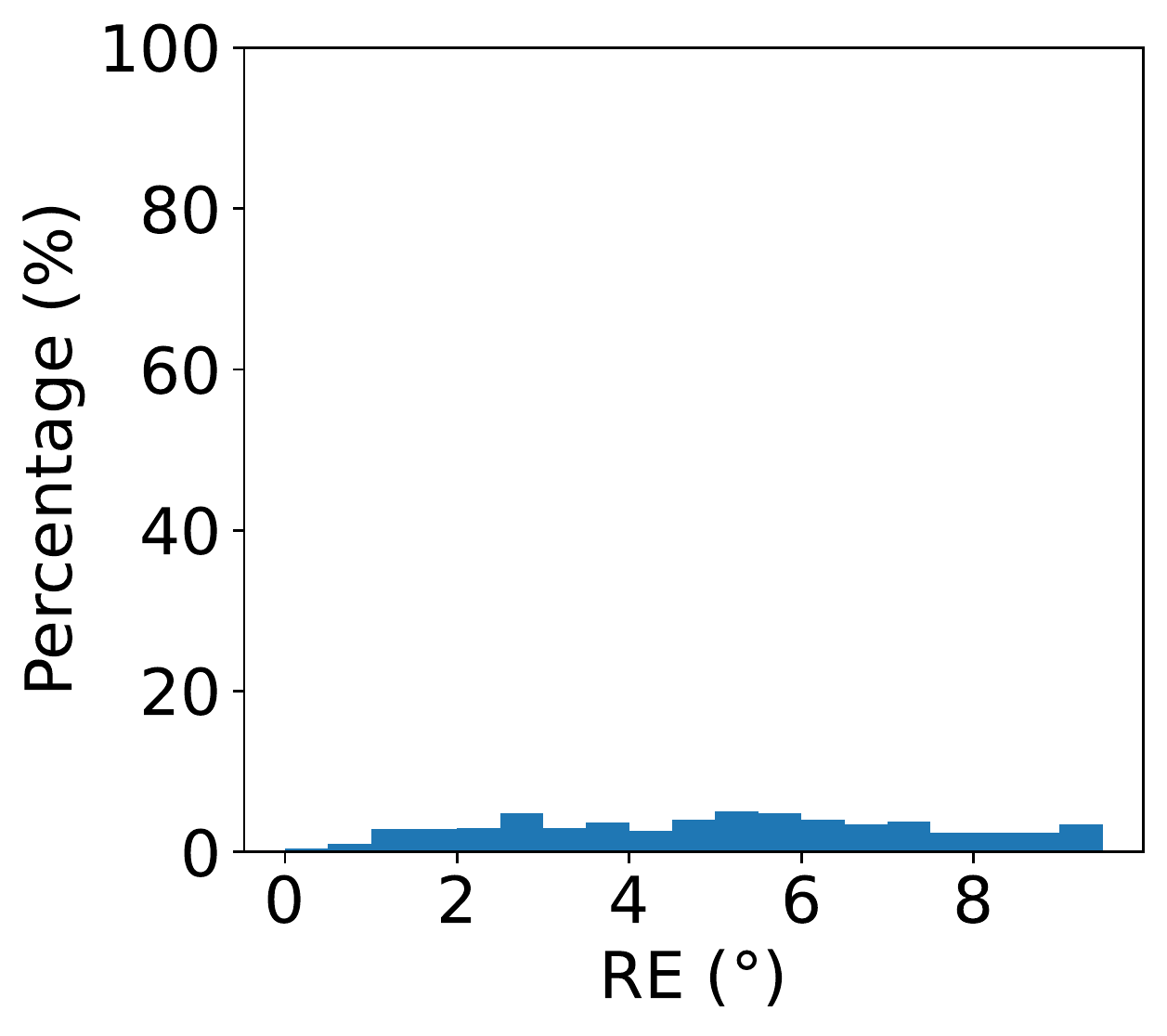}} ;
\node[] (a) at (10.5,6) {\includegraphics[width=0.24\textwidth]{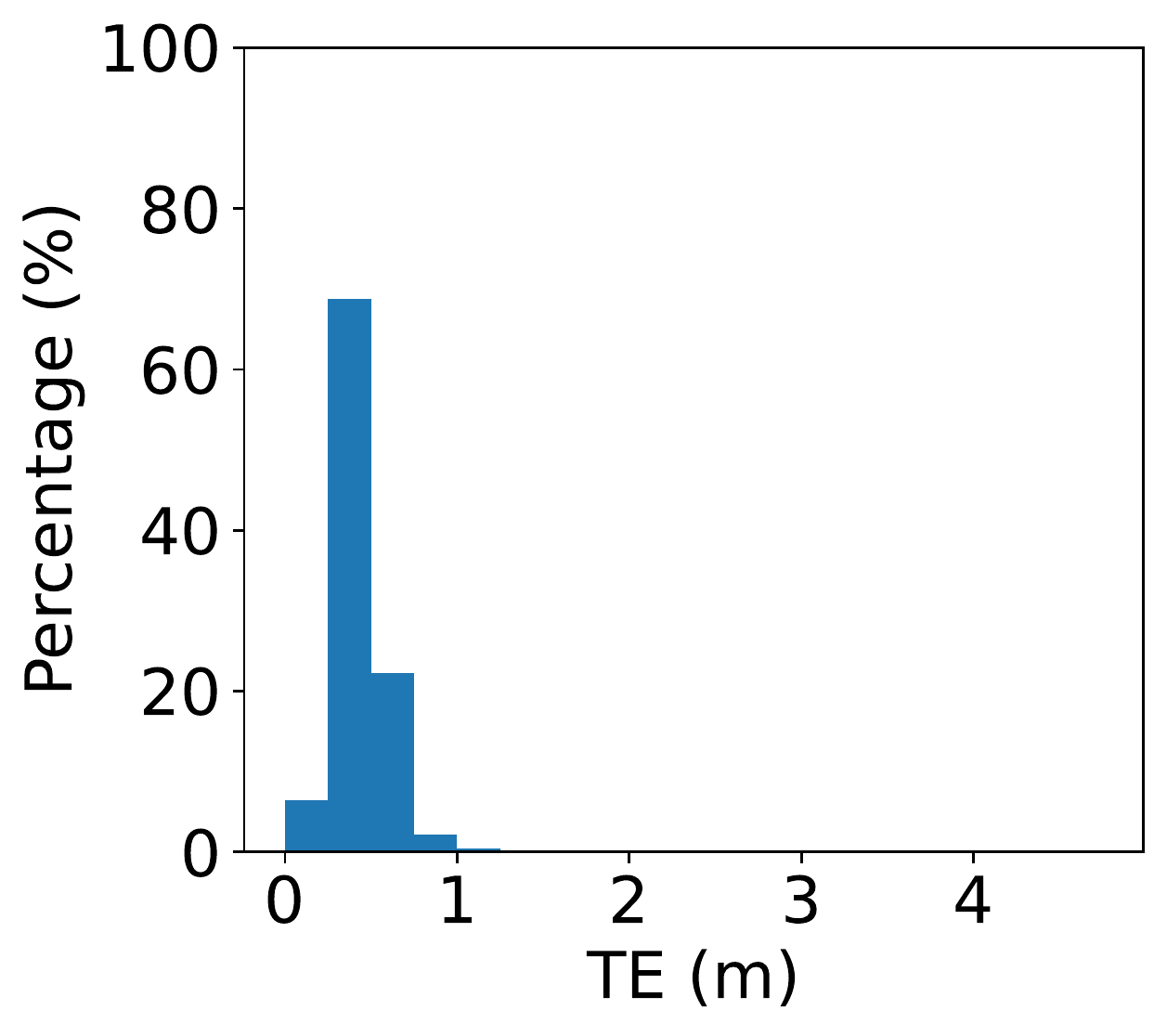}} ;
\node[] (a) at (10.5/6*5+0.05,4.5) {(b) CD \cite{CD}} ;

\node[] (a) at (0,2.5) {\includegraphics[width=0.24\textwidth]{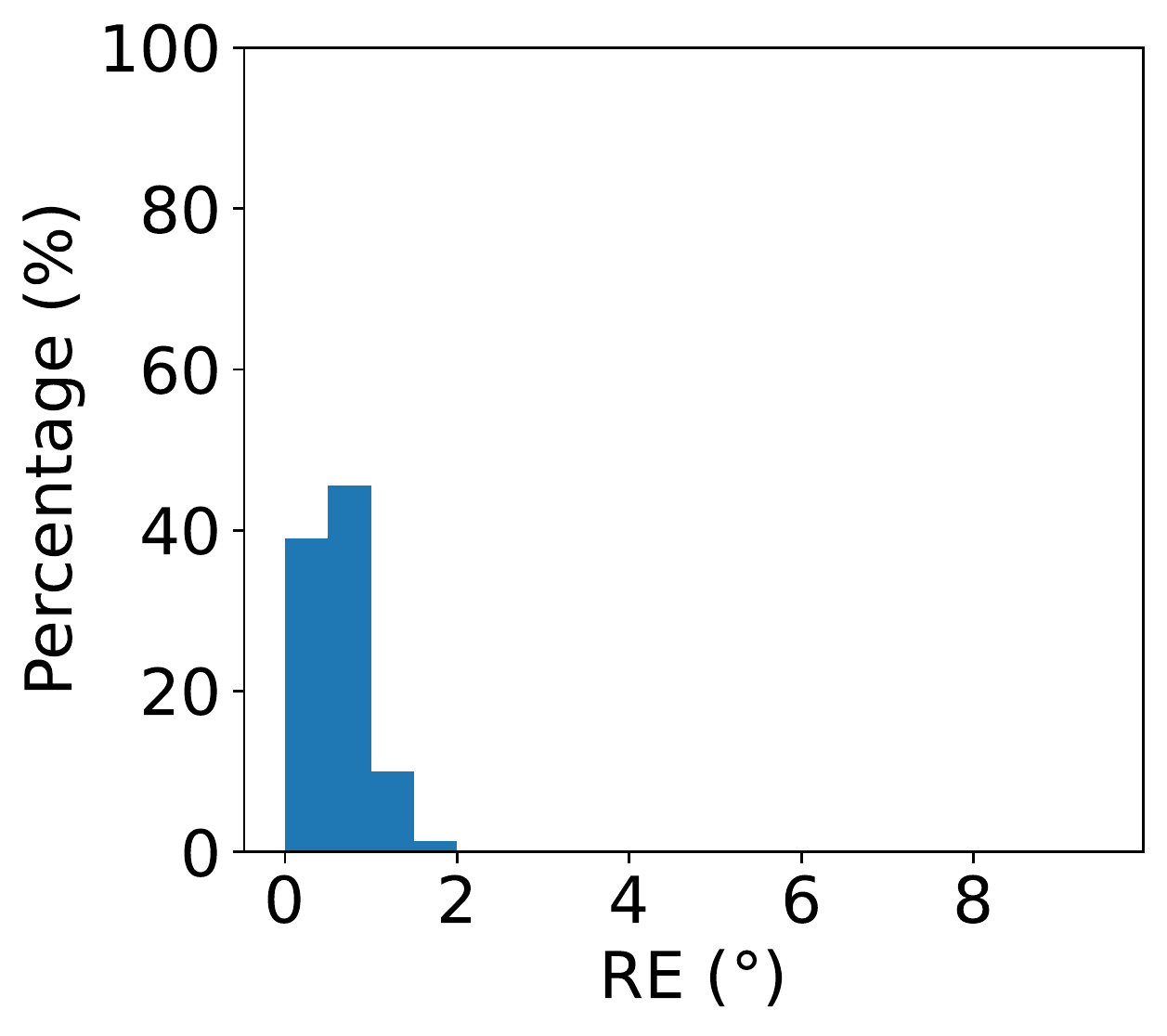}} ;
\node[] (a) at (10.5/3-0.1,2.5) {\includegraphics[width=0.24\textwidth]{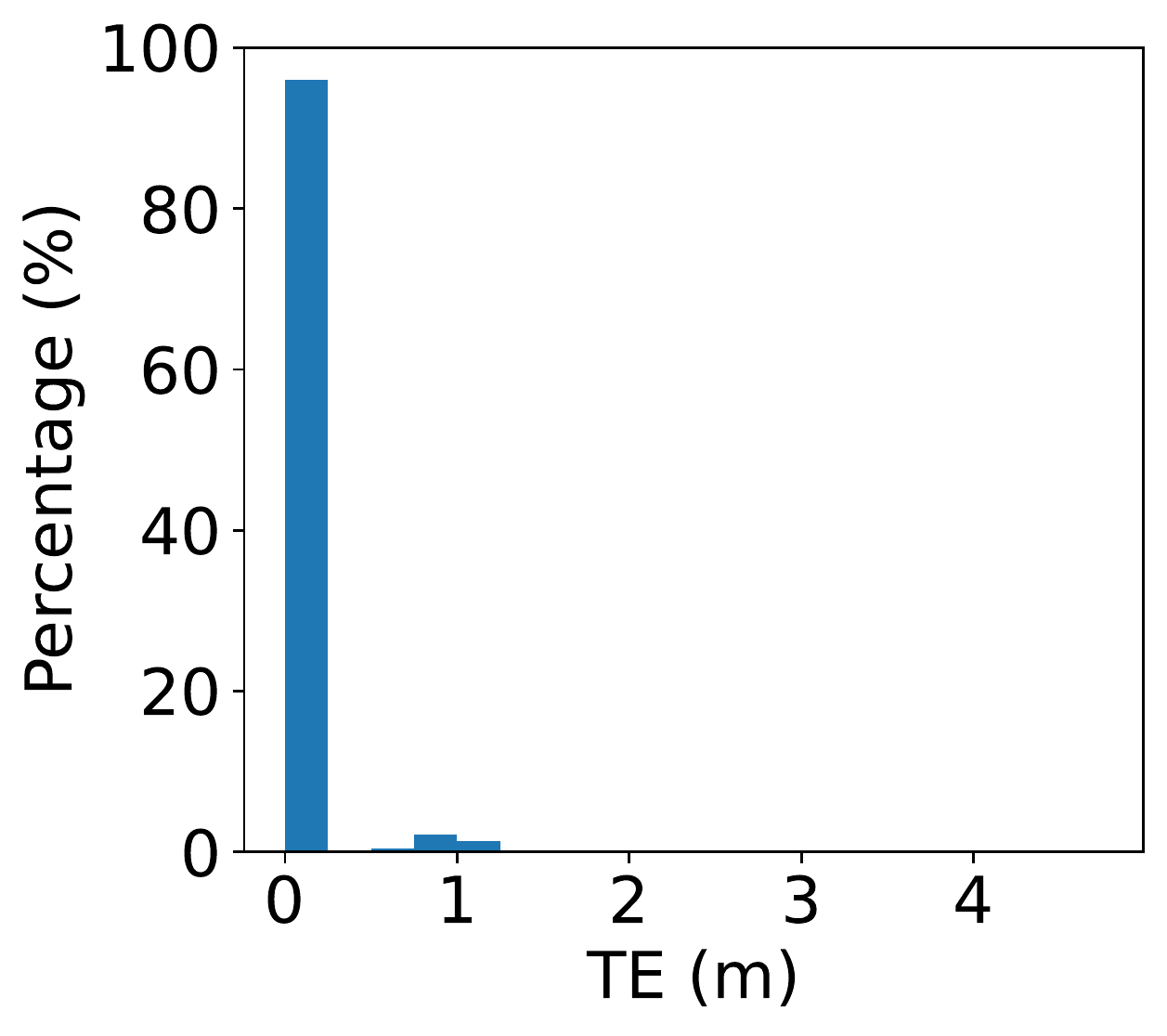}} ;
\node[] (a) at (10.5/6-0.05,1) {(c) ARL \cite{ARL}} ;

\node[] (a) at (10.5/3*2+0.1,2.5) {\includegraphics[width=0.24\textwidth]{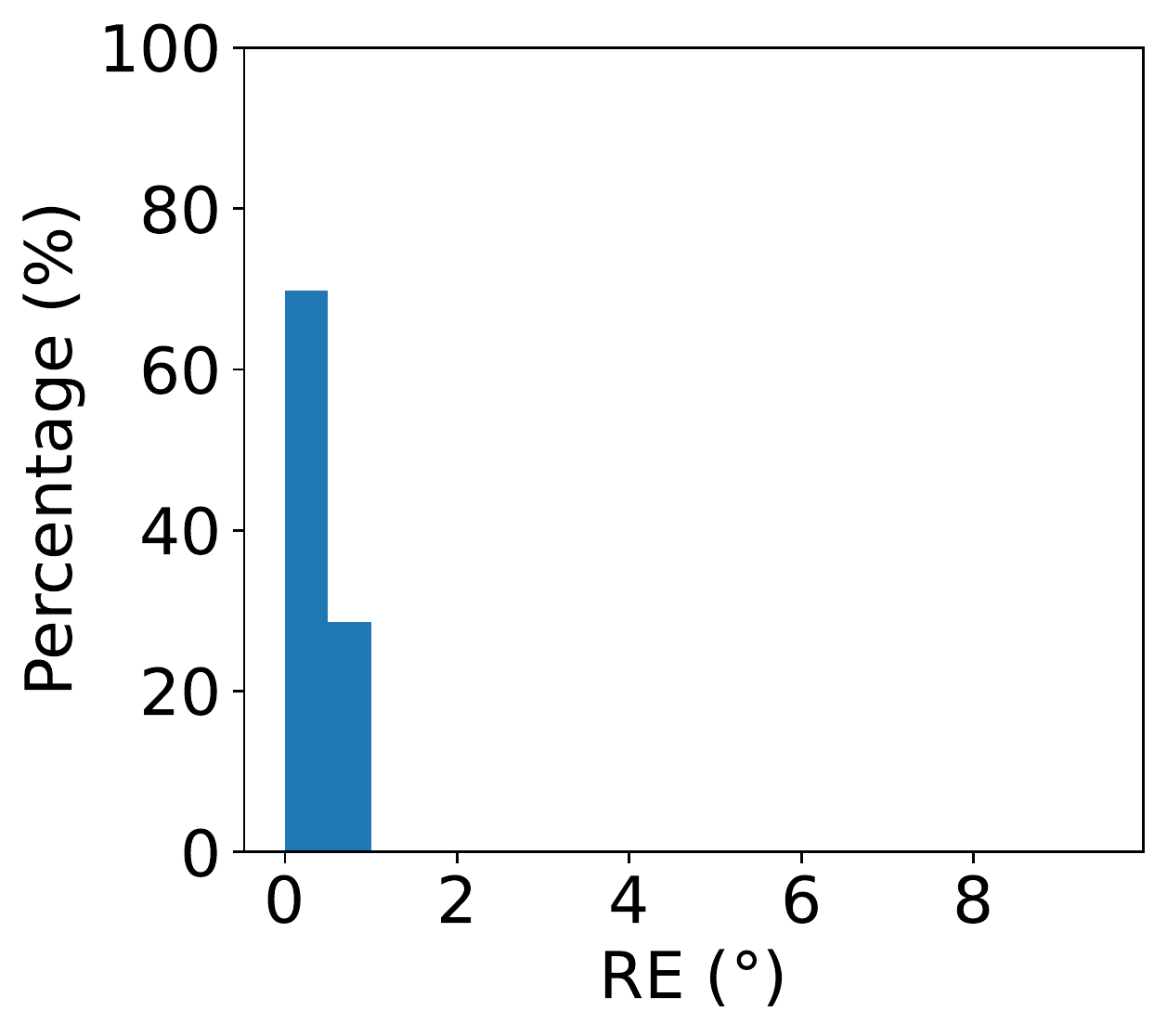}} ;
\node[] (a) at (10.5,2.5) {\includegraphics[width=0.24\textwidth]{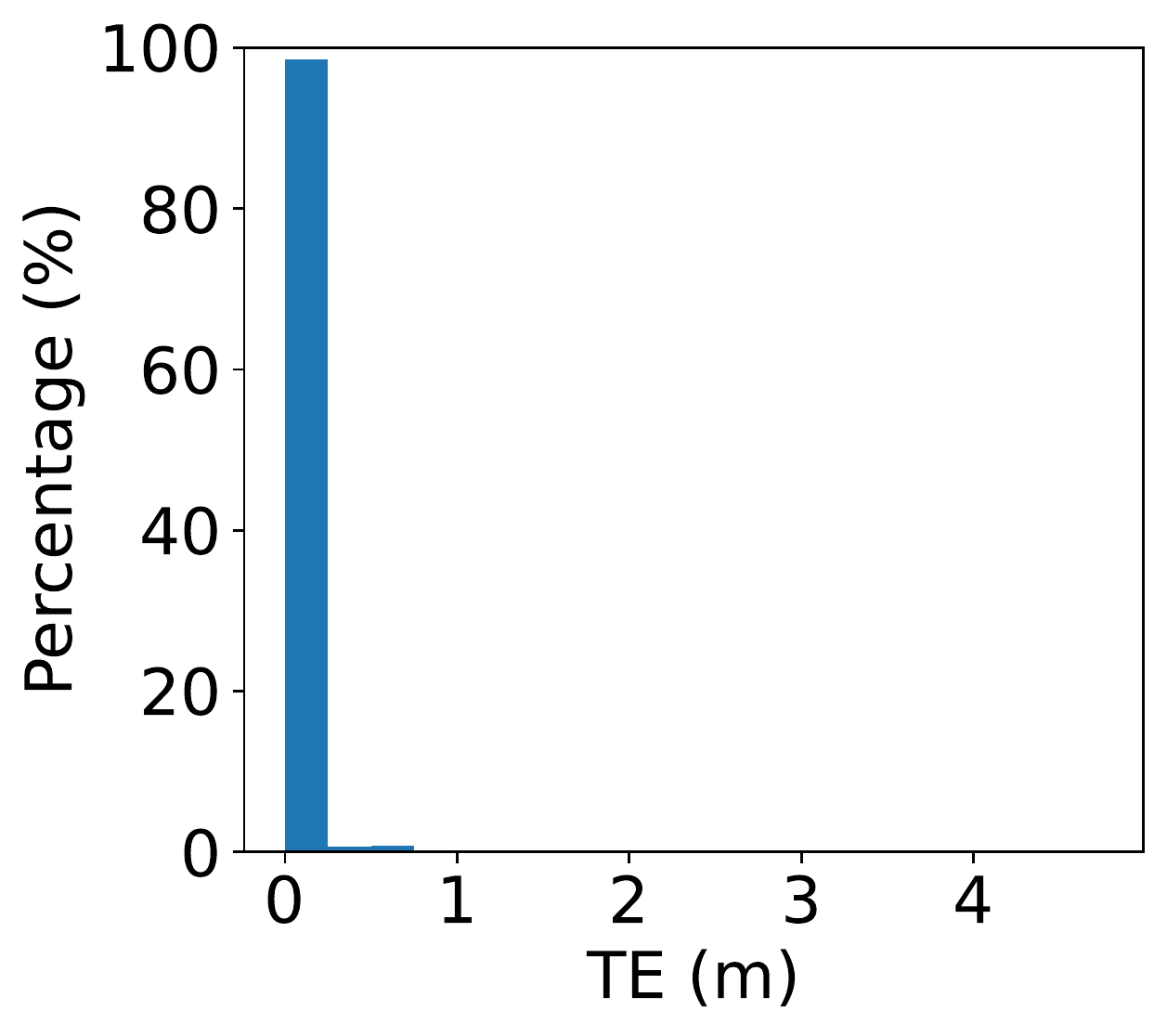}} ;
\node[] (a) at (10.5/6*5+0.05,1) {(d) Ours} ;

\node[] (a) at (0,-1) {\includegraphics[width=0.24\textwidth]{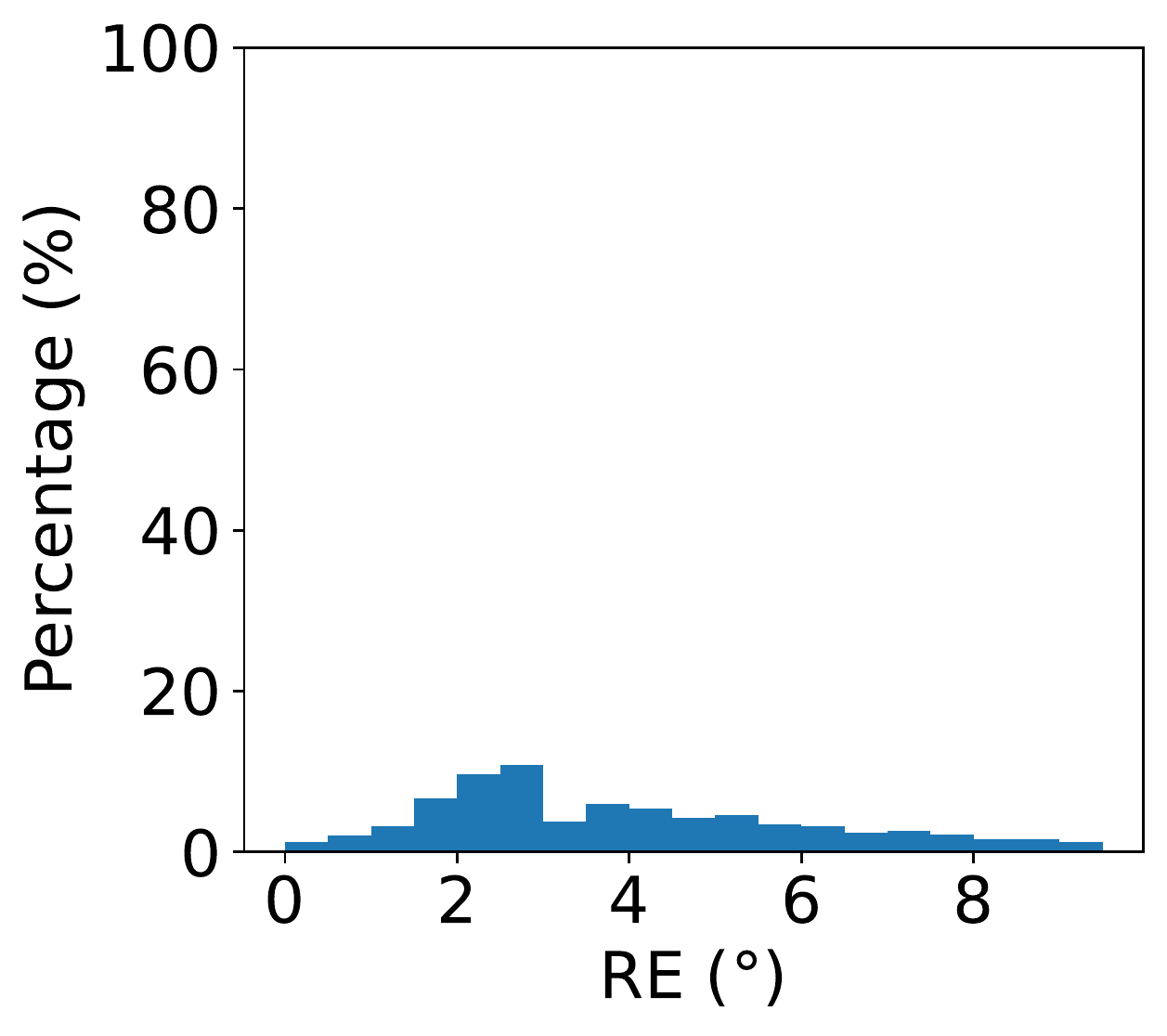}} ;
\node[] (a) at (10.5/3-0.1,-1) {\includegraphics[width=0.24\textwidth]{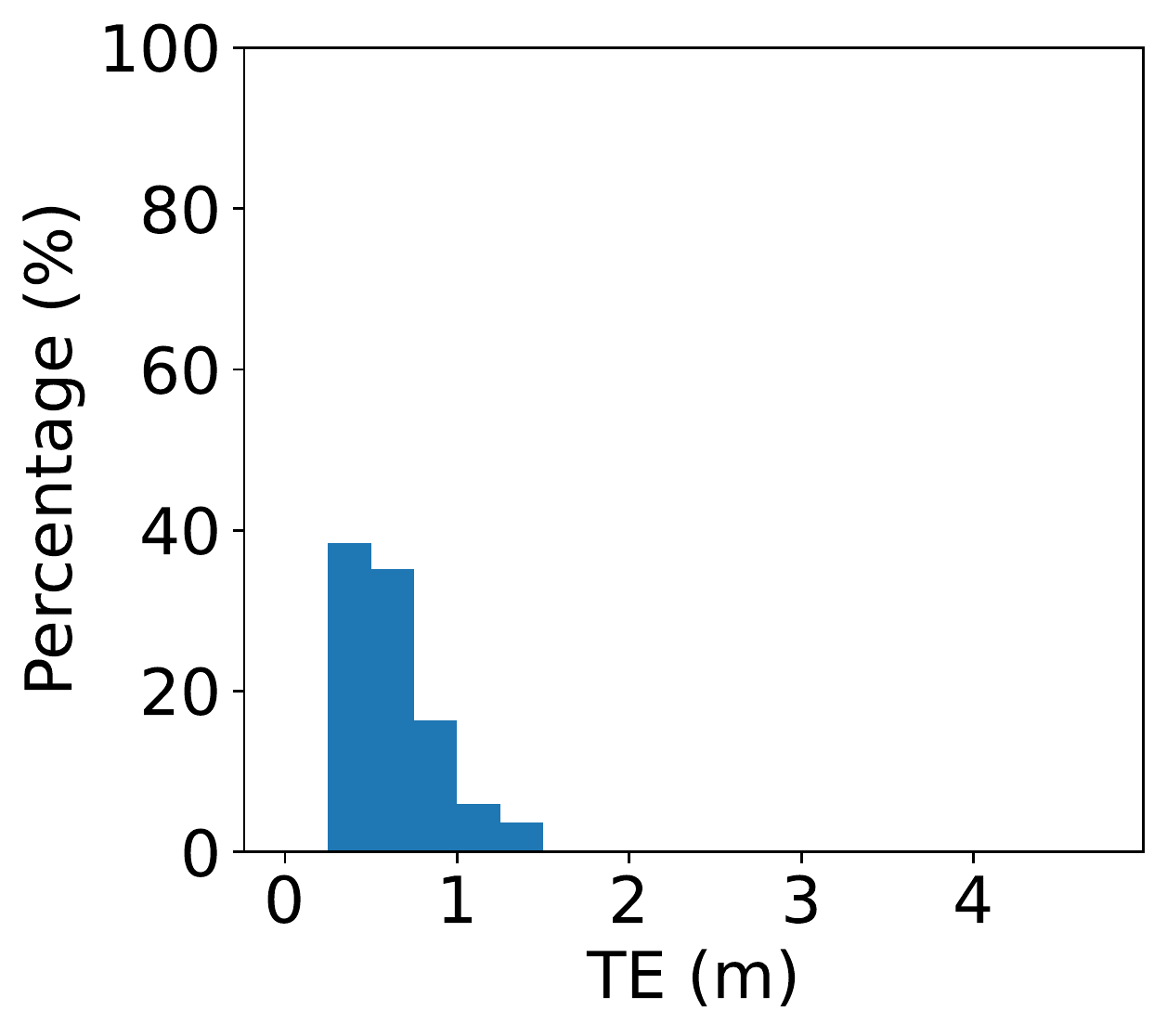}} ;
\node[] (a) at (10.5/6-0.05,-2.5) {(e) RPM-EMD} ;

\node[] (a) at (10.5/3*2+0.1,-1) {\includegraphics[width=0.24\textwidth]{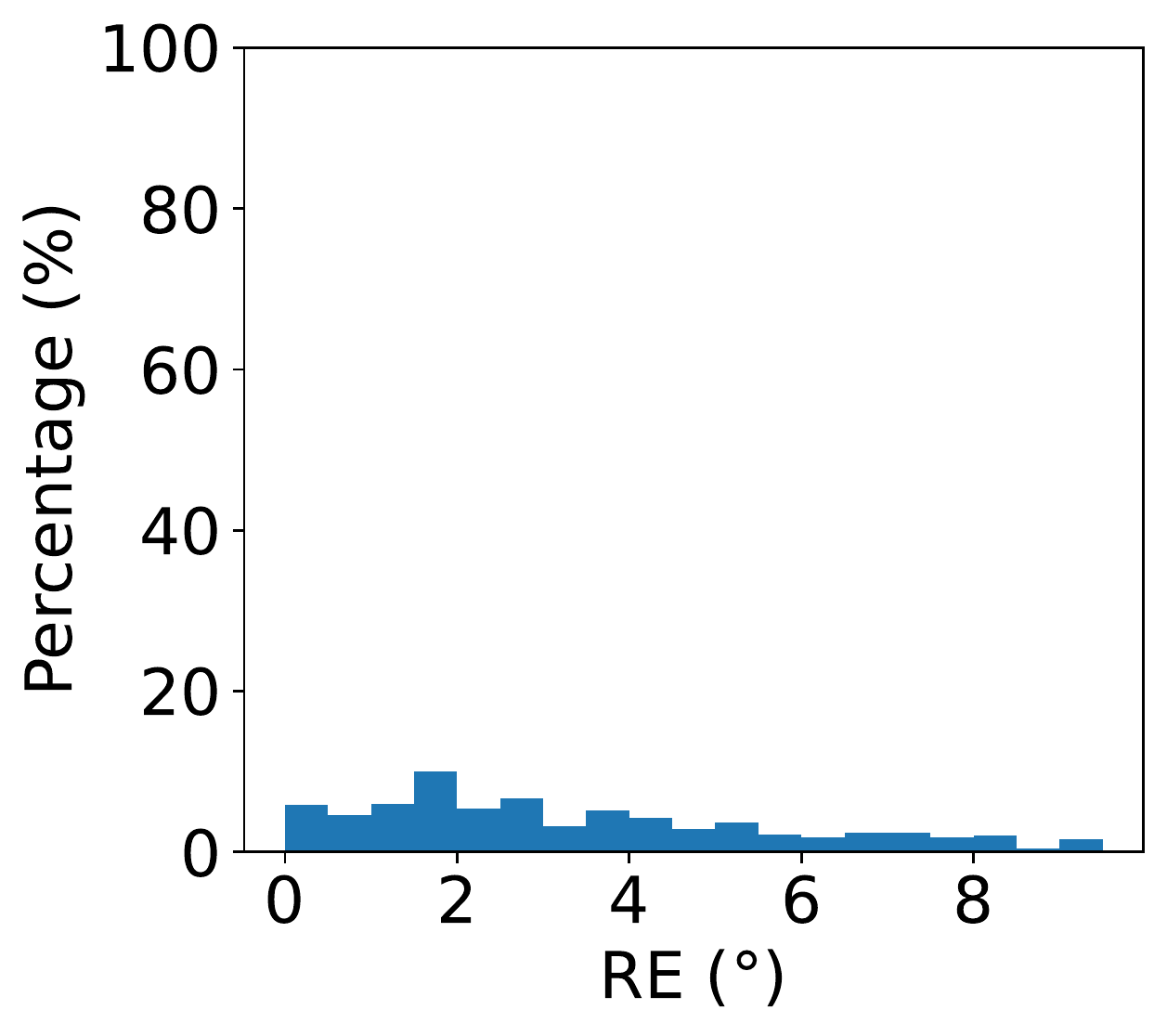}} ;
\node[] (a) at (10.5,-1) {\includegraphics[width=0.24\textwidth]{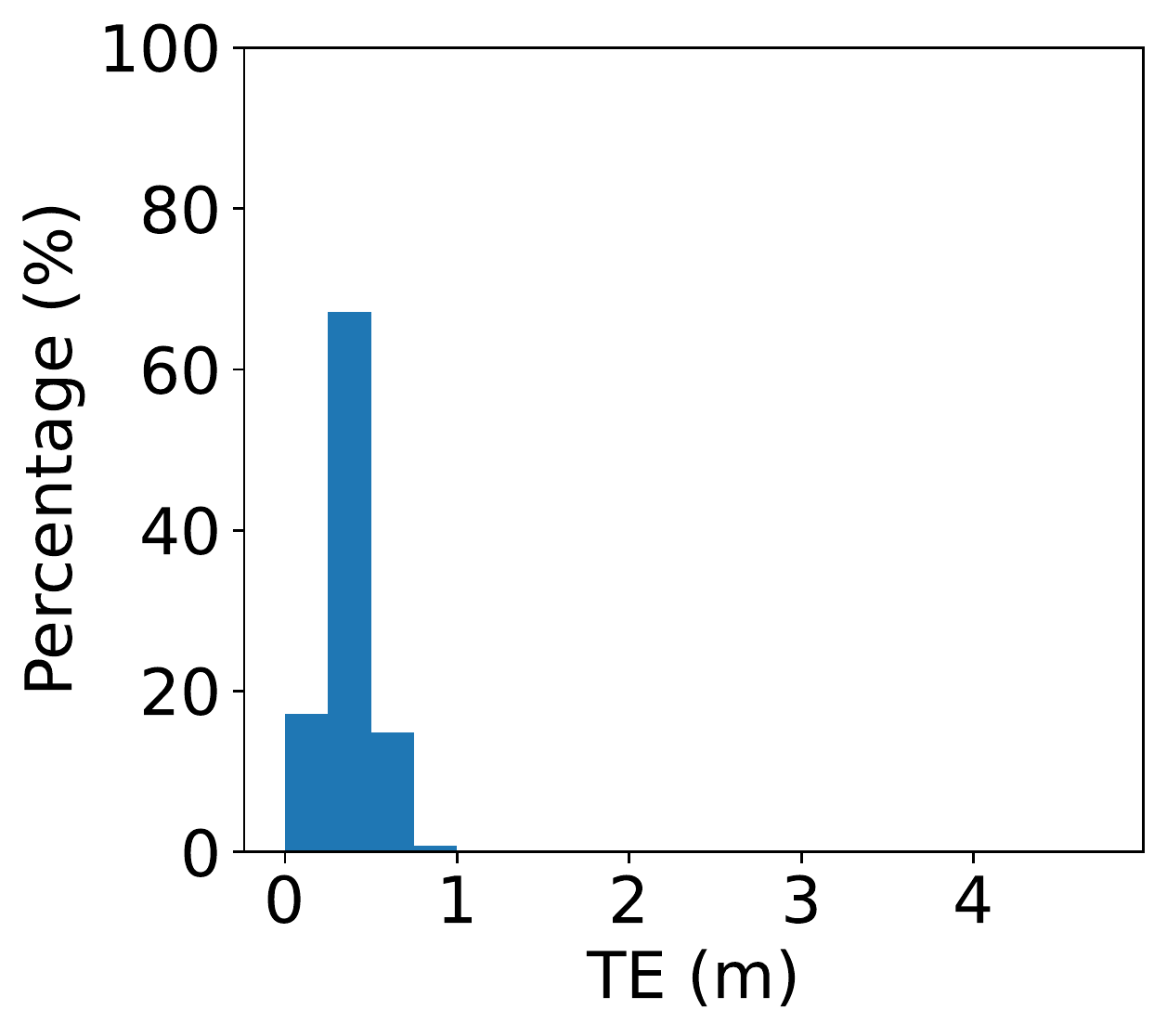}} ;
\node[] (a) at (10.5/6*5+0.05,-2.5) {(f) RPM-CD} ;

\node[] (a) at (0,-4.5) {\includegraphics[width=0.24\textwidth]{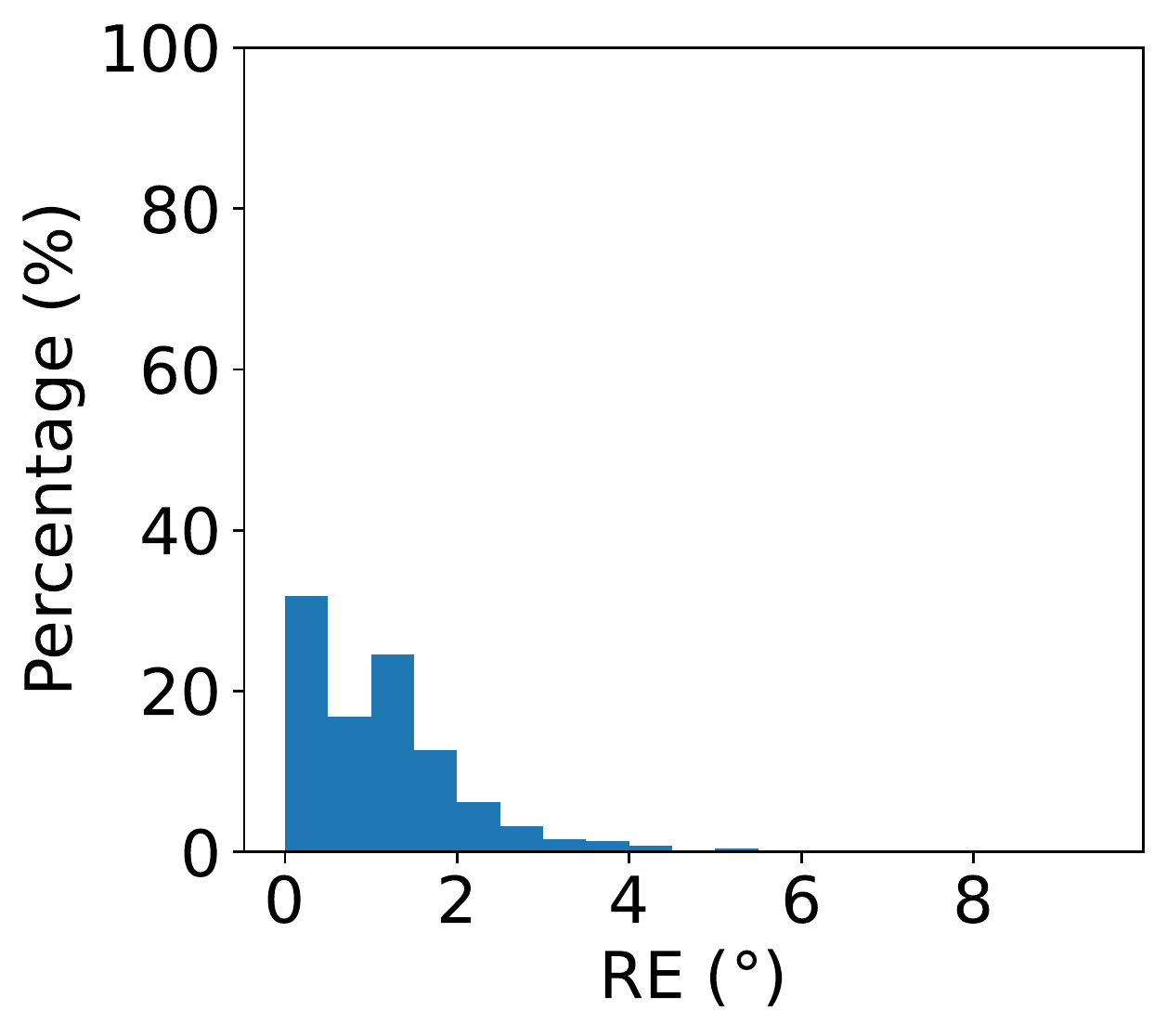}} ;
\node[] (a) at (10.5/3-0.1,-4.5) {\includegraphics[width=0.24\textwidth]{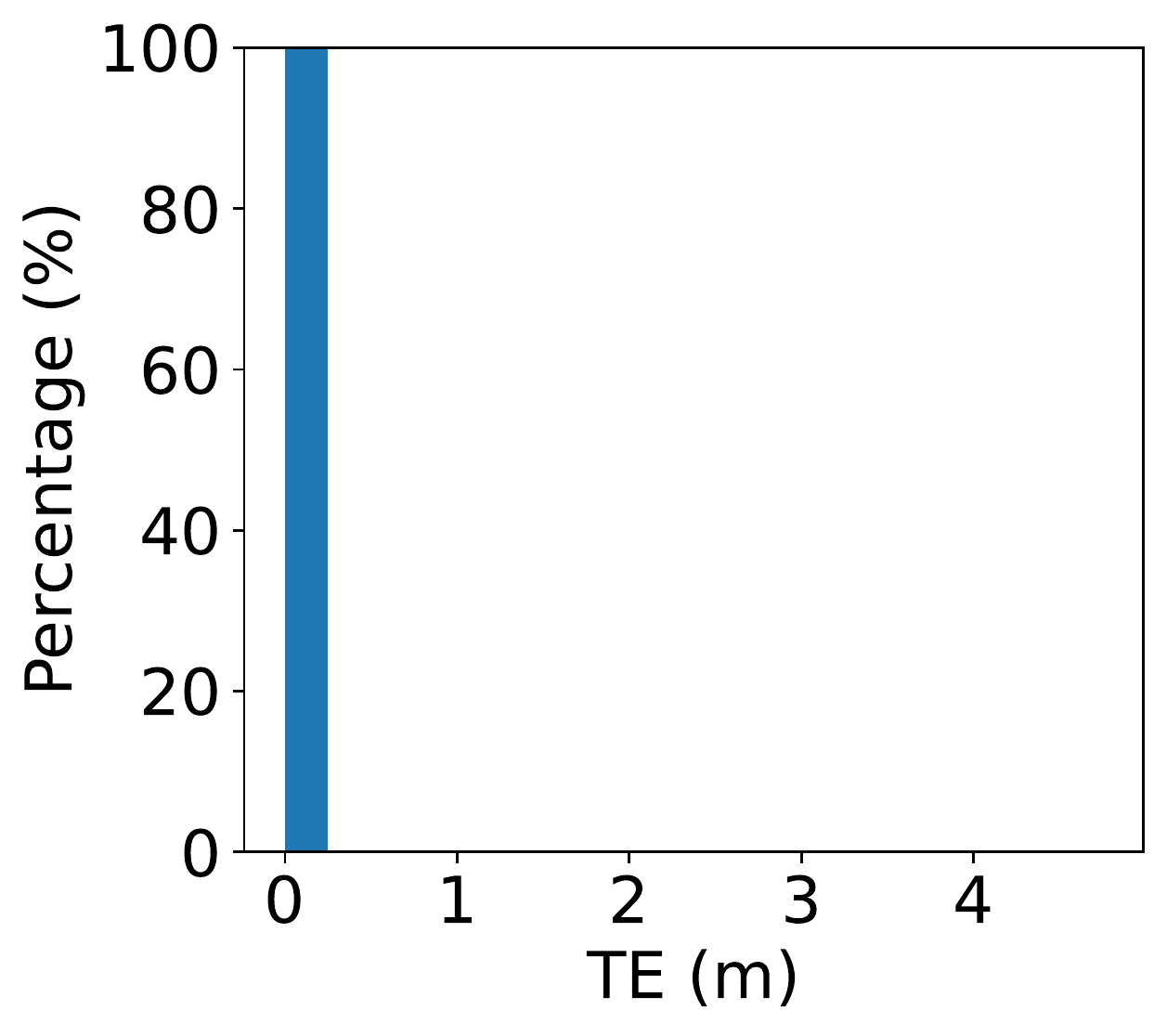}} ;
\node[] (a) at (10.5/6-0.05,-6) {(g) RPM-ARL} ;

\node[] (a) at (10.5/3*2+0.1,-4.5) {\includegraphics[width=0.24\textwidth]{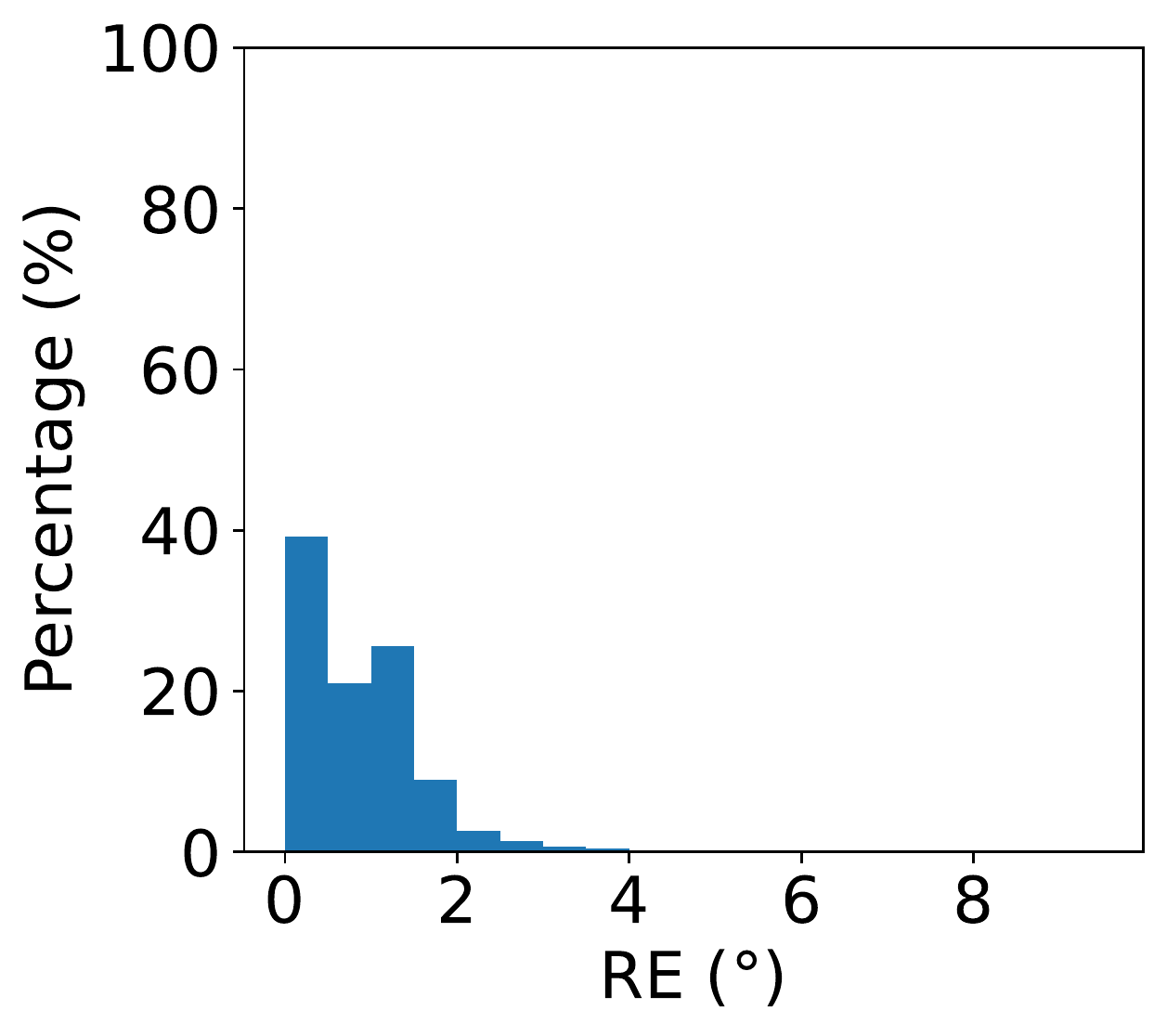}} ;
\node[] (a) at (10.5,-4.5) {\includegraphics[width=0.24\textwidth]{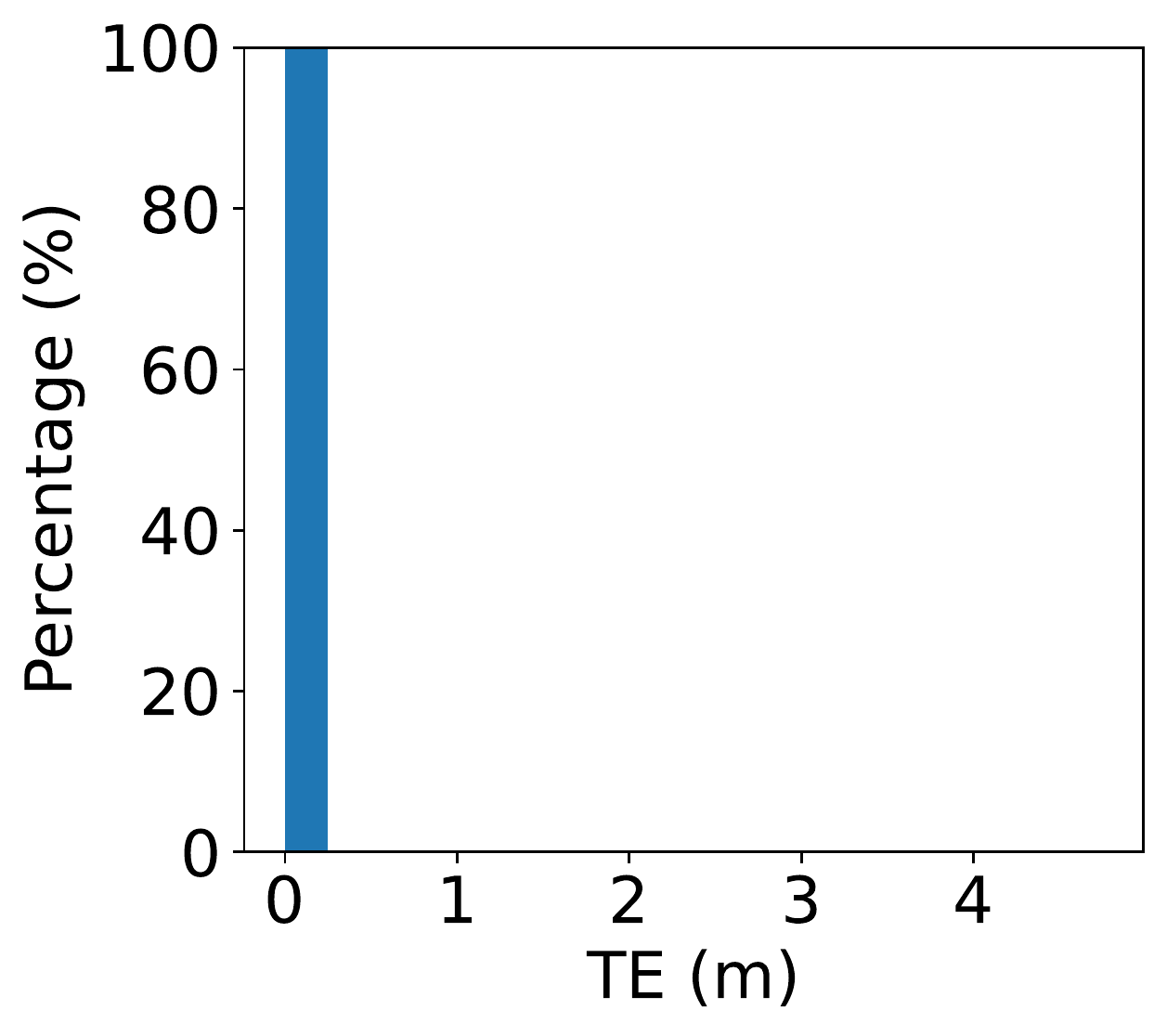}} ;
\node[] (a) at (10.5/6*5+0.05,-6) {(h) RPM-Ours} ;
\end{tikzpicture}
}
\vspace{-0.5cm}
\caption{Histograms of RE and TE on the optimization-based and unsupervised learning-based registration methods.
x-axis is RE($^\circ$) and TE(m), and y-axis is the percentage.} 
\label{REG:HIST} 
\end{figure}
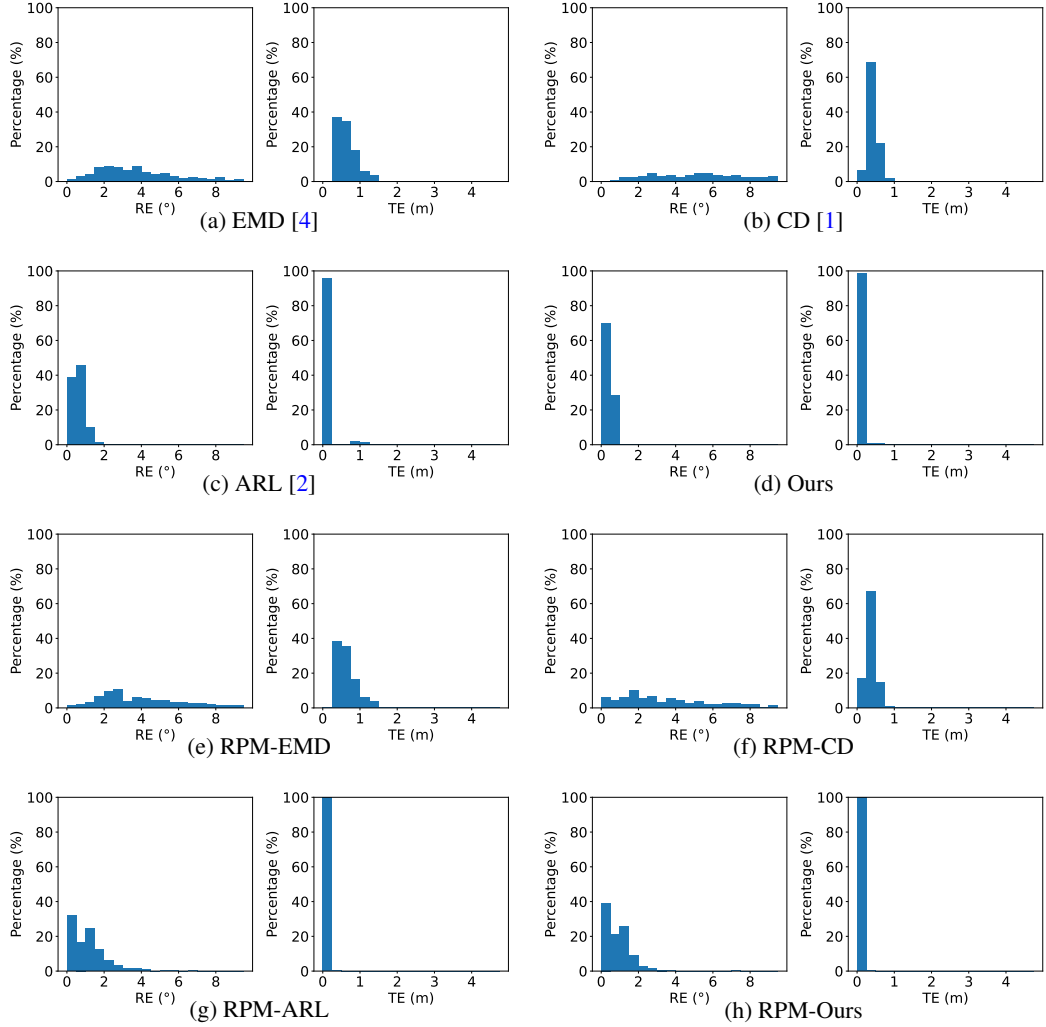

Based on RPM-Net \cite{RPMNET}, 
we construct an unsupervised learning-based rigid registration framework by modifying 
its loss, 
as shown in Fig. \ref{RPM:STRUC}. 
The loss at $i$-th iteration $\mathcal{L}_i$ is defined as
\begin{equation}
    \mathcal{L}_i=\lambda_1\mathcal{D}(\mathbf{R}_i\mathbf{P}_{\rm src}+\mathbf{t}_i,\mathbf{P}_{\rm tgt})+\lambda_2\mathcal{R}_{\rm inlier}(\mathbf{P}_{\rm src},\mathbf{P}_{\rm tgt}), \label{RPM:LOSS}
\end{equation}
where $\mathcal{R}_{\rm inlier}$ is the inlier regularization in RPM-Net (see Eq. (11) of the original paper of RPM-Net for more details), and the hyperparameters, $\lambda_1$ and $\lambda_2$, are set 10 and 0.01, respectively. Considering all iterations, the total loss is 
\begin{equation}
    \mathcal{L}_{\rm total}=\sum_{i=1}^{N_{\rm iter}}\left(\frac{1}{2}\right)^{(N_{\rm iter}-i)}\mathcal{L}_i,
\end{equation}
assigning later iterations with higher weights.
we set $N_{\rm iter}=2$ and $N_{\rm iter}=5$ during training and inference, respectively. 
\paragraph{Error Distribution.} 
The distributions of the registration error are shown in Fig. \ref{REG:HIST}, where it can be seen that under both optimization and unsupervised learning pipelines, 
the registration errors by our methods 
are concentrated in smaller ranges, compared with other methods.

\section{Scene Flow Estimation}
\paragraph{Spatial Smooth Regularization.}
Given the source point cloud $\mathbf{P}_{\rm src}\in\mathbb{R}^{N_{\rm src}\times 3}$ and its estimated scene flow $\mathbf{F}\in\mathbb{R}^{N_{\rm src}\times 3}$,  we define the spatial smooth regularization term $\mathcal{R}_{\rm smooth}(\mathbf{F})$ as 
\begin{equation}
    \mathcal{R}_{\rm smooth}(\mathbf{F})=\frac{1}{3N_{\rm src}K_{\rm s}}\sum_{\mathbf{x}\in\mathbf{P}_{\rm src}}\sum_{\mathbf{x}'\in\mathcal{N}(\mathbf{x})}\|\mathbf{F}(\mathbf{x})-\mathbf{F}(\mathbf{x}')\|_2^2,
\end{equation}
where $\mathcal{N}(\mathbf{x})$ is the operator returning $\mathbf{x}$'s $K_{\rm s}$-NN points in the $\mathbf{P}_{\rm src}$ with $K_{\rm s}=30$ in our experiments.
\section{Feature Representation}
\paragraph{Network Architecture.}
\begin{figure}[htbp]
    \centering
    \includegraphics[width=0.95\linewidth]{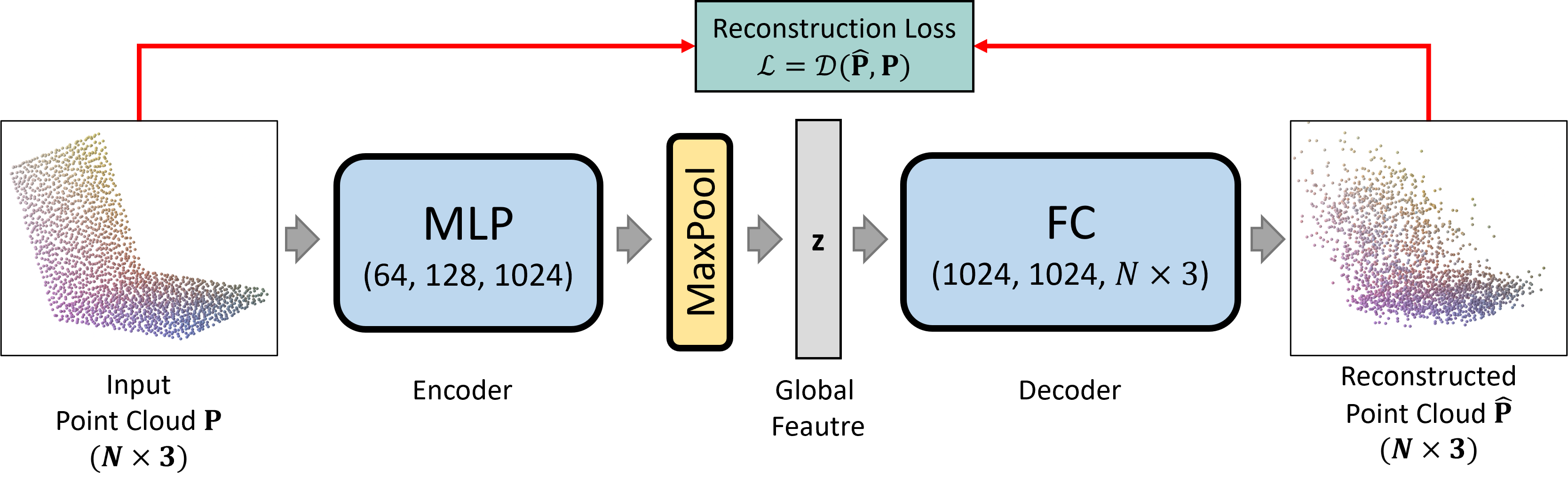}
    \vspace{-0.2cm}
    \caption{The network architecture of the auto-decoder for feature representation.}
    \label{AE:STRUCTURE}
\end{figure}
Fig. \ref{AE:STRUCTURE} shows the network architecture of the auto-decoder for feature representation, 
where 
the encode embeds an input point cloud $\mathbf{P}$ as a 
global feature $\mathbf{z}$, 
and the resulting global feature is then fed into the decoder to get the reconstructed point cloud $\hat{\mathbf{P}}$. The difference between $\mathbf{P}$ and $\hat{\mathbf{P}}$ is employed as the loss to train the auto-decoder: 
\begin{equation}
    \mathcal{L}_{\rm rec}=\mathcal{D}(\hat{\mathbf{P}},~\mathbf{P}),
\end{equation}
where $\mathcal{D}(\cdot,~\cdot)$ is a typical 
point cloud distance metric, e.g.,  EMD, CD, or our CLGD. After training, we adopt SVM to classify the global feature representations of point clouds to achieve classification.

\paragraph{Visualization of the Decoded 
Point Clouds.} 
\begin{figure}[htbp] \small 
\centering
{
\begin{tikzpicture}[]

\node[] (a) at (0,6) {\includegraphics[width=0.24\textwidth]{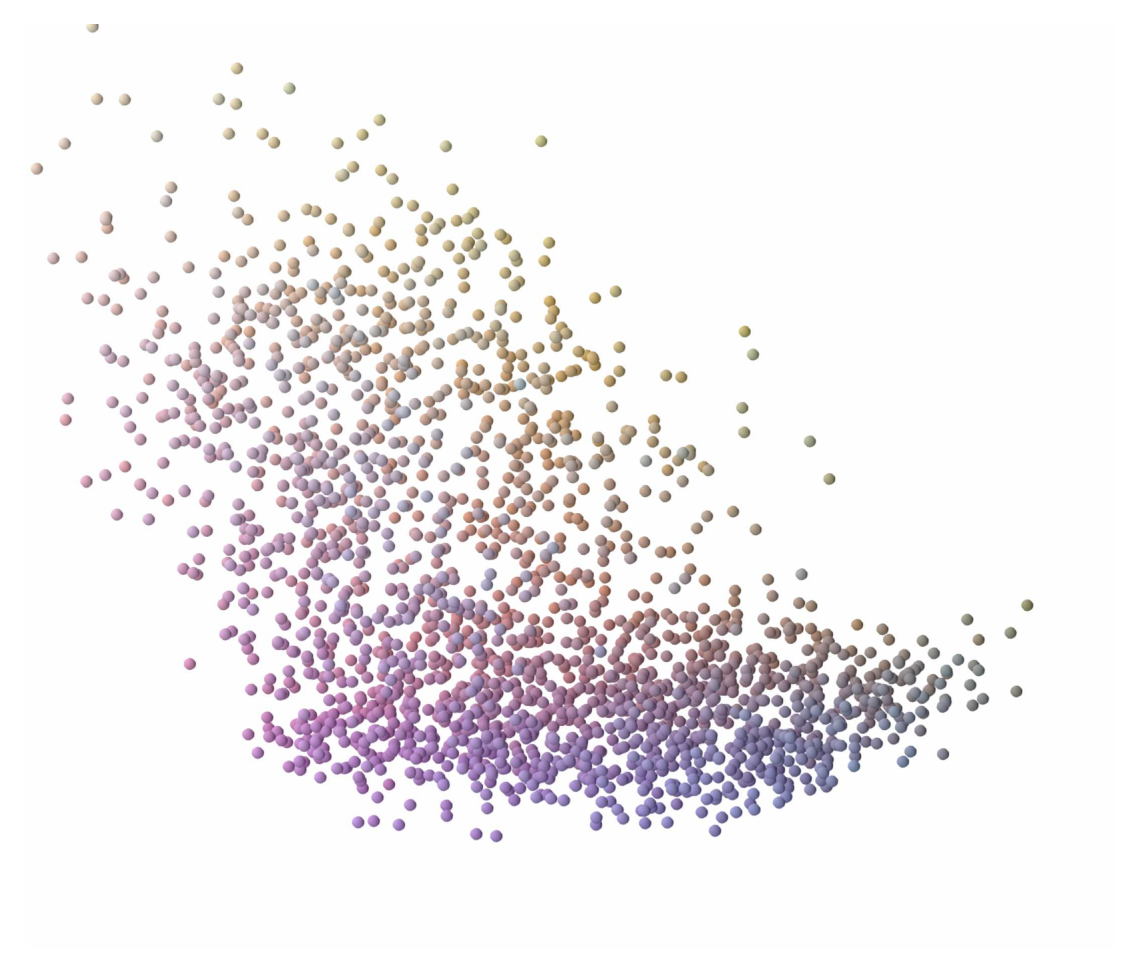}} ;
\node[] (a) at (0,2.3) {\includegraphics[width=0.24\textwidth]{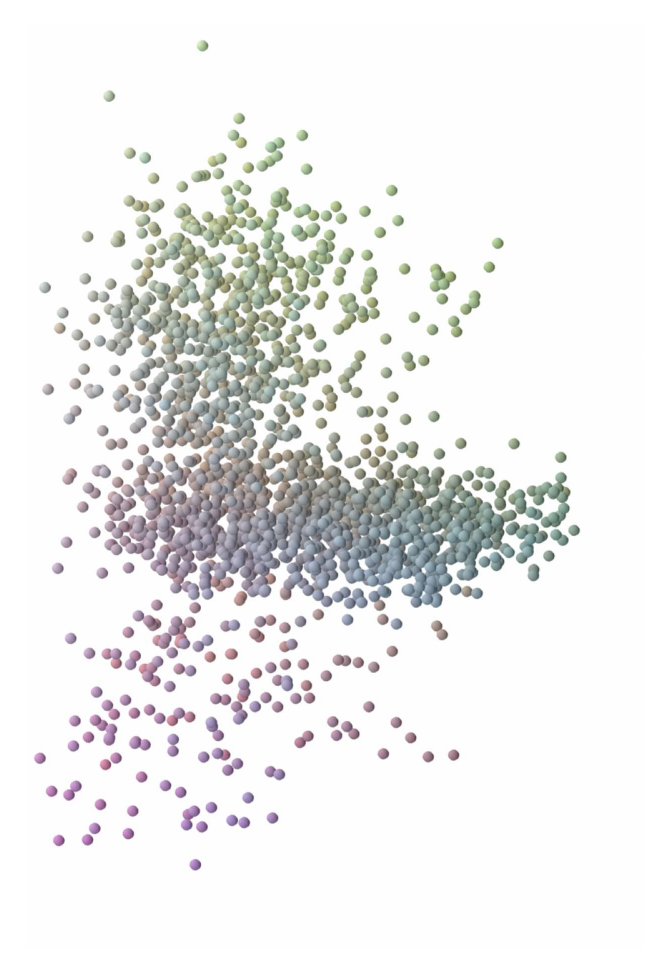}} ;
\node[] (a) at (0,-0.1) {(a) EMD} ;

\node[] (b) at (10.5/3,6) {\includegraphics[width=0.24\textwidth]{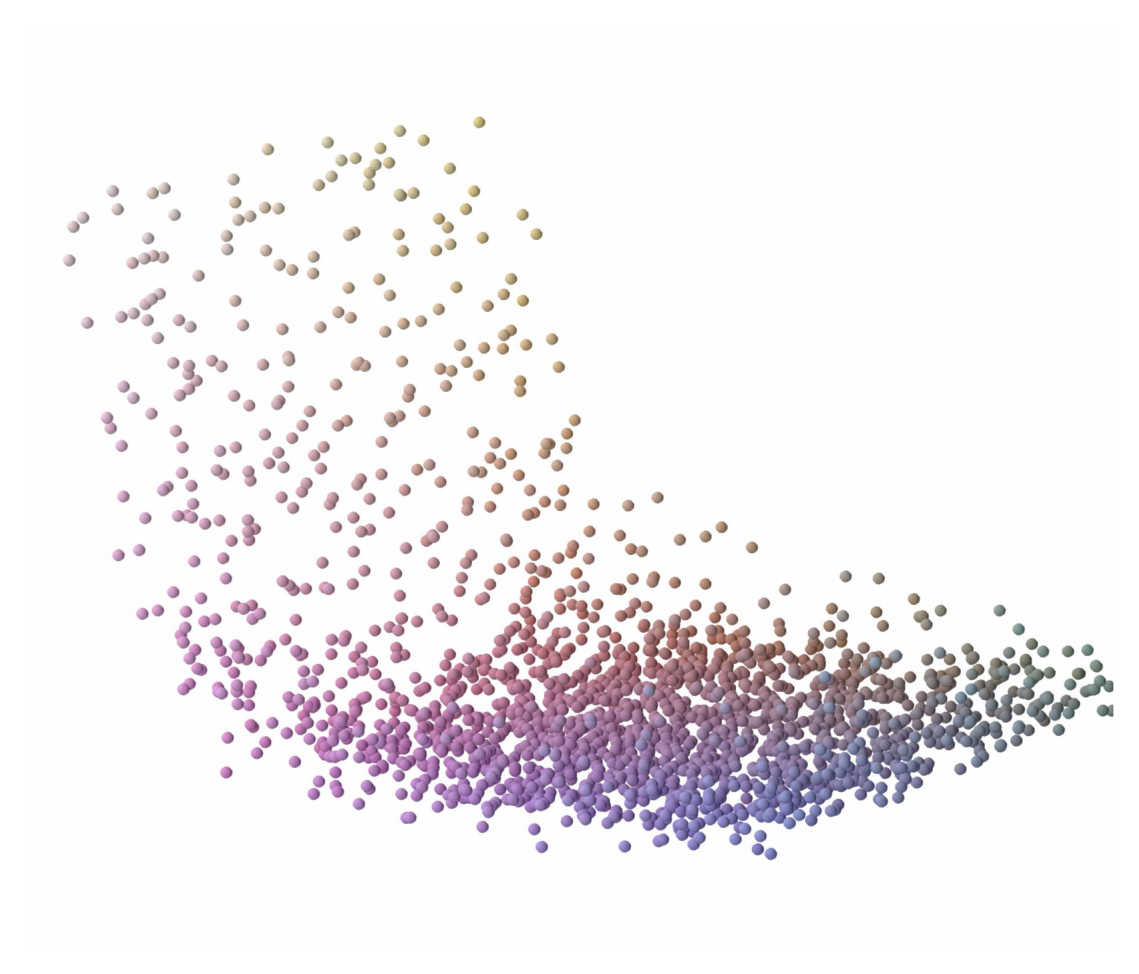}};
\node[] (b) at (10.5/3,2.3) {\includegraphics[width=0.24\textwidth]{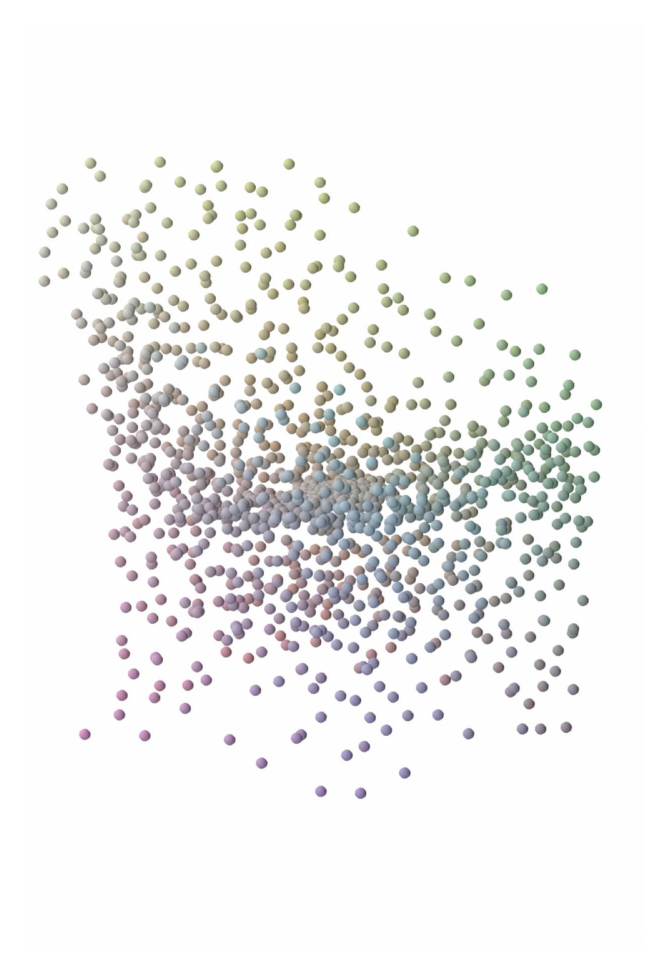}};
\node[] (b) at (10.5/3,-0.1) {(b) CD};

\node[] (c) at (10.5/3*2,6) {\includegraphics[width=0.24\textwidth]{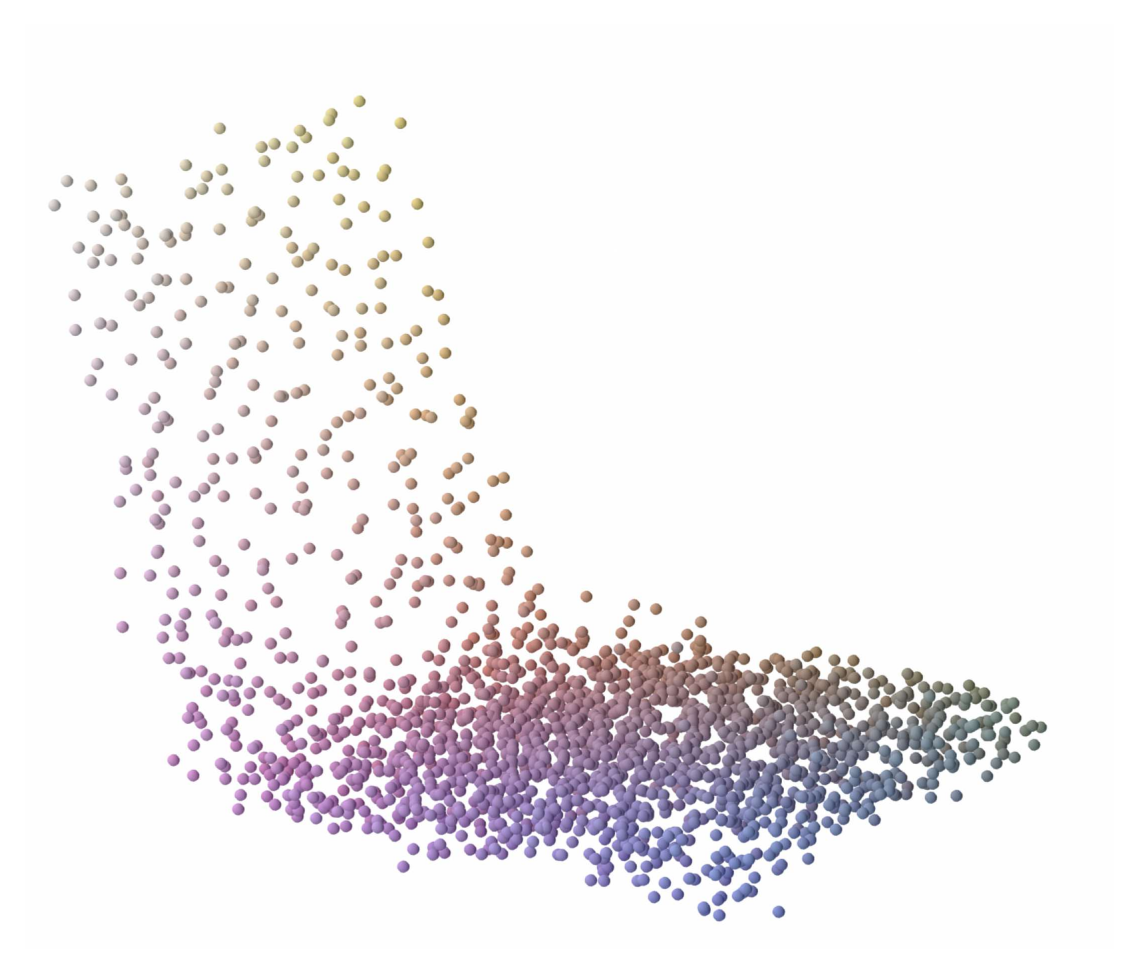}};
\node[] (c) at (10.5/3*2,2.3) {\includegraphics[width=0.24\textwidth]{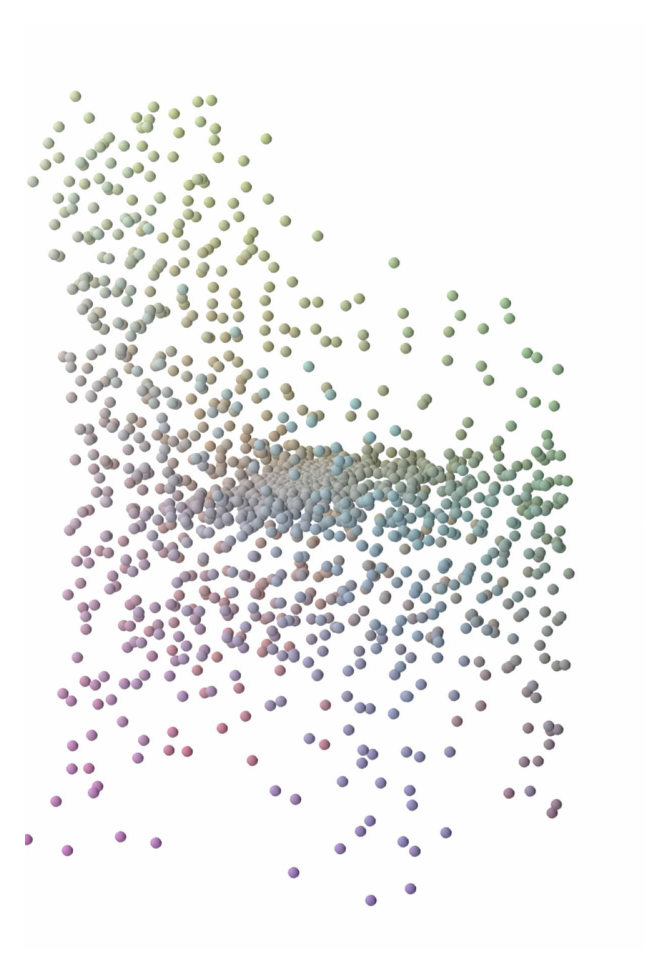} };
\node[] (c) at (10.5/3*2,-0.1) {(c) Ours };

\node[] (d) at (10.5/3*3,6) {\includegraphics[width=0.24\textwidth]{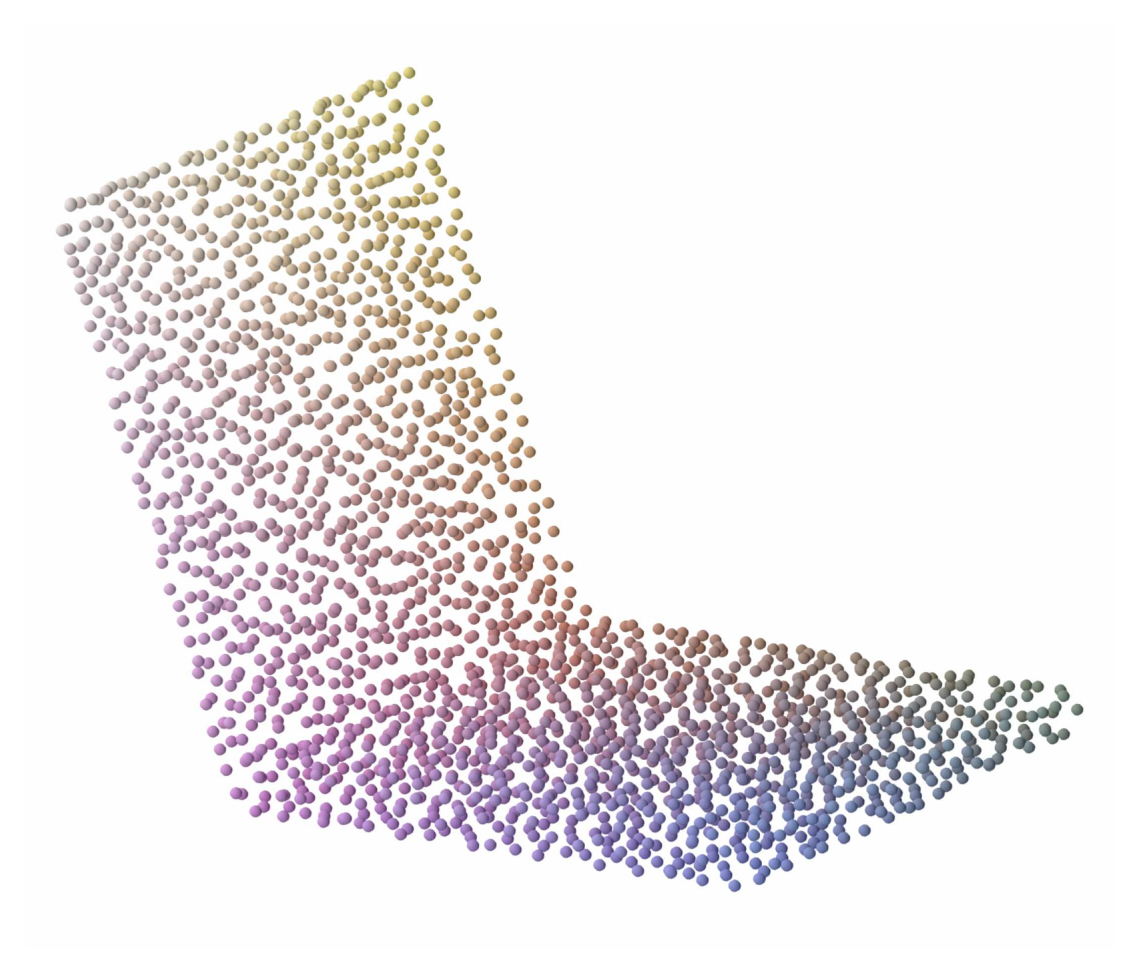}};
\node[] (d) at (10.5/3*3,2.3) {\includegraphics[width=0.24\textwidth]{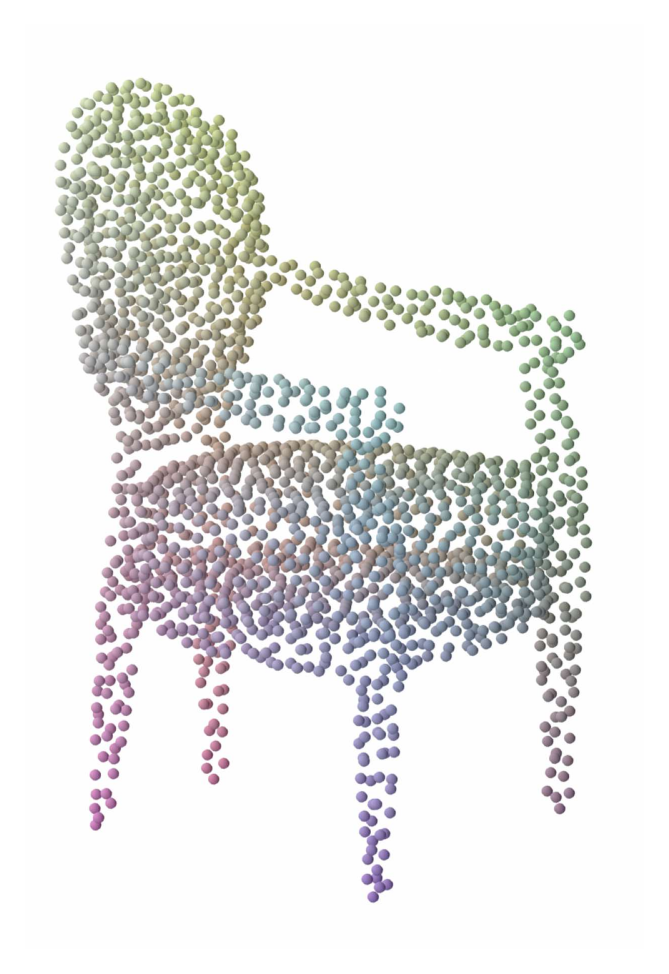} };
\node[] (d) at (10.5/3*3,-0.1) {(d) Input/GT };
\end{tikzpicture}
}
\vspace{-0.5cm}
\caption{Visual comparisons of decoded point clouds by the auto-encoder trained with different distance metrics.} 
\label{AE:FIG} 
\end{figure}
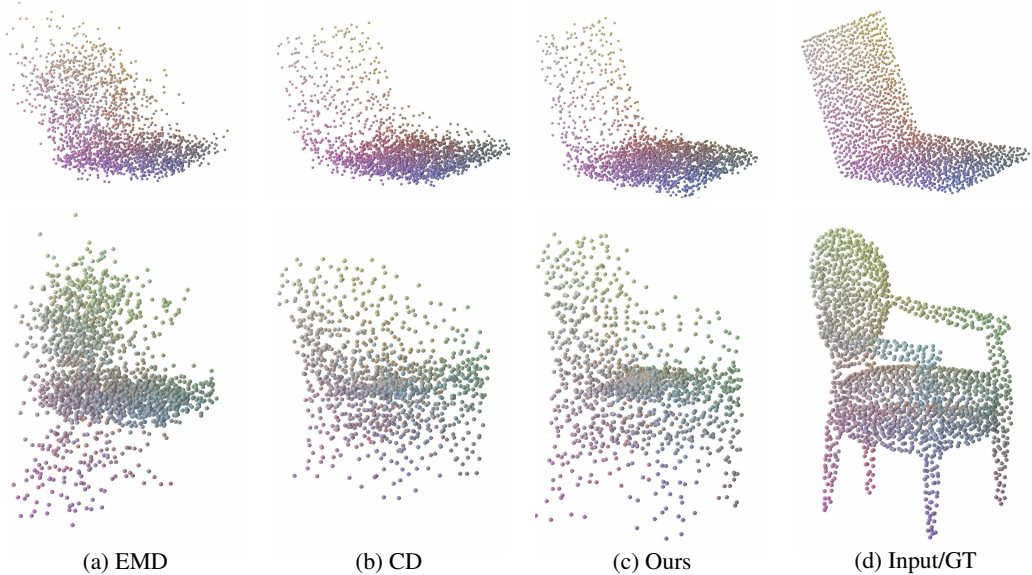
We also show the decoded results by the auto-decoder after trained with different distance metrics in Fig. \ref{AE:FIG}, where it can be seen that the auto-encoder trained with our CLGD can decode point clouds that are closer to input/ground-truth ones during inference, demonstrating its advantage. 

{
\small
\bibliographystyle{abbrv}
\bibliography{ref}
}